\documentclass[a4paper,fleqn]{cas-sc}

 \usepackage[numbers]{natbib}

\usepackage{graphicx}
\usepackage{hyperref}
\usepackage{amssymb}
\usepackage{amsmath}
\usepackage{bm}
\usepackage{subcaption} 

\DeclareCaptionLabelFormat{empty}{}
\def\tsc#1{\csdef{#1}{\textsc{\lowercase{#1}}\xspace}}
\tsc{WGM}
\tsc{QE}
\tsc{EP}
\tsc{PMS}
\tsc{BEC}
\tsc{DE}

\begin{document}
\let\WriteBookmarks\relax
\def\floatpagepagefraction{1}
\def\textpagefraction{.001}

\shorttitle{Physics-informed KAN PointNet}

\shortauthors{A. Kashefi \& T. Mukerji}

\title [mode = title]{Physics-informed KAN PointNet: Deep learning for simultaneous solutions to inverse problems in incompressible flow on numerous irregular geometries}




%

\author[1]{Ali Kashefi}
[orcid=0000-0003-0014-9051]
\cormark[1]
\ead{kashefi@stanford.edu}
\address[1]{Stanford University, Stanford, 94305, CA, USA}



\author
[1]
{Tapan Mukerji}[orcid=0000-0003-1711-1850]
\ead{mukerji@stanford.edu}



\cortext[cor1]{Corresponding author}


\begin{abstract}
Kolmogorov-Arnold Networks (KANs) have gained attention as a promising alternative to traditional multilayer perceptrons (MLPs) for deep learning applications in computational physics, particularly for solving inverse problems with sparse data, as exemplified by the physics-informed Kolmogorov-Arnold network (PIKAN). However, the capability of KANs to simultaneously solve inverse problems over multiple irregular geometries within a single training run remains unexplored. To address this gap, we introduce the physics-informed Kolmogorov-Arnold PointNet (PI-KAN-PointNet), in which shared KANs are integrated into the PointNet architecture to capture the geometric features of computational domains. The loss function comprises the squared residuals of the governing equations, computed via automatic differentiation, along with sparse observations and partially known boundary conditions. We construct shared KANs using Jacobi polynomials and investigate their performance by considering Jacobi polynomials of different degrees and types in terms of both computational cost and prediction accuracy. As a benchmark test case, we consider natural convection in a square enclosure with a cylinder, where the cylinder's shape varies across a dataset of 135 geometries. PI-KAN-PointNet offers two main advantages. First, it overcomes the limitation of current PIKANs, which are restricted to solving only a single computational domain per training run, thereby reducing computational costs. Second, when comparing the performance of PI-KAN-PointNet with that of the physics-informed PointNet using MLPs, we observe that, with approximately the same number of trainable parameters and comparable computational cost in terms of the number of epochs, training time per epoch, and memory usage, PI-KAN-PointNet yields more accurate predictions, particularly for values on unknown boundary conditions involving nonsmooth geometries.
\end{abstract}




\maketitle

\section{Introduction and motivation}
\label{Sect1}

Physics-informed Neural Networks (PINNs), first introduced by Raissi et al. in 2019 \cite{raissi2019physics}, are increasingly recognized as a valuable tool for solving inverse problems in diverse scientific and industrial domains. These include solid mechanics \cite{haghighat2021physics,rao2021physics,jiang2022physics,tandale2022physics,vadyala2021review,xu2023transfer,flaschel2022discovering,cao2022physics,wu2022effective,bai2022physics,he2023mflp,bolandi2022physics,niu2023modeling,rezaei2022mixed,jeong2023physics,qiu2023sensenet,fernandez2023physics}, both incompressible and compressible flow \cite{jin2021nsfnets,kashefi2022physics,lou2021physics,jagtap2022physics,jagtap2020conservative,raissi2019physics,patel2022thermodynamically,qiu2022physics,ouyang2023reconstruction,buhendwa2021inferring}, chemistry \cite{weng2022multiscale,ji2021stiff}, heat transfer \cite{kashefi2022physics,wang2021reconstruction,cai2021physics}, and flow in porous media \cite{kashefi2022prediction,almajid2022prediction,yu2022gradient}. The central idea behind PINNs for inverse problems is to train a neural network to minimize both the residuals of the governing partial differential equations and the distance between the solution provided by the neural network and sparse observational data at sensor locations in specific norms such as the Euclidean norm. Additional constraints, such as boundary or initial conditions, may also be incorporated into this optimization process. The choice of the neural network architecture in PINNs significantly influences the performance and capability of PINN models. A prevalent choice is the fully connected neural network \cite{rao2021physics,haghighat2021physics,xu2021explore,yuan2022pinn,linka2022bayesian,eivazi2022physics}. However, a notable limitation of using fully connected networks is that they are generally suited to solving inverse problems for a single specific geometry. For each new geometry, the network must be retrained from the ground up, which incurs significant computational costs. This challenge is particularly pronounced when exploring a broad range of geometric parameters for optimizing industrial designs. The issue of high computational expense in such scenarios was addressed by \citep{gao2021phygeonet,kashefi2022physics,kashefi2023PIPNelasticity}. To resolve this issue, \citet{kashefi2022physics} introduced Physics-informed PointNet (PIPN). In PIPN, instead of using a fully connected neural network, PointNet \cite{qi2017pointnet,kashefi2021PointNet} is employed. Since PointNet \cite{qi2017pointnet} and its advanced versions \cite{qi2017pointnet++,qi2017frustum} can capture the geometric features of inputs in a latent space, they can differentiate between various geometries. As a result, inverse problems across multiple geometries can be solved simultaneously in a single training process. \citet{kashefi2022physics} solved an inverse heat transfer convection problem over 108 irregular geometries simultaneously. Similarly, \citet{kashefi2023PIPNelasticity} used PIPN to solve an inverse linear elasticity problem over 532 domains with irregular geometries. The architecture of PointNet \cite{qi2017pointnet}, derived from the field of computer graphics, is more complex than fully connected neural networks and involves several components. The primary components of PointNet are fundamentally built using shared Multilayer Perceptrons (MLPs) \cite{goodfellow2016deep}. In this article, we introduce  Physics-Informed Kolmogorov-Arnold PointNet (PI-KAN-PointNet), where PointNet replaces MLPs with Kolmogorov-Arnold Networks (KANs) \citep{liu2024kan,liu2024kan2}.

KANs \citep{liu2024kan,liu2024kan2} have recently been introduced as a novel alternative to conventional MLPs \citep{cybenko1989approximation,hornik1989multilayer,Goodfellow2016}. KANs are rooted in the Kolmogorov-Arnold representation theorem \citep{arnold2009representation,arnold2009functions,kolmogorovSuperposition,hecht1987kolmogorov,girosi1989representation,braun2009constructive,ismayilova2024kolmogorov,borri2024one}. Unlike MLPs, which rely on training weights and biases with fixed activation functions, KANs focus on training the activation functions themselves \citep{liu2024kan}. KANs have been incorporated into Convolutional Neural Networks (CNNs) \citep{azam2024suitability,bodner2024CNNkan}, graph neural networks \citep{kiamari2024gkan,bresson2024kagnns,zhang2024graphKAN,de2024kolmogorovGraph}, and PointNet \citep{kashefi2024KANpointnet,kashefi2024pointnetKAN3D}. This approach has proven effective across a variety of fields, including physics-informed machine learning \citep{wang2024KANinformed,shukla2024comprehensive,howard2024finite,toscano2024inferring,wang2024kolmogorov,rigas2024adaptive,patra2024physics,toscano2024kkans,yu2024sinc,guo2024physicsR1,nie2025pd,zhang2025physics,cui2025physics,zhang2025physicsV2,shuai2025physics}, deep operator networks \citep{abueidda2024deepokan,shukla2024comprehensive}, neural ordinary differential equations \citep{koenig2024kan,koenig2025leankan}, image classification \citep{azam2024suitability,cheon2024kolmogorovRemote,seydi2024exploringPolynomial,seydi2024KANwavelets,ta2024bsrbf,cheon2024Vision,lobanov2024hyperkan,yu2024kan,ExploreClassification,altarabichi2024rethinking}, image segmentation \citep{li2024UKAN,tang20243d}, image detection \citep{wang2024spectralkan}, audio classification \citep{yu2024kan}, and numerous other scientific and industrial applications \citep{bozorgasl2024wav,vaca2024kolmogorov,genet2024tkan,samadi2024smooth,liu2024ikan,li2024KANradial,aghaei2024Fractional,xu2024kolmogorovPower,xu2024fourierkan,peng2024predictivePump,genet2024temporalKAN,nehma2024leveragingOperator,herbozo2024kan,liu2024initialHumanActivity,poeta2024Table,kundu2024KANquantum,aghaei2024RationalKAN,li2024coeff,pratyush2024calmphoskan,liu2024complexity}. Concerning the usage of KANs in the PointNet framework, \citet{kashefi2024pointnetKAN3D} and \citet{shi2025kanGraphic} proposed, respectively, PointNet-KAN and PointKAN for classification and segmentation of three-dimensional point sets in the area of computer graphics. Moreover, \citet{kashefi2024KANpointnet} introduced KA-PointNet, wherein KAN layers were integrated into PointNet for the supervised learning of fluid dynamic fields. The model employed a loss function solely based on the mean squared error between the predicted and reference velocity and pressure fields. However, no prior work has used KAN layers in PointNet for physics-informed machine learning. In this article, we propose PI-KAN-PointNet, which combines KANs with PointNet and incorporates a loss function based on physical laws to solve inverse problems on irregular geometries, where only sparse data are available.

Several researchers have made significant advancements in the application of KANs in physics-informed deep learning, particularly in solving inverse problems. \citet{wang2024kolmogorov} introduced a physics-informed Kolmogorov-Arnold network for solving inverse problems in solid mechanics, such as nonlinear hyperelasticity. \citet{shukla2024comprehensive} used a physics-informed Kolmogorov-Arnold network with Chebyshev polynomials as the basis for KANs to simulate two-dimensional lid-driven cavity problems and solve the Allen-Cahn equation. Similarly, \citet{toscano2024inferring} applied the same version of the physics-informed Kolmogorov-Arnold network to infer velocity and temperature fields in three-dimensional turbulent flow. \citet{howard2024finite} developed a domain decomposition algorithm to reduce the computational cost of solving inverse problems using physics-informed Kolmogorov-Arnold networks. The interpretability of KANs in the context of physics-informed deep learning was explored by \citet{ranasinghe2024ginn}. To improve the efficiency of physics-informed Kolmogorov-Arnold networks, \citet{rigas2024adaptive} introduced a grid adaptation method for constructing B-splines as the basis of KANs.

As mentioned in the first paragraph, we propose PI-KAN-PointNet, a novel architecture that replaces traditional MLPs with KANs within the physics-informed PointNet framework (i.e., PIPN). Although PIPN \citep{kashefi2022physics} demonstrates robust performance in solving inverse problems on multiple sets of irregular geometries, there is room for improvement. One limitation of PIPN \citep{kashefi2022physics} is that the activation function in the final layer is fixed; regardless of the underlying physics, the network must predict the solution using a predetermined function, typically a sigmoid or hyperbolic tangent function. A similar constraint applies to the intermediate layers. In PINNs and PIPN, the hyperbolic tangent function is used in all layers primarily because the Navier–Stokes equations involve a second spatial derivative, and activation functions such as the rectified linear unit (ReLU) would yield a zero second derivative, causing the loss function to diverge. However, the hyperbolic tangent function was chosen mainly due to its availability in deep learning frameworks such as TensorFlow \citep{tensorflow2015-whitepaper} and PyTorch \citep{paszke2019pytorch}, rather than based on a principled understanding of its optimality or its relationship to the physics of the problem. On the other hand, because the activation functions in KAN layers are learnable, they can be optimized during training based on the physics of the problem. In particular, allowing flexibility in the final layer enhances the network’s ability to solve inverse problems, especially when the activation function is responsible for predicting unknown boundary conditions. Furthermore, in the context of geometric deep learning, when the solution depends on the geometry of the problem, such flexibility further aids in predicting variables on boundaries with irregular or non-smooth shapes. This means that the activation functions learn both the physics and the geometric features of the problem. Moreover, depending on the type of partial differential equation, one can select an appropriate polynomial degree for KAN layers, alleviating concerns about the differentiability of the activation function when using automatic differentiation to formulate the governing equations in the loss function of physics-informed PointNet. Additionally, in a deep neural network such as PointNet, different layers can employ varying polynomial degrees, resulting in a diverse set of activation functions tailored to the network’s design, rather than relying on a fixed activation function (or a limited set of options) for all layers.

To construct shared KAN layers in PointNet, we employ Jacobi polynomials. The test case addressed in this study involves solving the inverse problem of natural convection over 135 geometries simultaneously, with the goal of predicting the velocity, pressure, and temperature fields over these 135 domains. The governing equations consist of the conservation of mass, momentum, and energy in two dimensions, which are coupled together. We utilize PyTorch's automatic differentiation \cite{paszke2019pytorch} to incorporate the partial differential equations into the loss function of PI-KAN-PointNet. Additionally, we explore the impact of different types and degrees of Jacobi polynomials on the performance of PI-KAN-PointNet. The depth of the network is examined to determine the optimal architecture for solving this inverse problem and minimizing computational expenses. Moreover, a comprehensive comparison is made between the proposed PI-KAN-PointNet and the physics-informed PointNet with MLPs (i.e., PIPN \cite{kashefi2022physics}). Finally, we examine a hybrid architecture that integrates KAN and MLP layers within the PI-KAN-PointNet framework. Specifically, we analyze the performance of PI-KAN-PointNet, where MLP layers are incorporated in the encoder while KAN layers are used in the decoder, and vice versa. Overall, our key contributions can be summarized as follows:

\begin{itemize}

   \item We introduce PI‑KAN‑PointNet and evaluate its performance against the corresponding physics‑informed PointNet that uses MLPs to solve inverse problems over irregular geometries.

    \item PI‑KAN‑PointNet is a novel framework that, for the first time, integrates KAN, PointNet, and the enforcement of physical laws into the loss function via automatic differentiation.

    \item We assess the efficiency of PI‑KAN‑PointNet in solving steady incompressible flow inverse problems simultaneously across more than a hundred irregular geometries in a single training run, thereby eliminating the need to retrain classical physics-informed KANs for each new domain.

    \item We extensively evaluate the hyperparameters of PI‑KAN‑PointNet, focusing on the polynomial degree and type in the shared KANs, and assess performance in terms of accuracy and computational cost.

    \item We additionally examine a hybrid physics-informed PointNet model that incorporates both KAN and MLP components.
    
    \item We publicly release our code to support reproducibility, enable future research, and facilitate educational use.
    
\end{itemize}

The remainder of this article is organized as follows. Section \ref{Sect2} provides an overview of the governing equations for natural convection. Section \ref{Sect3} details the architecture of KAN layers. Section \ref{Sect4} elaborates on integrating KAN layers into PointNet and establishing PI-KAN-PointNet for solving inverse problems. Section \ref{Sect5} covers the computational setup, dataset, and details of the deep learning framework. Section \ref{Sect61} presents an analysis of PI-KAN-PointNet’s performance, including the impact of various parameters on result accuracy. Section \ref{Sect62} compares PI-KAN-PointNet with physics-informed PointNet using MLPs. Section \ref{Sect63} discusses the performance of alternative architectures for PI-KAN-PointNet. Section \ref{Sect64} proposes a physics-informed PointNet combined with both MLPs and KANs, where one serves as the encoder and the other as the decoder, and vice versa. Finally, Section \ref{Sect7} summarizes the research findings and discusses potential future directions.

\section{Benchmark case: Natural convection in a square enclosure containing a cylinder}
\label{Sect2}

To assess the capabilities of physics-informed PointNet with KAN for solving inverse problems, we consider the benchmark test case of natural convection in a square enclosure with a cylinder, which leads to thermally driven flow. A physics-informed neural network we used by \citet{cai2021physics} to solve an inverse problem involving two-dimensional forced convection heat transfer (see Fig. 2 of \cite{cai2021physics}). Moreover, \citet{cai2021physics} utilized their proposed physics-informed neural network to reconstruct the thermal fields of steady and unsteady mixed convection heat transfer for flow past a circular cylinder. Additionally, \citet{wang2021reconstruction} applied a physics-informed neural network to solve an inverse problem in natural convection within a square enclosure containing a circular inner cylinder.

The governing equations for the conservation of mass, momentum, and energy for an incompressible, steady-state flow of a Newtonian fluid in two-dimensional space are given by:

\begin{equation}
\nabla \cdot \textbf{\textit{u}} = 0 \quad \text{in} \, V,
\label{Eq1}
\end{equation}

\begin{equation}
\rho \big(\textbf{\textit{u}} \cdot \nabla \big) \textbf{\textit{u}} - \mu \Delta \textbf{\textit{u}} + \nabla p = \textbf{\textit{f}} \quad \text{in} \, V,
\label{Eq2}
\end{equation}

\begin{equation}
\rho \big(\textbf{\textit{u}} \cdot \nabla \big) T - \frac{\kappa}{c_p} \Delta T = 0 \quad \text{in} \, V,
\label{Eq3}
\end{equation}
where $\textbf{\textit{u}}$ is the velocity vector with components $u$ and $v$ in the $x$ and $y$ directions, respectively. The variables $p$ and $T$ represent the pressure and temperature, respectively, while $\textbf{\textit{f}}$ denotes the external body force, with $f^x$ and $f^y$ corresponding to its components in the $x$ and $y$ directions. The fluid density is denoted by $\rho$, the dynamic viscosity by $\mu$, the thermal conductivity by $\kappa$, and the specific heat at constant pressure by $c_p$. Energy dissipation is neglected in Eq. \ref{Eq3}. The fluid domain $V$ is a non-simply connected region defined as:

\begin{equation}
V := H - W,
\label{Eq4}
\end{equation}
where $H$ represents the square enclosure and the inner cylinder is specified with the space $W$. The square domain has a side length $L$.

Consider a hot cylinder with a surface temperature $T_h$ situated within a cold square enclosure with a surface temperature $T_c$. The buoyancy force induces natural convection, with the solution to the energy equation (Eq. \ref{Eq3}) contributing to the source term of the momentum conservation equation (Eq. \ref{Eq2}) via the Boussinesq approximation \cite{lee2010natural}. Consequently, the velocity and temperature fields are coupled in the governing PDEs (Eqs. \ref{Eq1}--\ref{Eq3}). According to the Boussinesq approximation \cite{lee2010natural}, the forcing terms are given by:

\begin{equation}
\label{Eq5}
f^x = 0,
\end{equation}

\begin{equation}
\label{Eq6}
f^y = \rho G \beta (T - T_{\textrm{ref}}),
\end{equation}
where $G$ denotes the magnitude of gravitational acceleration, $\beta$ represents the thermal expansion coefficient, and $T_{\textrm{ref}}$ is the reference temperature. The Rayleigh number ($Ra$) is calculated as:

\begin{equation}
Ra = \frac{\rho^2 c_p G \beta (T_h - T_c) L^3}{\kappa \mu},
\label{Eq7}
\end{equation}
and the Prandtl number ($Pr$) is given by:

\begin{equation}
Pr = \frac{c_p \mu}{\kappa}.
\label{Eq8}
\end{equation}

We formulate the inverse problem as follows: given velocity boundary conditions on all boundaries, temperature boundary conditions solely on the outer hot surface, and limited measurements of velocity, temperature, and pressure fields at sensor locations, the goal is to reconstruct the complete velocity, temperature, and pressure fields at specific inquiry points, with a particular interest in determining the temperature distribution on the surface of the inner cylinder. Mathematically, this constitutes an ill-posed problem.

\section{Kolmogorov-Arnold network (KAN) layer}
\label{Sect3}

In this section, we explain the Kolmogorov-Arnold Network (KAN) layer framework. For clarity, let us consider a KAN with a single hidden layer. The input to the network is a vector $\mathbf{r}$ of size $d_\text{input}$, and the output is a vector $\mathbf{s}$ of size $d_\text{output}$. In this configuration, the relationship between the input and output in the one-layer KAN is expressed as:

\begin{equation}
    \mathbf{s}_{d_\text{output}\times 1} =  \mathbf{\Phi}_{d_\text{output}\times d_\text{input}} \mathbf{r}_{d_\text{input}\times 1},
    \label{Eq9}
\end{equation}
where the matrix $\mathbf{\Phi}_{d_\text{output}\times d_\text{input}}$ is structured as follows:

\begin{equation}
        \mathbf{\Phi}_{d_\text{output}\times d_\text{input}} = 
    \left[
\begin{array}{cccc}
\psi_{1,1}(\cdot) & \psi_{1,2}(\cdot) & \cdots & \psi_{1,d_\text{input}}(\cdot) \\
\psi_{2,1}(\cdot) & \psi_{2,2}(\cdot) & \cdots & \psi_{2,d_\text{input}}(\cdot) \\
\vdots & \vdots & \ddots & \vdots \\
\psi_{d_\text{output},1}(\cdot) & \psi_{d_\text{output},2}(\cdot) & \cdots & \psi_{d_\text{output},d_\text{input}}(\cdot) \\
\end{array}
\right],
\label{Eq10}
\end{equation}
where each element $\psi(z)$ is defined as:

\begin{equation}
\psi(z) = \sum_{i=0}^n \Lambda_i  P_i^{(\alpha,\beta)}(z),
\label{Eq11}
\end{equation}
with $P_i^{(\alpha,\beta)}(z)$ being the Jacobi polynomial of order $i$, $n$ representing the degree of the polynomial, and $\Lambda_i$ serving as trainable parameters. As a result, the total number of trainable parameters in this KAN layer is $(n+1) \times d_\text{output} \times d_\text{input}$. The Jacobi polynomials $P_n^{(\alpha,\beta)}(z)$ are computed using the following recursive formula \citep{Szego1939Orthogonal}:

\begin{equation}
    P_n^{(\alpha,\beta)}(z) = (A_n z + B_n)P_{n-1}^{(\alpha,\beta)}(z) + C_n P_{n-2}^{(\alpha,\beta)}(z),
    \label{Eq12}
\end{equation}
where the coefficients $A_n$, $B_n$, and $C_n$ are defined as:

\begin{equation}
     A_n = \frac{(2n+\alpha+\beta-1)(2n+\alpha+\beta)}{2n(n+\alpha+\beta)},
     \label{Eq13}
\end{equation}

\begin{equation}
    B_n = \frac{(2n+\alpha+\beta-1)(\alpha^2 - \beta^2)}{2n(n+\alpha+\beta)(2n+\alpha+\beta-2)},
    \label{Eq14}
\end{equation}

\begin{equation}
    C_n = \frac{-2(n+\alpha-1)(n+\beta-1)(2n+\alpha+\beta)}{2n(n+\alpha+\beta)(2n+\alpha+\beta-2)}.
    \label{Eq15}
\end{equation}

The initial conditions for the recursion are:

\begin{equation}
    P_0^{(\alpha,\beta)}(z) = 1,
     \label{Eq16}
\end{equation}

\begin{equation}
    P_1^{(\alpha,\beta)}(z) = \frac{1}{2}(\alpha+\beta+2)z + \frac{1}{2}(\alpha-\beta).
     \label{Eq17}
\end{equation}
Given this recursive formulation, the Jacobi polynomials $P_i^{(\alpha,\beta)}(z)$ for $0 \leq i \leq n$ are successively generated. Since Jacobi polynomials require inputs in the range $[-1, 1]$, the input vector $\mathbf{r}$ is scaled accordingly before entering the KAN layer. To achieve this, we apply the hyperbolic tangent function:

\begin{equation}
    \tanh(z) = \frac{e^{2z} - 1}{e^{2z} + 1}.
    \label{Eq18}
\end{equation}
This scaling strategy has been adopted in previous studies (e.g., \citep{shukla2024comprehensive,KANwithTANH,aghaei2024Fractional}). Lastly, when $\alpha = \beta = 0$, the Jacobi polynomials reduce to Legendre polynomials \citep{MiltonHandbook,Szego1939Orthogonal}. Chebyshev polynomials of the first and second kinds are obtained with $\alpha = \beta = -0.5$ and $\alpha = \beta = 0.5$, respectively, while Gegenbauer (ultraspherical) polynomials emerge when $\alpha = \beta$ \citep{Szego1939Orthogonal}.

Multiple KAN layers can be stacked to form a KAN component. In our implementation within PointNet (see Sect. \ref{Sect4}), we utilize shared KAN components. For the definition and detailed explanation of shared KAN layers, one might refer to \citep{kashefi2024KANpointnet}. A shared KAN component with two layers of sizes $\mathcal{B}_1$ and $\mathcal{B}_2$ is denoted as $(\mathcal{B}_1, \mathcal{B}_2)$, and a component with three layers is expressed as $(\mathcal{B}_1, \mathcal{B}_2, \mathcal{B}_3)$.


\begin{figure}[!htbp]
  \centering 
        \includegraphics[width=\textwidth]{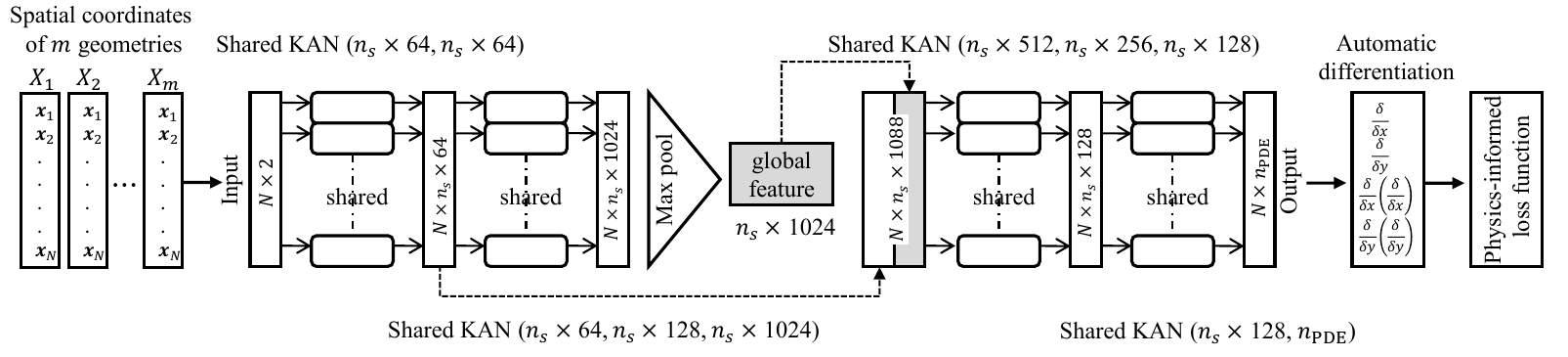}
  \caption{Architecture of the physics-informed Kolmogorov-Arnold PointNet (PI-KAN-PointNet). Shared KANs with the labels $(\mathcal{B}_1, \mathcal{B}_2)$ and $(\mathcal{B}_1, \mathcal{B}_2, \mathcal{B}_3)$ are explained in the text. $n_{\text{PDE}}$ denotes the number of variables in partial differential equations. $N$ is the number of points in the point clouds. $n_s$ is the global scaling parameter used to control the network size. PI-KAN-PointNet utilizes automatic differentiation to compute spatial derivatives (e.g., $\frac{\delta \Tilde{u}}{ \delta y}$, $\frac{\delta \Tilde{p}}{\delta x}$, $\frac{\delta}{\delta y}\left(\frac{\delta \Tilde{v}}{\delta y}\right)$, etc.), which are then used to formulate the physics-informed loss function (see Eq.\ref{Eq30}).}
  \label{Fig1}
\end{figure}

\section{Architecture of physics-informed PointNet with KAN layers}
\label{Sect4}


The main objective of PI-KAN-PointNet is to obtain the solution of inverse problems over the $m$ number of domains with irregular geometries, simultaneously. Figure \ref{Fig1} illustrates the architecture of PI-KAN-PointNet. In a set containing $m$ domain, each domain $V_i$ ($1 \leq i \leq m$) is represented by a point cloud $\mathcal{X}_i$ with $N$ points, where $\mathcal{X}_i = \left\{ \boldsymbol{x}_j \in \mathbb{R}^d \right\}_{j=1}^{N}$. The spatial dimension of $V_i$, and thus of $\mathcal{X}_i$, is denoted by $d$. In this study, we focus on two-dimensional problems, setting $d = 2$. Therefore, $\boldsymbol{x}_j$ ($1 \leq j \leq N$) represents the spatial coordinates of each point in the point cloud $\mathcal{X}_i$. We denote the $x$ and $y$ components of $\boldsymbol{x}_j$ as $x_j$ and $y_j$, respectively. In PI-KAN-PointNet, an end-to-end mapping is provided from $\mathcal{X}_i$ to $\mathcal{Y}_i$, where $\mathcal{Y}_i = \left\{ \boldsymbol{y}_j \in \mathbb{R}^{n_{\text{PDE}}} \right\}_{j=1}^{N}$. The variable $n_\text{PDE}$ indicates the number of fields we aim to find. In this study, we focus on predicting the two-dimensional velocity vector, pressure and temperature fields, making $n_\text{PDE} = 4$. Therefore, $\boldsymbol{y}_j$ is the vector of predicted fields at the spatial point $\boldsymbol{x}_j$, comprising the components $u_j$, $v_j$, $p_j$, and $T_j$. Mathematically, this process is described as:

\begin{equation}
\left(u_j, v_j, p_j, T_j\right) = f \left(\left(x_j, y_j\right), g\left(\mathcal{X}_i\right)\right); \quad \forall \left(x_j, y_j\right) \in \mathcal{X}_i \text{ and } \forall \left(u_j, v_j, p_j, T_j\right) \in \mathcal{Y}_i \text{ with } 1 \leq i \leq m \text{ and } 1 \leq j \leq N,
\label{Eq19}
\end{equation}
where $f$ represents the mapping function in PI-KAN-PointNet.

KA-PointNet is designed to be invariant to any permutation of the input vector $\mathcal{X}_i$, meaning that if the points in $\mathcal{X}_i$ are permuted, the geometric structure remains the same, and hence the solution $\mathcal{Y}_i$ should also remain unchanged. This permutation invariance is achieved using a symmetric function in conjunction with shared KANs. The function $g$ serves as a symmetric encoder of the geometric features of the point cloud $\mathcal{X}_i$. Following the approach introduced in Ref. \cite{qi2017pointnet}, we define $g$ as the maximum function:

\begin{equation}
g\left(\mathcal{X}_i\right) = \max \left(h\left(x_1, y_1\right), \ldots, h\left(x_N, y_N\right)\right); \quad \forall \left(x_j, y_j\right) \in \mathcal{X}_i \text{ with } 1 \leq i \leq m \text{ and } 1 \leq j \leq N,
\label{Eq20}
\end{equation}
where $h$ represents two shared KAN layers in the first branch of PI-KAN-PointNet (see Fig. \ref{Fig3}). In the context of PI-KAN-PointNet, $g\left(\mathcal{X}_i\right)$ is referred to as the global feature. The key concept behind PI-KAN-PointNet, as a geometric deep learning model, is that the predicted fields at each spatial point depend not only on the spatial coordinates of that point but also on the overall geometric structure of the domain, formed by all points, including the specific point. This idea is evident in Eqs. (\ref{Eq19})--(\ref{Eq20}).


The batch size, $B$, represents the number of point clouds (i.e., $\mathcal{X}_i$ and $\mathcal{Y}_i$ pairs) processed by KA-PointNet at each epoch. As illustrated in Fig. \ref{Fig3}, the input to KA-PointNet is a three-dimensional tensor of size $B \times N \times 2$. Following this, two sequential shared KANs with sizes (64, 64) and (64, 128, 1024) are applied, as shown in Fig. \ref{Fig3}. The global feature, of size 1024, is generated by applying the maximum function. This global feature is then concatenated with an intermediate feature tensor of size $B \times N \times 64$, resulting in a tensor of size $B \times N \times 1088$. Next, two additional shared KANs are applied, with sizes (512, 256, 128) and (128, $n_\text{PDE}$), respectively, yielding a tensor of size $B \times N \times n_\text{PDE}$, as depicted in Fig. \ref{Fig3}. After each KAN layer (except the last layer in the network), batch normalization \citep{ioffe2015batch} is applied to ensure stable training and avoid divergence of the training loss.



For each $V_i$ ($1 \leq i \leq m$) and its associated pair $\mathcal{X}_i$ and $\mathcal{Y}_i$, the residuals of the continuity equation $\big(r_i^{\text{continuity}}\big)$, momentum conservation equations in the $x-$ direction $\big(r_i^{\text{momentum}_x}\big)$ and in the $y-$directions $\big(r_i^{\text{momentum}_y}\big)$, energy conservation equation $\big(r_i^{\text{energy}}\big)$, along with the residuals of the Dirichlet boundary conditions of the velocity $\big(r_i^{\text{velocity}_{\text{BC}}}\big)$, and temperature $\big(r_i^{\text{temperature}_{\text{outer-BC}}}\big)$ and sparse observations of the velocity $\big(r_i^{\text{velocity}_{\text{obs}}}\big)$, pressure $\big(r_i^{\text{pressure}_{\text{obs}}}\big)$, and temperature $\big(r_i^{\text{temperature}_{\text{obs}}}\big)$ are respectively defined as follows:
\begin{equation}
r_i^{\text{continuity}} = \frac{1}{M_1} \sum_{k=1}^{M_1} \left (\frac{\delta \Tilde{u}_k}{\delta x_k} + \frac{\delta \Tilde{v}_k}{\delta y_k}  \right)^2,
\label{Eq21}
\end{equation}
\begin{equation}
r_i^{\text{momentum}_x} = \frac{1}{M_1} \sum_{k=1}^{M_1} \left ( \rho \left(\Tilde{u}_k \frac{\delta \Tilde{u}_k}{\delta x_k} + \Tilde{v}_k \frac{\delta \Tilde{u}_k}{\delta y_k} \right) +  \frac{\delta \Tilde{p}_k}{\delta x_k}-\mu \left(\frac{\delta}{\delta x_k} \left(\frac{\delta \Tilde{u}_k}{\delta x_k} \right) + \frac{\delta}{\delta y_k}\left(\frac{\delta \Tilde{u}_k}{\delta y_k} \right) \right)  - f_k^x \right)^2,
\label{Eq22}
\end{equation}
\begin{equation}
r_i^{\text{momentum}_y} = \frac{1}{M_1} \sum_{k=1}^{M_1} \left ( \rho \left(\Tilde{u}_k \frac{\delta \Tilde{v}_k}{\delta x_k} + \Tilde{v}_k \frac{\delta \Tilde{v}_k}{\delta y_k} \right) +  \frac{\delta \Tilde{p}_k}{\delta y_k}-\mu \left(\frac{\delta}{\delta x_k} \left(\frac{\delta \Tilde{v}_k}{\delta x_k} \right) + \frac{\delta}{\delta y_k}\left(\frac{\delta \Tilde{v}_k}{\delta y_k} \right) \right)  - f_k^y \right)^2,
\label{Eq23}
\end{equation}
\begin{equation}
r_i^{\text{energy}} = \frac{1}{M_1} \sum_{k=1}^{M_1} \left ( \rho \left(\Tilde{u}_k \frac{\delta \widetilde{T}_k}{\delta x_k} + \Tilde{v}_k \frac{\delta \widetilde{T}_k}{\delta y_k} \right) -\frac{\kappa}{c_p} \left(\frac{\delta}{\delta x_k} \left(\frac{\delta \widetilde{T}_k}{\delta x_k} \right) + \frac{\delta}{\delta y_k}\left(\frac{\delta \widetilde{T}_k}{\delta y_k} \right) \right) \right)^2,
\label{Eq24}
\end{equation}
\begin{equation}
r_i^{\text{velocity}_{\text{BC}}} = \frac{1}{M_2} \sum_{k=1}^{M_2} \left (\left(\Tilde{u}_k - u_k \right)^2 + \left(\Tilde{v}_k - v_k \right)^2  \right),
\label{Eq25}
\end{equation}
\begin{equation}
r_i^{\text{temperature}_{\text{outer-BC}}} = \frac{1}{M_3} \sum_{k=1}^{M_3} \left (\widetilde{T}_k - T_k \right)^2,
\label{Eq26}
\end{equation}
\begin{equation}
r_i^{\text{velocity}_{\text{obs}}} = \frac{1}{M_4} \sum_{k=1}^{M_4} \left (\left(\Tilde{u}_k - u_k \right)^2 + \left(\Tilde{v}_k - v_k \right)^2  \right),
\label{Eq27}
\end{equation}
\begin{equation}
r_i^{\text{pressure}_{\text{obs}}} = \frac{1}{M_5} \sum_{k=1}^{M_5} \left (\Tilde{p}_k - p_k \right)^2,
\label{Eq28}
\end{equation}
\begin{equation}
r_i^{\text{temperature}_{\text{obs}}} = \frac{1}{M_5} \sum_{k=1}^{M_5} \left (\widetilde{T}_k - T_k \right)^2,
\label{Eq29}
\end{equation}
where $M_1$ is the number of interior points of $\mathcal{X}_i$. $M_2$ denotes the number of points located on the inner and outer boundaries of $\mathcal{X}_i$, while $M_3$ exclusively indicates the number of points placed on the outer boundaries of $\mathcal{X}_i$. $M_4$ is the number of sensors located in the computational domain to measure the velocity values, sparsely. $M_5$ is similarly defined for the temperature and pressure. Obviously, $M_1+M_2=N$. In this study, we use a fixed value for $M_1$, $M_2$, $M_3$, $M_4$, and $M_5$ and over all $\mathcal{X}_i$. Note that while $M_1$, $M_2$, $M_3$, $M_4$, and $M_5$ could conceptually vary from one point cloud to another, $N$ has to be fixed over all $\mathcal{X}_i$. The solutions obtained by PI-KAN-PointNet are shown by $(\Tilde{u},\Tilde{v},\Tilde{p},\widetilde{T})$, whereas the ground truth obtained by a numerical solver, lab experiment, or analytical solutions are indicated by $(u,v,p,T)$. The PI-KAN-PointNet loss function is specified as a combination of the residuals introduced above summed over all $V_i$ ($1 \leq i \leq m$). We indicate the automatic differentiation operator by $\delta$ computed using the PyTorch software \cite{paszke2019pytorch}. In line with the definition of the inverse problem in Sect. \ref{Sect2}, the corresponding loss function ($\mathcal{L}$) is expressed as:

\begin{equation}
\label{Eq30}
\begin{split}
\mathcal{L} = \frac{1}{m} \sum_{i=1}^m \big(r_i^{\text{continuity}} +  r_i^{\text{momentum}_x} + r_i^{\text{momentum}_y} + r_i^{\text{velocity}_{\text{BC}}} +  r_i^{\text{velocity}_{\text{obs}}} +  r_i^{\text{pressure}_{\text{obs}}} + r_i^{\text{energy}} \\
+  r_i^{\text{temperature}_{\text{outer-BC}}} + r_i^{\text{temperature}_{\text{obs}}}\big).
\end{split}
\end{equation}
Note that no boundary condition is specified for the pressure in the loss function (Eq. \ref{Eq30}).

\section{Computational setup}
\label{Sect5}

We define the square enclosure as $H:= [-1 \text{ m}, 1 \text{ m}] \times [-1 \text{ m}, 1 \text{ m}]$ with the side length $L$ of 2 m. Zero-velocity Dirichlet conditions are applied at all boundaries. We set the density ($\rho$), thermal expansion coefficient ($\beta$), gravitational acceleration ($G$), specific heat ($c_p$), and hot temperature ($T_h$) to 1.00, while the cold temperature ($T_c$) and reference temperature ($T_{\textrm{ref}}$) are set to 0.00. The dynamic viscosity ($\mu$) and thermal conductivity ($\kappa$) are both set to $2 \sqrt{2} \times 10^{-2.5}$. All units used in this study are expressed in the International System of Units (SI). This parameter set results in a Rayleigh number of $Ra = 10^5$ and a Prandtl number of $Pr = 1.0$. We consider three different shapes for $W$: equilateral heptagon, equilateral octagon, and equilateral nonagon. To expand the range of geometries, we also rotate the inner cylinders. Details of the generated geometries are provided in Table \ref{Table1}. A total of 135 geometries are generated. For the point clouds $V_i$ ($1 \leq i \leq m$), we set $N = 5000$, $M_1 = 4340$, $M_2 = 660$, $M_3 = 492$, $M_4 = 105$, and $M_5 = 130$. A similar data set was used by \citet{kashefi2022physics}.


Due to computational constraints, we were limited to a maximum batch size of 7. Consequently, we set the batch size parameter, $\mathcal{B}$, to 7 for this study. The Adam optimizer \cite{kingma2014adam} is employed with hyperparameters set to $\beta_1=0.9$, $\beta_2=0.999$, and $\hat{\epsilon}=10^{-8}$. For a comprehensive understanding of the mathematical definitions of $\beta_1$, $\beta_2$, and $\hat{\epsilon}$, one may refer to Ref. \cite{kingma2014adam}. We use a constant learning rate of 0.0005 and iterate for 2500 epochs. We report the results at the iteration where the minimum loss is obtained during these 2500 epochs. From an engineering perspective, this is reasonable because iterations cannot continue indefinitely due to limited computational resources.



\begin{figure}[!htbp]
  \centering 
      \begin{subfigure}[b]{0.95\textwidth}
        \centering
        \includegraphics[width=\textwidth]{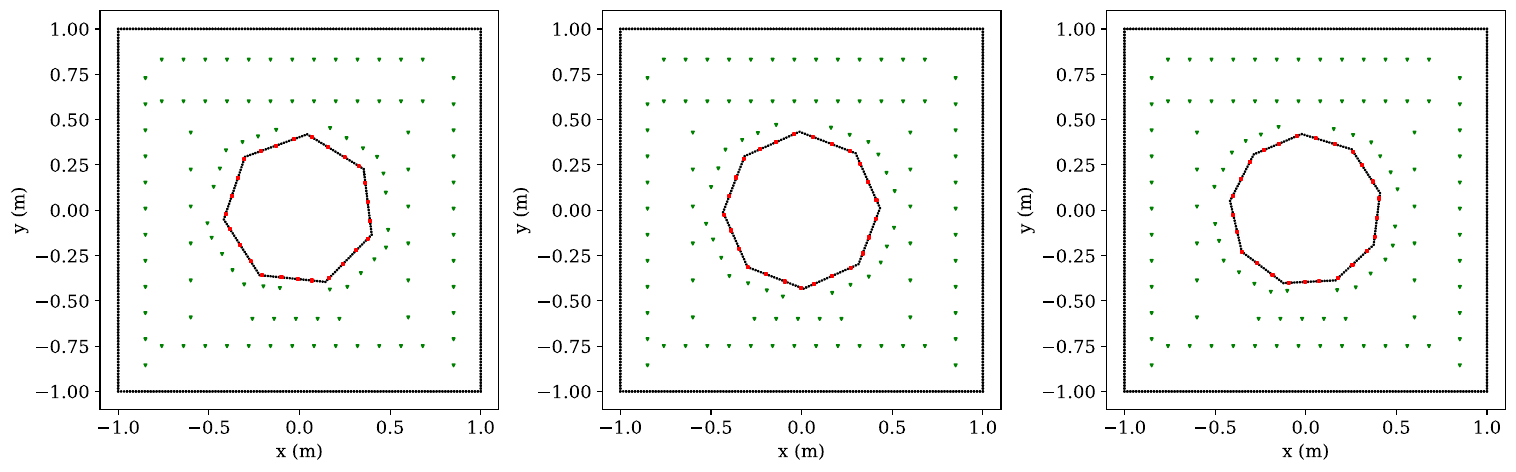}
    \end{subfigure}
    
      \caption{Examples of sensor locations for three domains. Green triangles represent sensors measuring velocity, pressure, and temperature, whereas red squares represent sensors measuring only pressure and temperature.}
  \label{FigSensor}
\end{figure}

We deploy 105 sensors inside the domains and 25 sensors on the surface of the inner cylinders, as depicted in Fig. \ref{FigSensor}. Sensors inside the domains (shown as green triangles in Fig. \ref{FigSensor}) measure velocity, pressure, and temperature, while sensors on the inner surfaces (shown as red squares in Fig. \ref{FigSensor}) measure only pressure and temperature. The placement of 25 sensors on the inner cylinder surface and 25 sensors surrounding the cylinder is slightly adjusted according to the cross-sectional shape, while the remaining 80 sensors are fixed across all the geometries.

For generating sparse data at virtual sensor locations and validating the results obtained by PI-KAN-PointNet, we use one of our high-fidelity finite element numerical solvers. This solver has been previously employed in the literature \cite{kashefi2018finite, kashefi2020coarse, kashefi2020coarseb, kashefi2020coarseC, kashefi2021coarseE}. Note that the point clouds used are a subset of the grid vertices from finite element meshes employed by the numerical solver.


\begin{table}[h]
\caption{Overview of the 135 generated geometries ($m=135$). The symbol $\Omega$ denotes the counterclockwise rigid rotation of space $W$ around its geometric center.}
\centering
\begin{tabular}{l l l l}
\toprule
Shape of $W$ & Side length & $\Omega$ (variation in orientation) & Number of data \\
\midrule
Equilateral nonagon & $0.365 \times \sin{\frac{\pi}{9}} \times \csc \frac{\pi}{7}  \textrm{ m}$ & $1^\circ,2^\circ,\cdots,39^\circ,40^\circ$ & $40$ \\
Equilateral octagon & $0.8(\sqrt{2}-1)  \textrm{ m}$ & $1^\circ,2^\circ,\cdots,44^\circ,45^\circ$ & $45$ \\
Equilateral heptagon & $0.365 \textrm{ m}$ & $1^\circ,2^\circ,\cdots,49^\circ,50^\circ$ & $50$ \\
\bottomrule
\end{tabular}
\label{Table1}
\end{table}


\section{Results and discussion}
\label{Sect6}

\begin{figure}[!htbp]
  \centering 
      \begin{subfigure}[b]{0.24\textwidth}
        \centering
        \includegraphics[width=\textwidth]{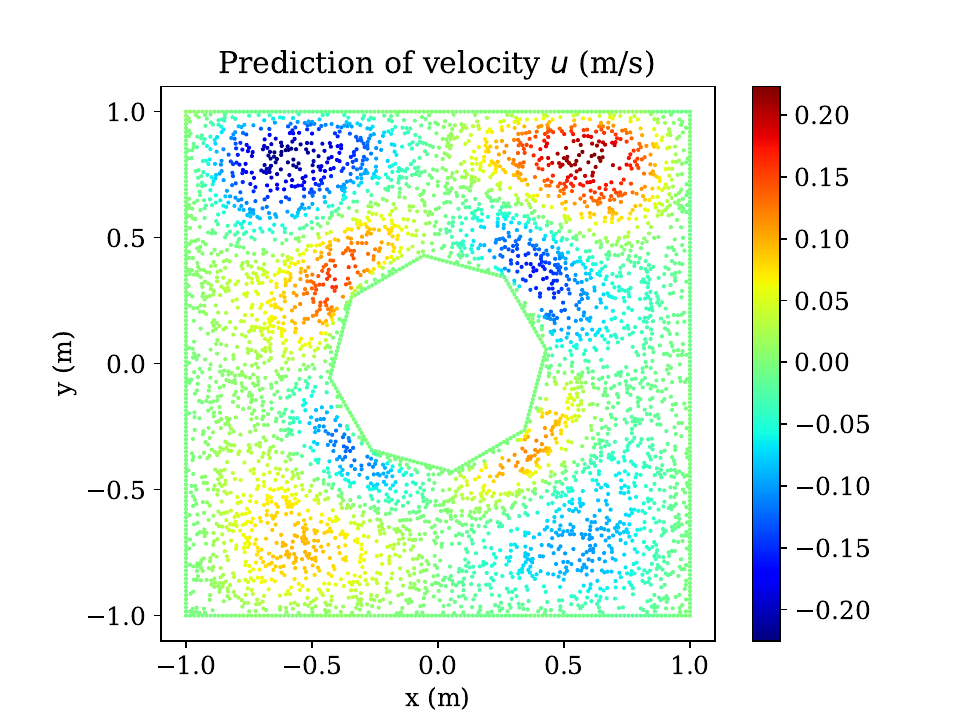}
    \end{subfigure}
    \begin{subfigure}[b]{0.24\textwidth}
        \centering
        \includegraphics[width=\textwidth]{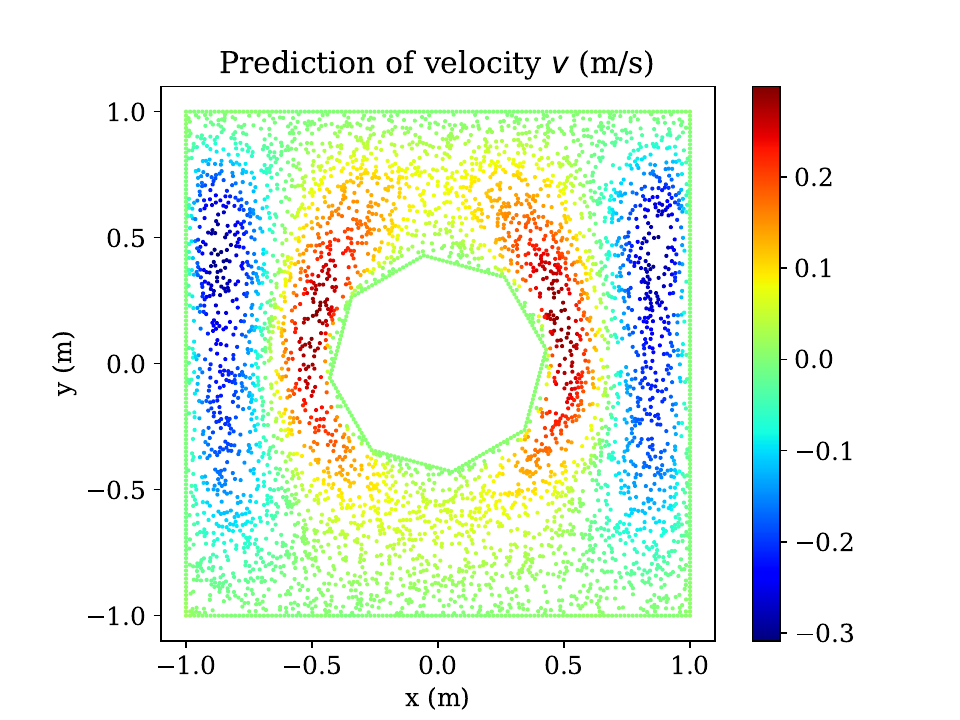}
    \end{subfigure}
    \begin{subfigure}[b]{0.24\textwidth}
        \centering
        \includegraphics[width=\textwidth]{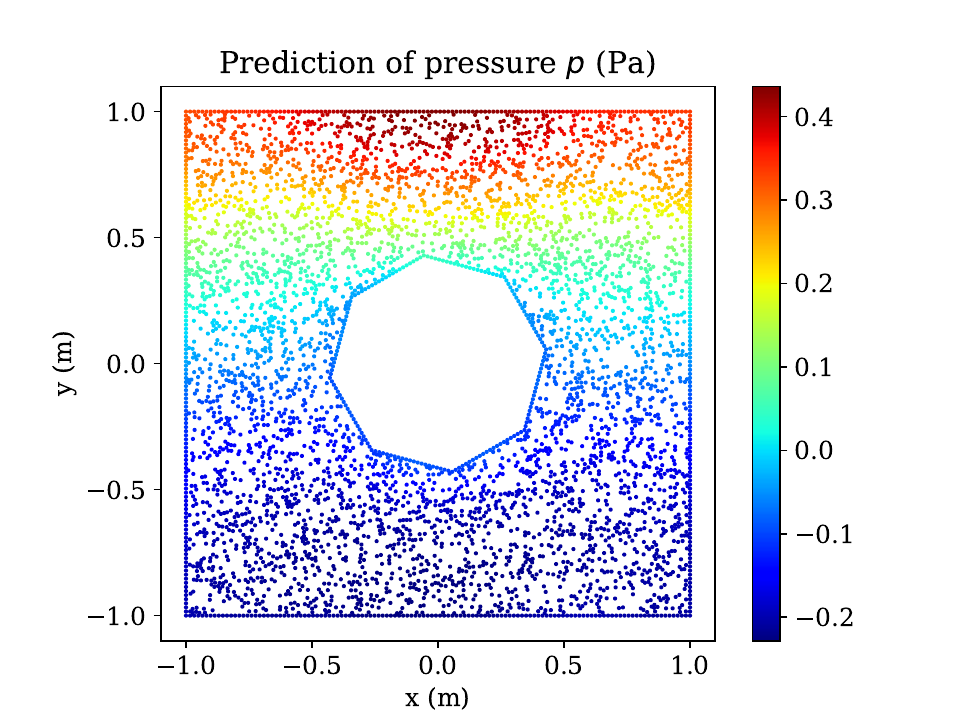}
    \end{subfigure}
     \begin{subfigure}[b]{0.24\textwidth}
        \centering
        \includegraphics[width=\textwidth]{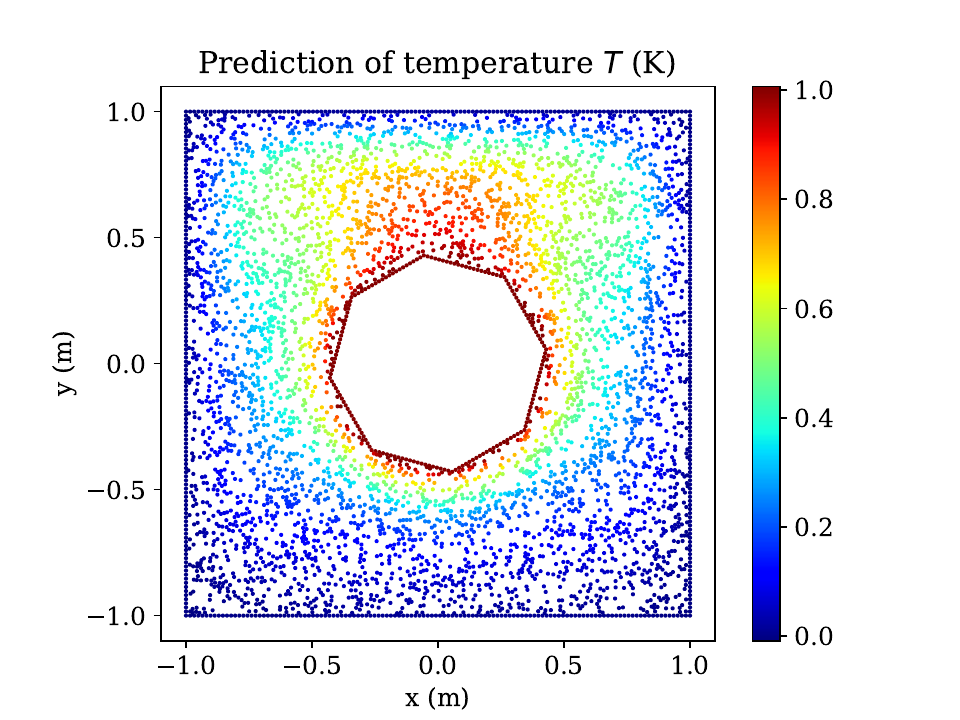}
    \end{subfigure}


  \centering 
      \begin{subfigure}[b]{0.24\textwidth}
        \centering
        \includegraphics[width=\textwidth]{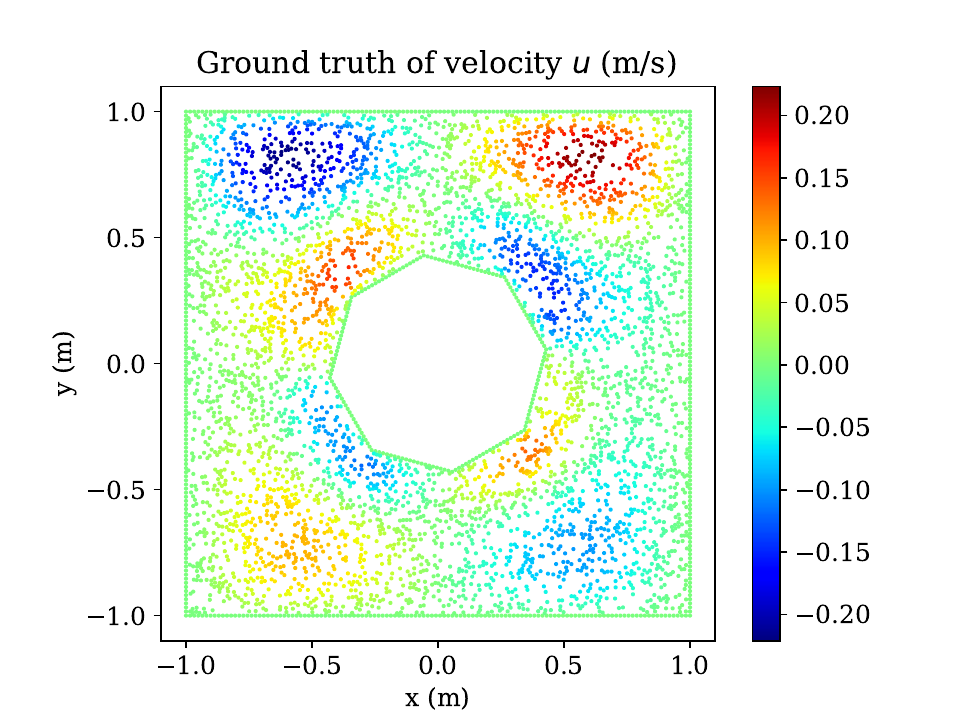}
    \end{subfigure}
    \begin{subfigure}[b]{0.24\textwidth}
        \centering
        \includegraphics[width=\textwidth]{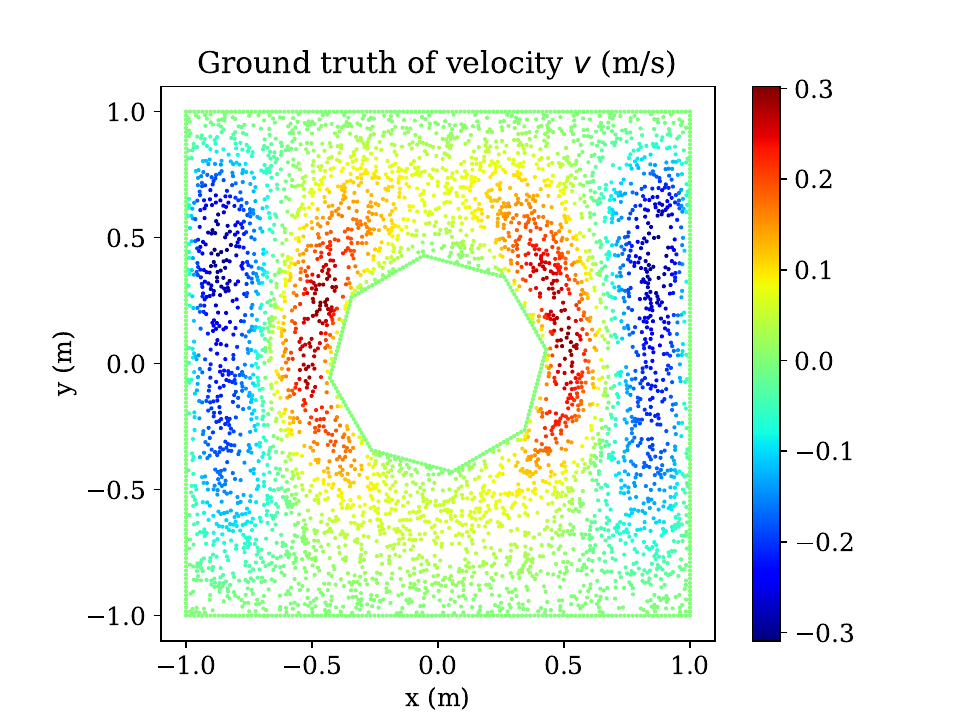}
    \end{subfigure}
    \begin{subfigure}[b]{0.24\textwidth}
        \centering
        \includegraphics[width=\textwidth]{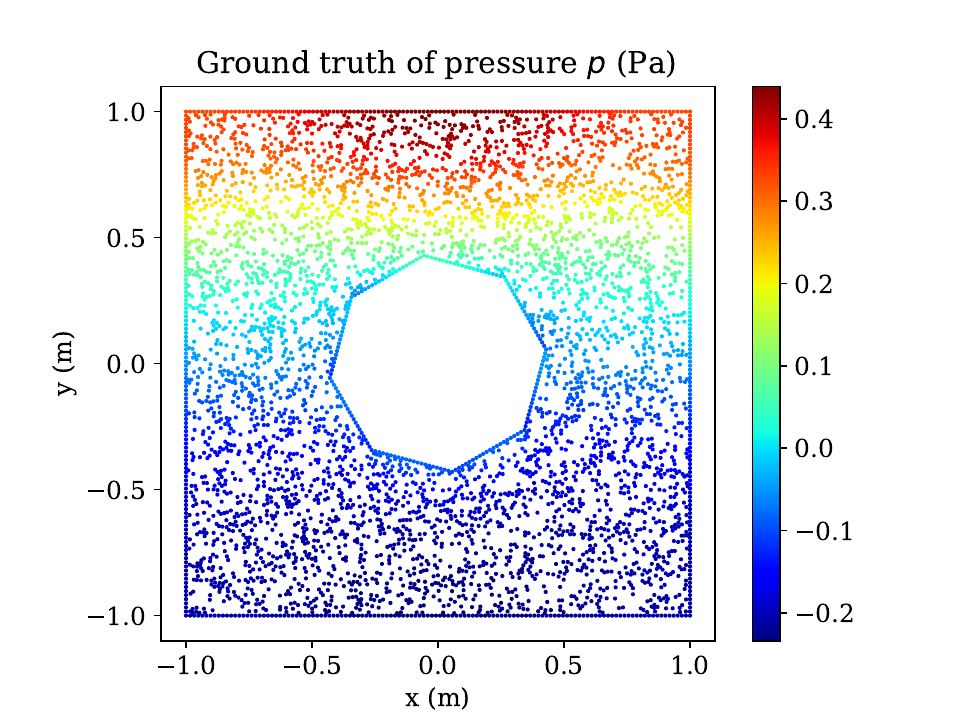}
    \end{subfigure}
     \begin{subfigure}[b]{0.24\textwidth}
        \centering
        \includegraphics[width=\textwidth]{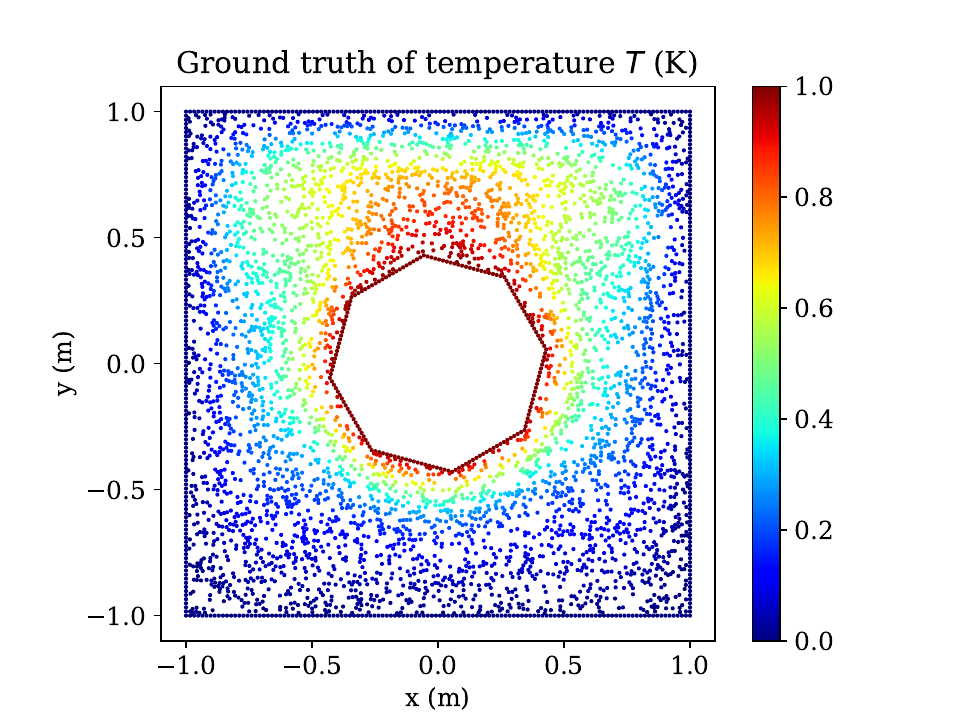}
    \end{subfigure}
    

  \centering 
      \begin{subfigure}[b]{0.24\textwidth}
        \centering
        \includegraphics[width=\textwidth]{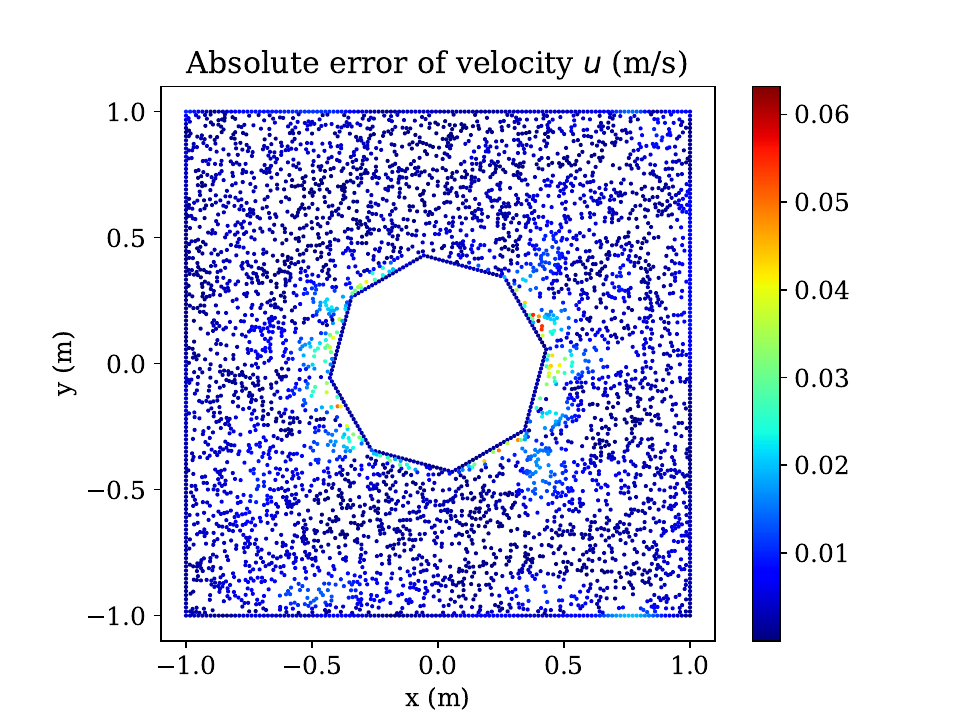}
    \end{subfigure}
    \begin{subfigure}[b]{0.24\textwidth}
        \centering
        \includegraphics[width=\textwidth]{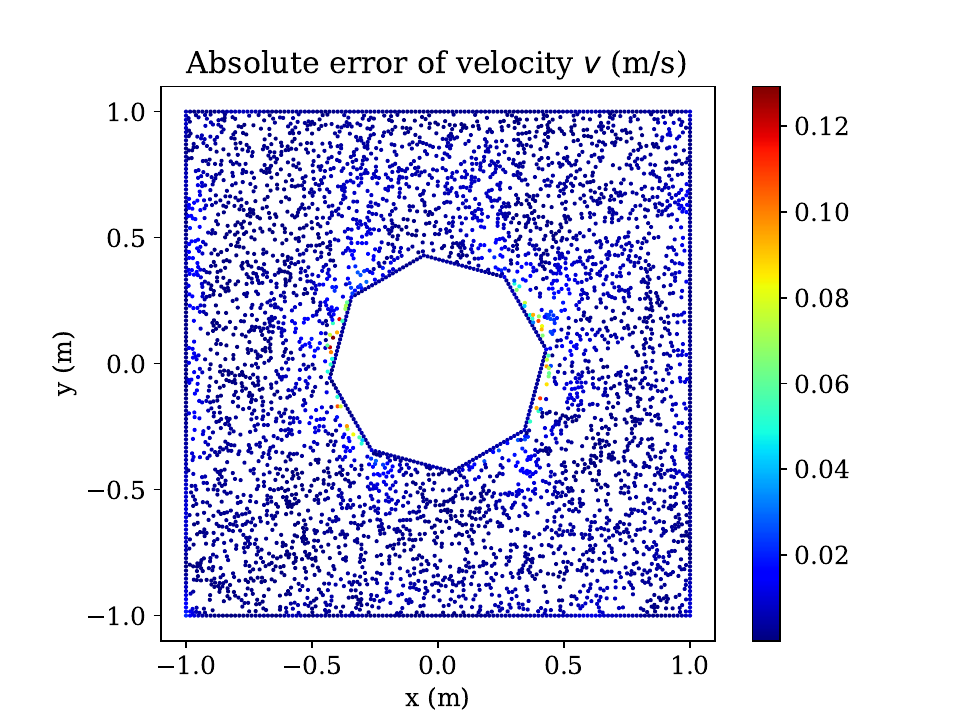}
    \end{subfigure}
    \begin{subfigure}[b]{0.24\textwidth}
        \centering
        \includegraphics[width=\textwidth]{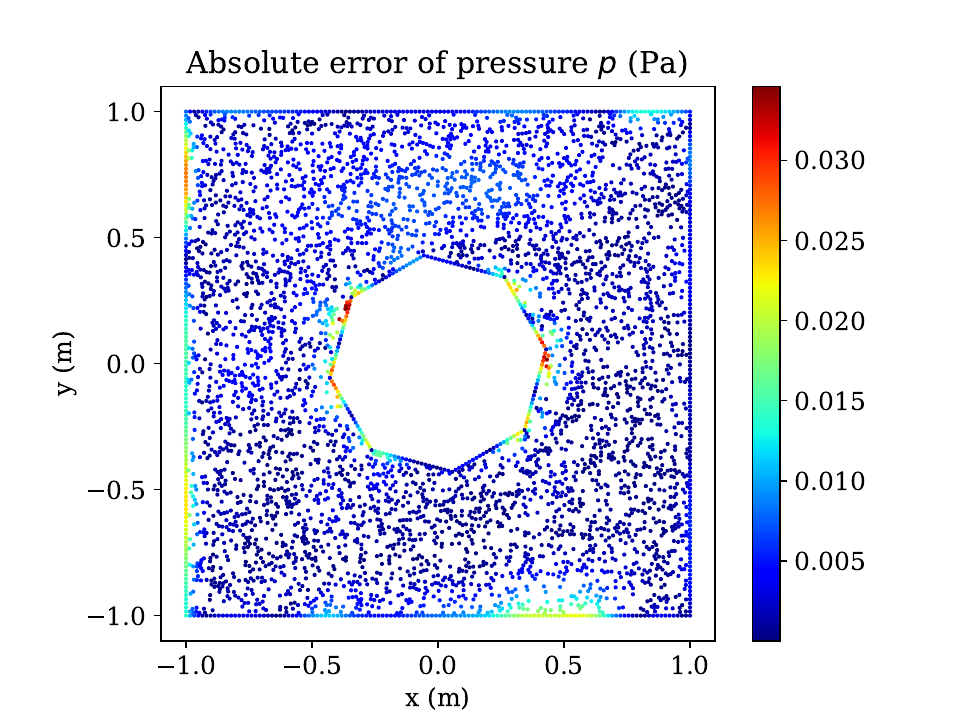}
    \end{subfigure}
     \begin{subfigure}[b]{0.24\textwidth}
        \centering
        \includegraphics[width=\textwidth]{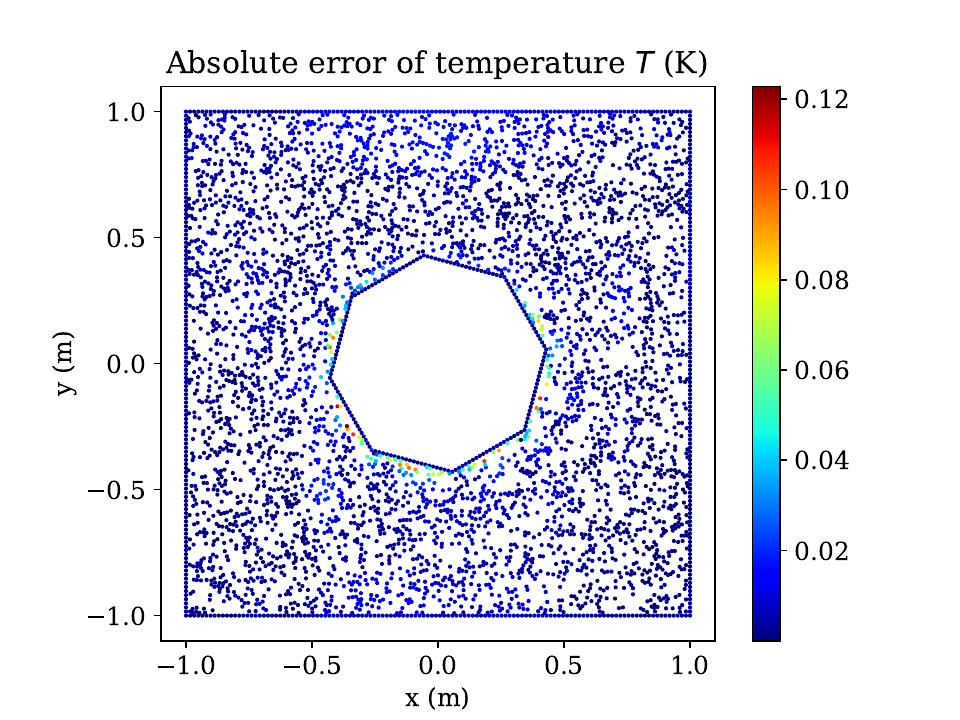}
    \end{subfigure}

  \caption{The first set of examples comparing the ground truth and predictions of the physics-informed KAN PointNet for the velocity, pressure, and temperature fields. The Jacobi polynomial used has a degree of 2, with $\alpha = \beta = -0.5$. Here, $n_s = 0.5$.}
  \label{Fig2}
\end{figure}


\begin{figure}[!htbp]
  \centering 
      \begin{subfigure}[b]{0.24\textwidth}
        \centering
        \includegraphics[width=\textwidth]{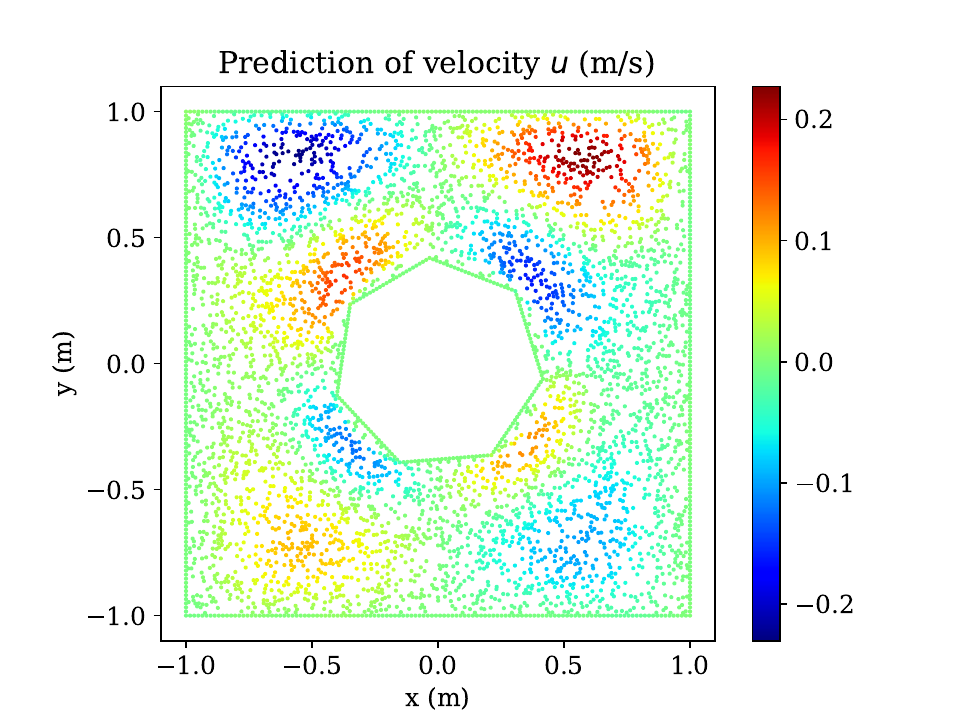}
    \end{subfigure}
    \begin{subfigure}[b]{0.24\textwidth}
        \centering
        \includegraphics[width=\textwidth]{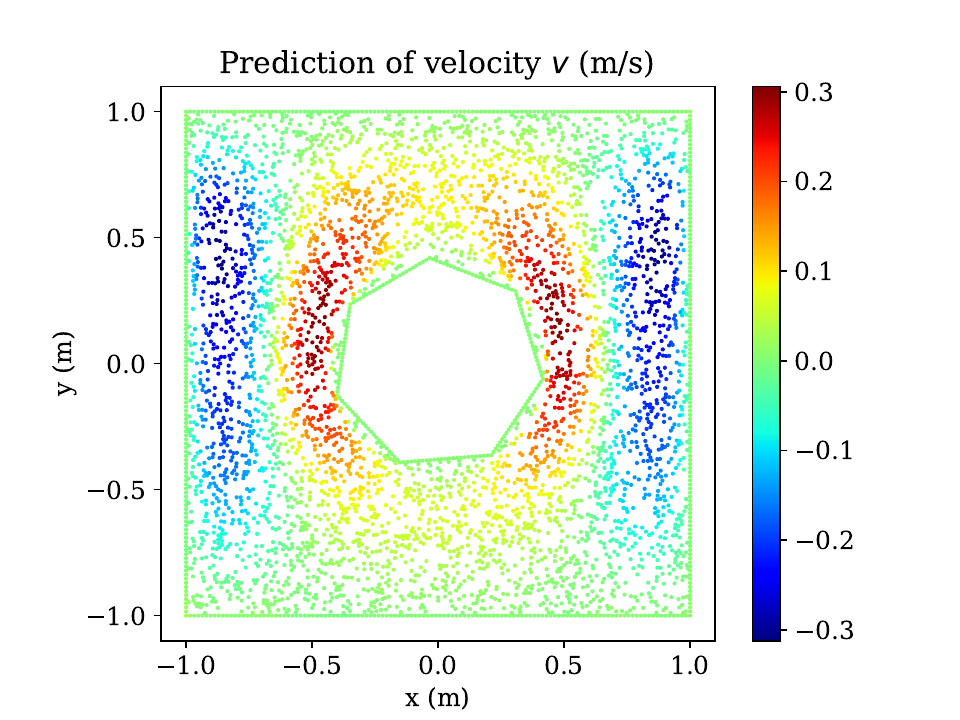}
    \end{subfigure}
    \begin{subfigure}[b]{0.24\textwidth}
        \centering
        \includegraphics[width=\textwidth]{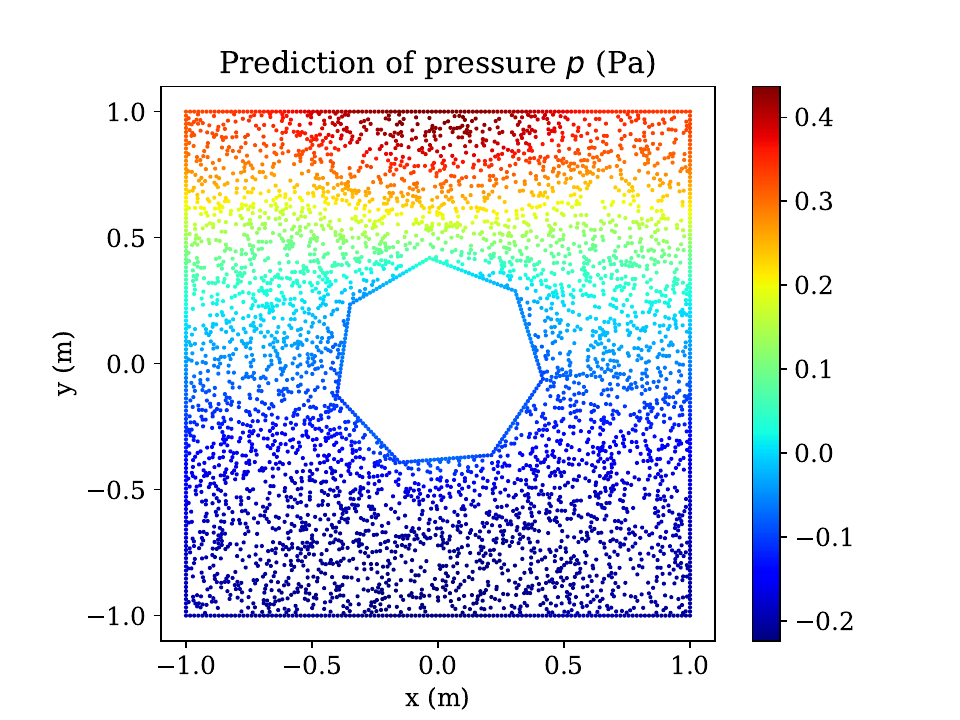}
    \end{subfigure}
     \begin{subfigure}[b]{0.24\textwidth}
        \centering
        \includegraphics[width=\textwidth]{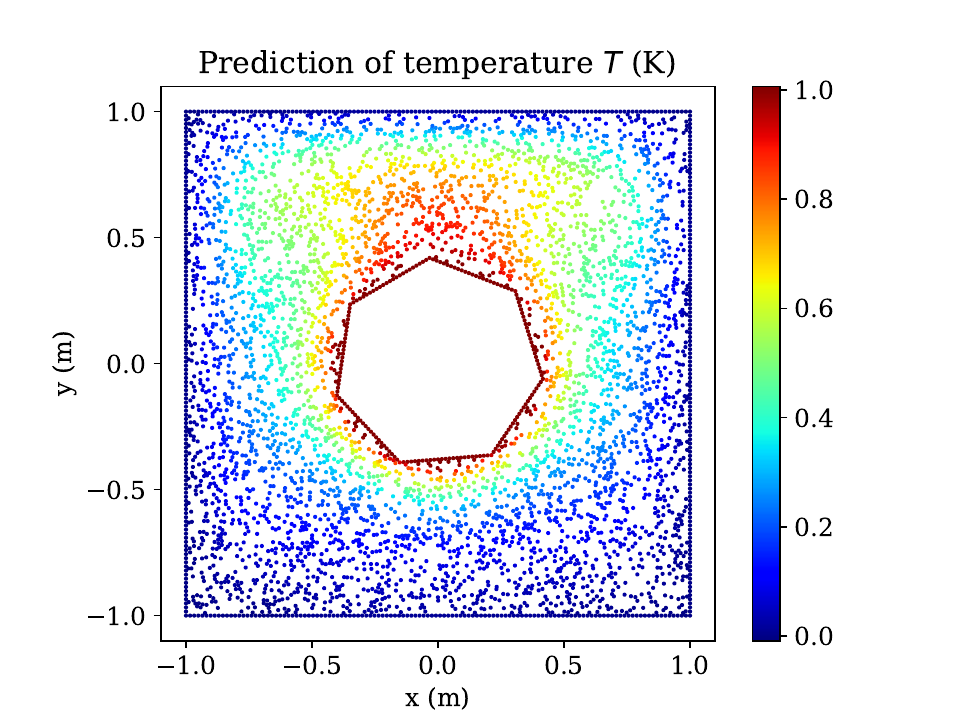}
    \end{subfigure}


  \centering 
      \begin{subfigure}[b]{0.24\textwidth}
        \centering
        \includegraphics[width=\textwidth]{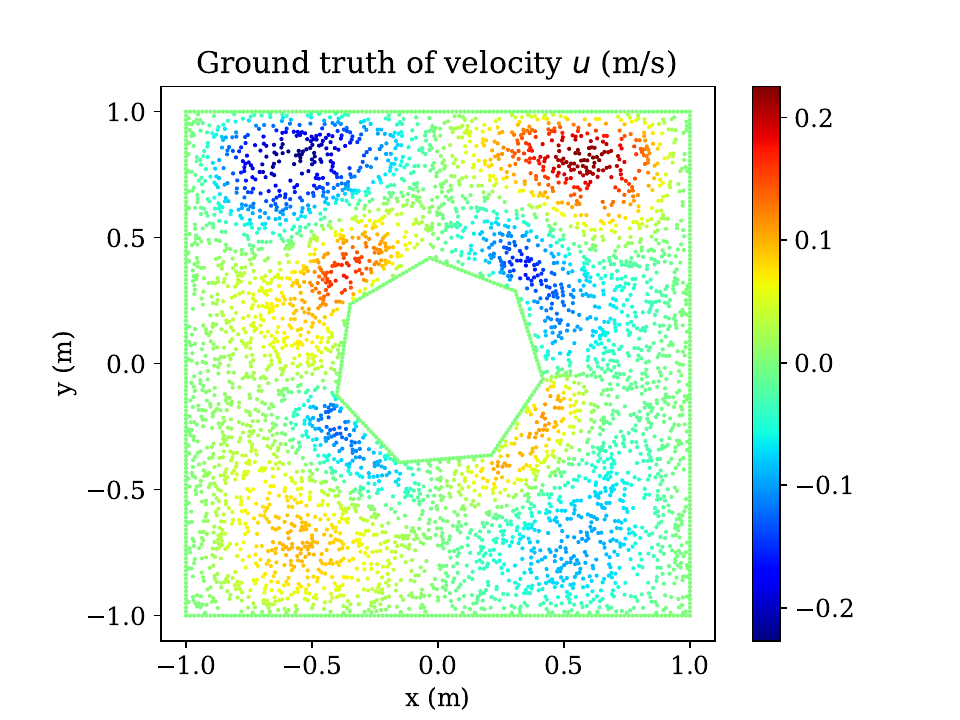}
    \end{subfigure}
    \begin{subfigure}[b]{0.24\textwidth}
        \centering
        \includegraphics[width=\textwidth]{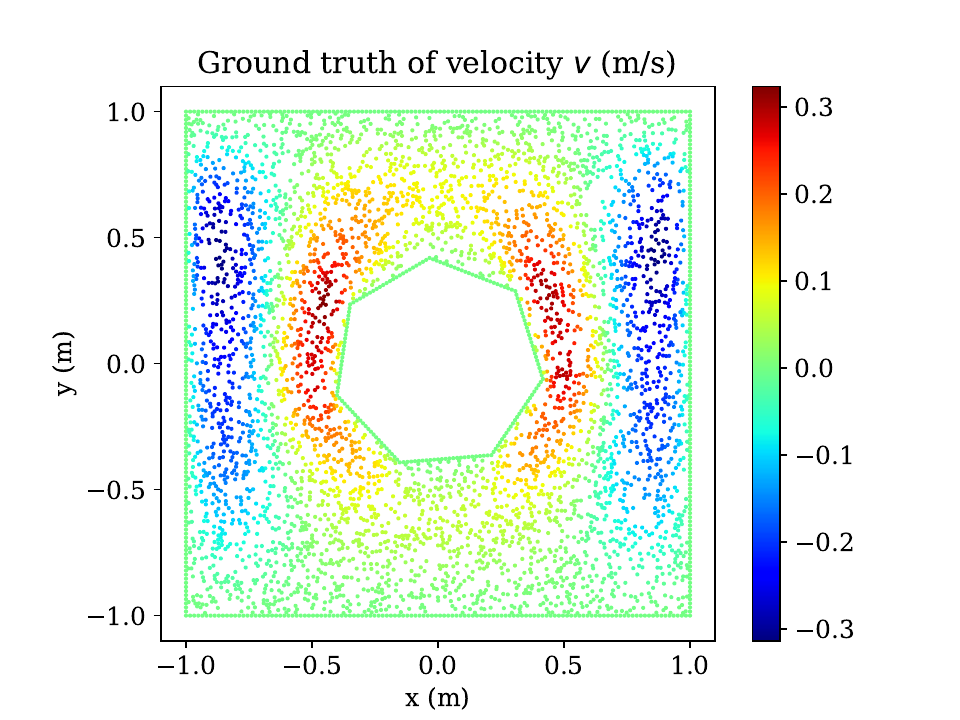}
    \end{subfigure}
    \begin{subfigure}[b]{0.24\textwidth}
        \centering
        \includegraphics[width=\textwidth]{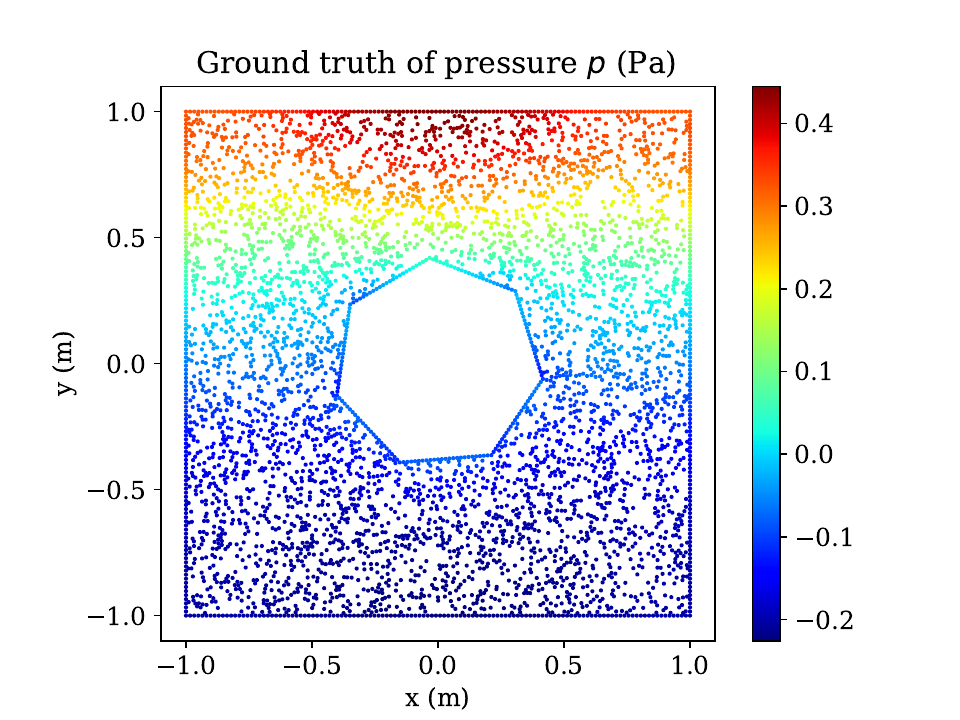}
    \end{subfigure}
     \begin{subfigure}[b]{0.24\textwidth}
        \centering
        \includegraphics[width=\textwidth]{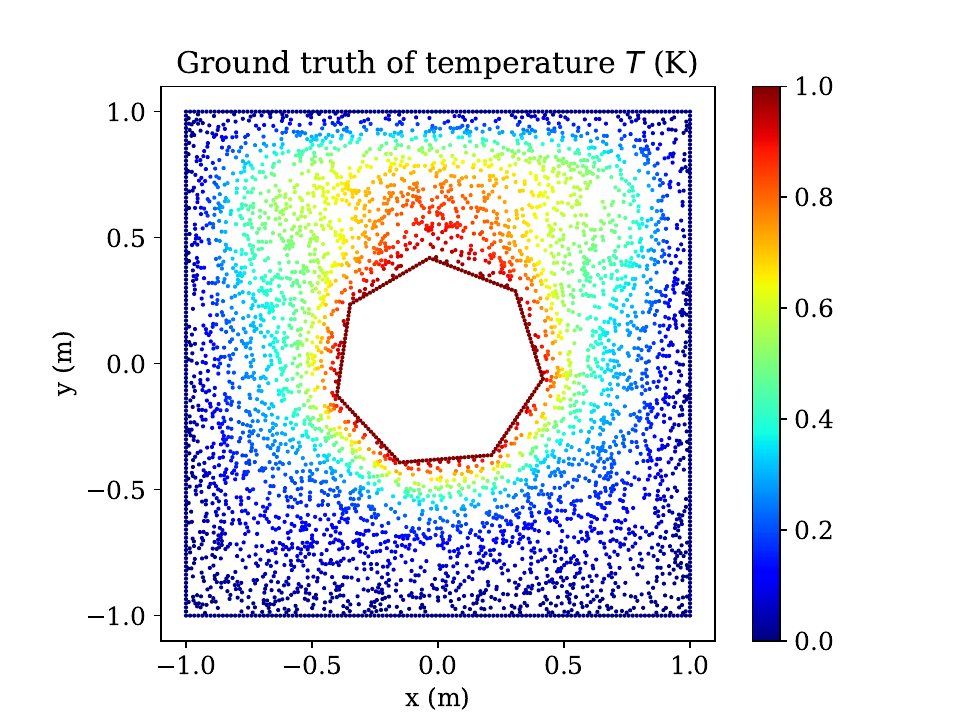}
    \end{subfigure}
    

  \centering 
      \begin{subfigure}[b]{0.24\textwidth}
        \centering
        \includegraphics[width=\textwidth]{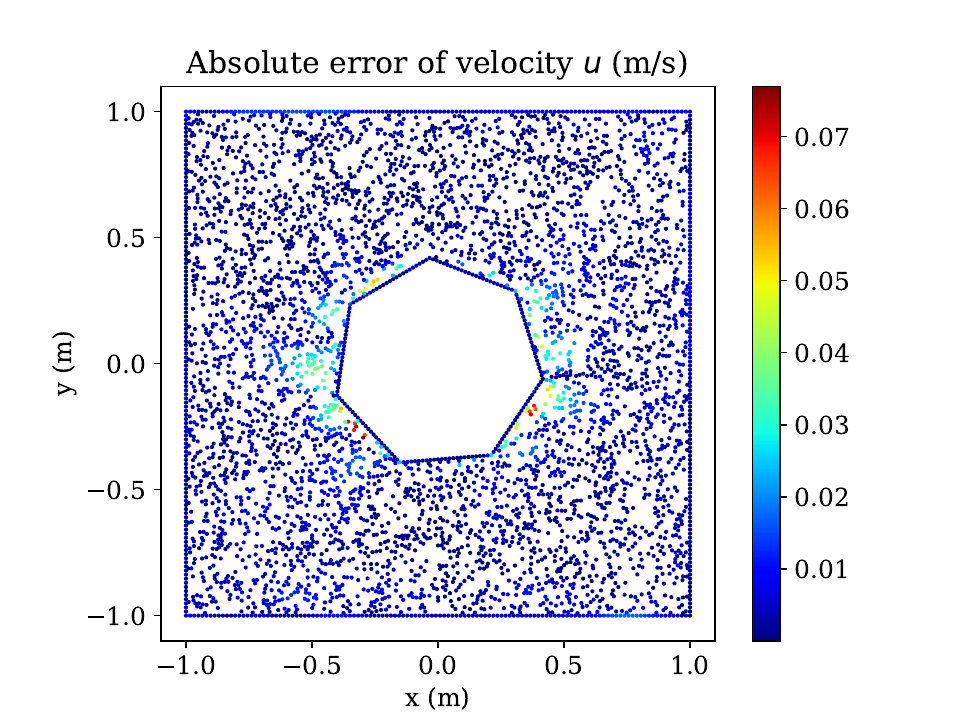}
    \end{subfigure}
    \begin{subfigure}[b]{0.24\textwidth}
        \centering
        \includegraphics[width=\textwidth]{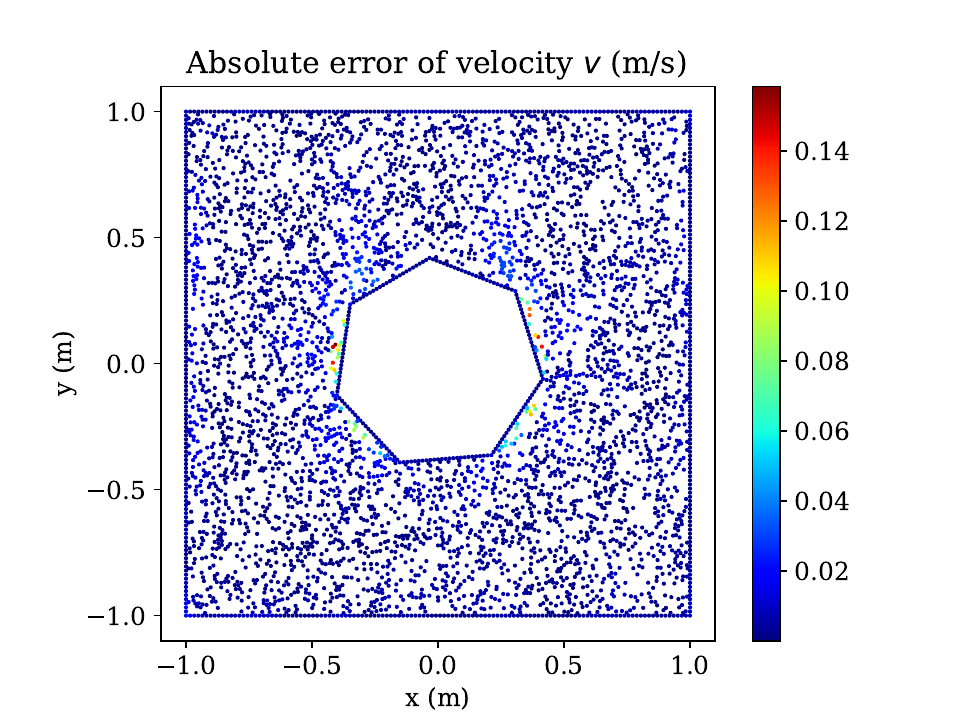}
    \end{subfigure}
    \begin{subfigure}[b]{0.24\textwidth}
        \centering
        \includegraphics[width=\textwidth]{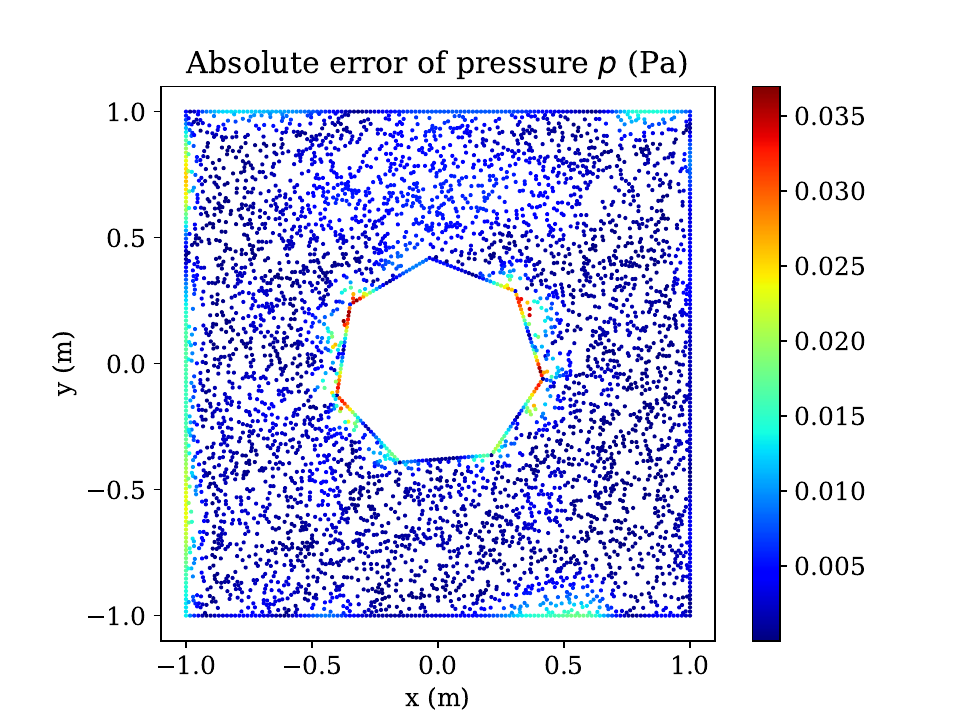}
    \end{subfigure}
     \begin{subfigure}[b]{0.24\textwidth}
        \centering
        \includegraphics[width=\textwidth]{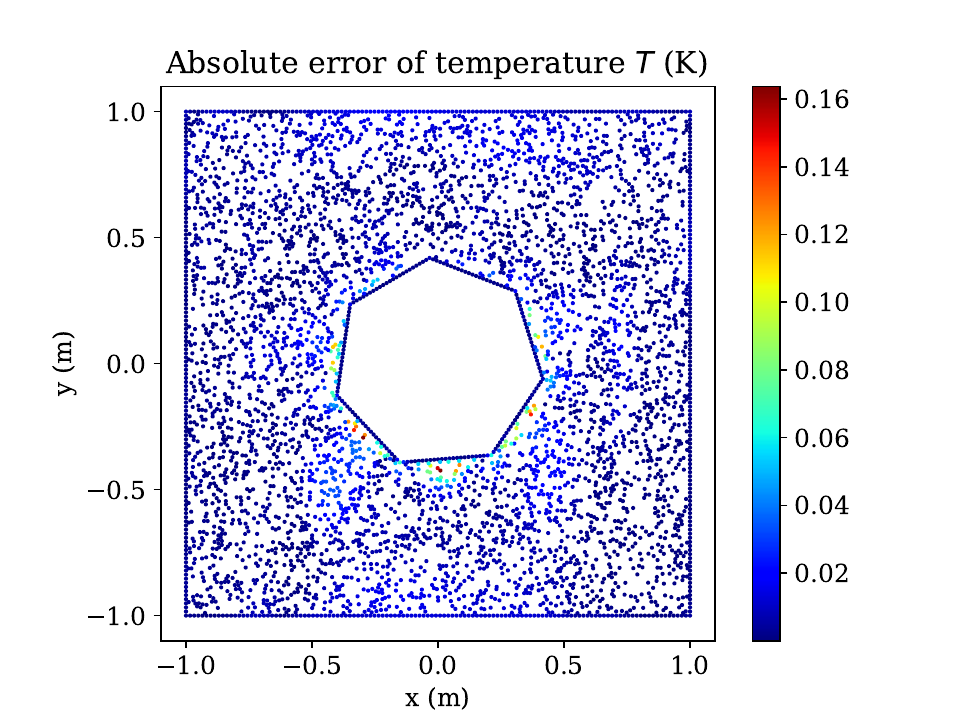}
    \end{subfigure}

  \caption{The second set of examples comparing the ground truth and predictions of the physics-informed KAN PointNet for the velocity, pressure, and temperature fields. The Jacobi polynomial used has a degree of 2, with $\alpha = \beta = -0.5$. Here, $n_s = 0.5$.}
  \label{Fig3}
\end{figure}


\begin{figure}[!htbp]
  \centering 
      \begin{subfigure}[b]{0.24\textwidth}
        \centering
        \includegraphics[width=\textwidth]{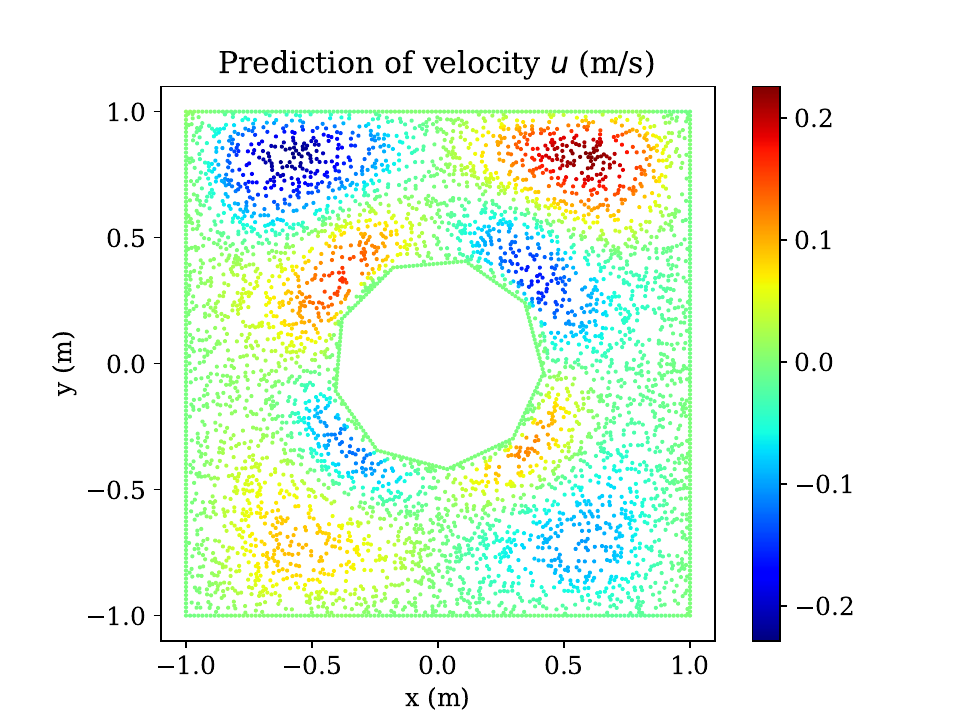}
    \end{subfigure}
    \begin{subfigure}[b]{0.24\textwidth}
        \centering
        \includegraphics[width=\textwidth]{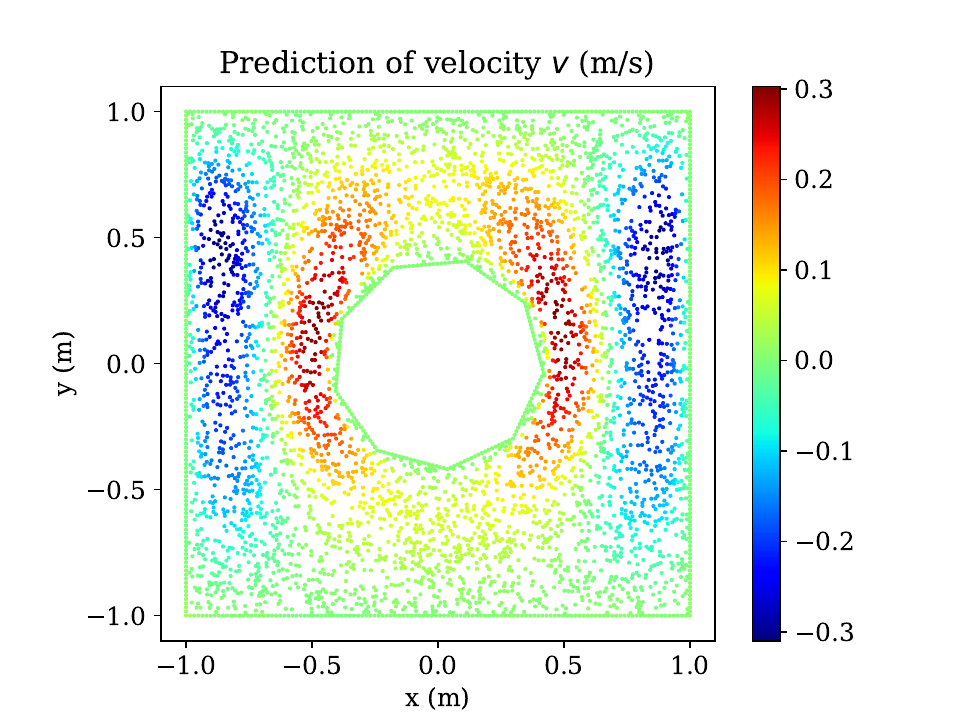}
    \end{subfigure}
    \begin{subfigure}[b]{0.24\textwidth}
        \centering
        \includegraphics[width=\textwidth]{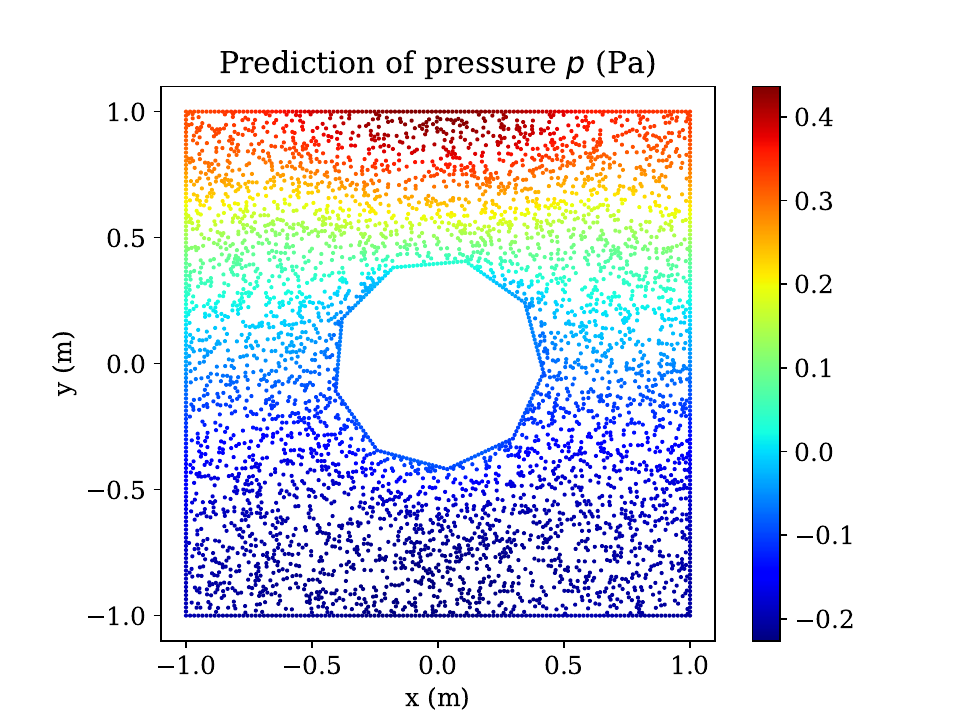}
    \end{subfigure}
     \begin{subfigure}[b]{0.24\textwidth}
        \centering
        \includegraphics[width=\textwidth]{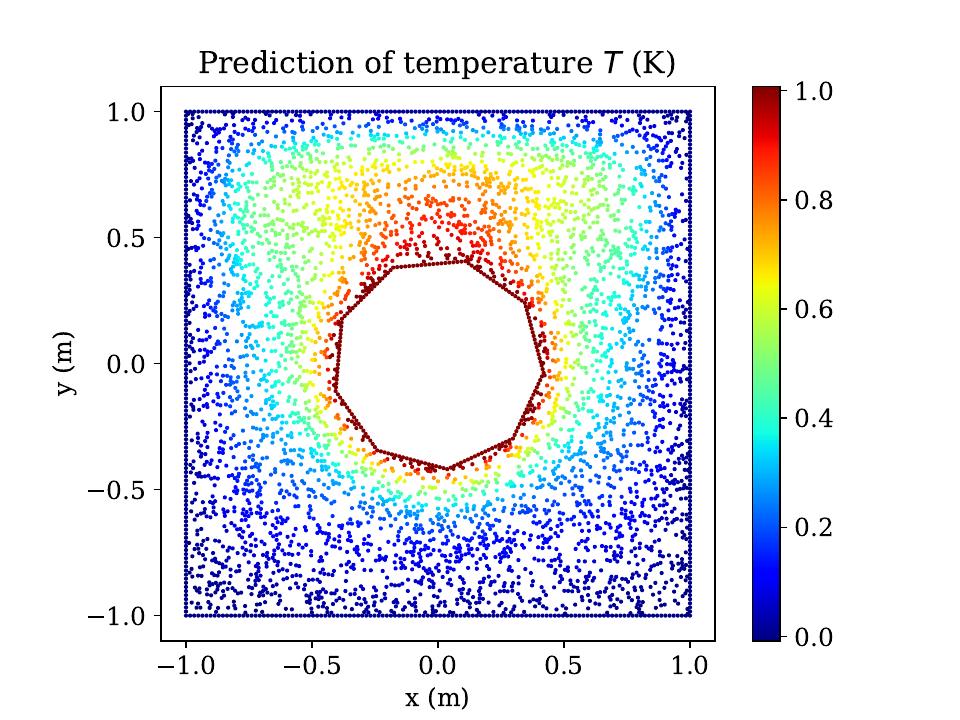}
    \end{subfigure}


  \centering 
      \begin{subfigure}[b]{0.24\textwidth}
        \centering
        \includegraphics[width=\textwidth]{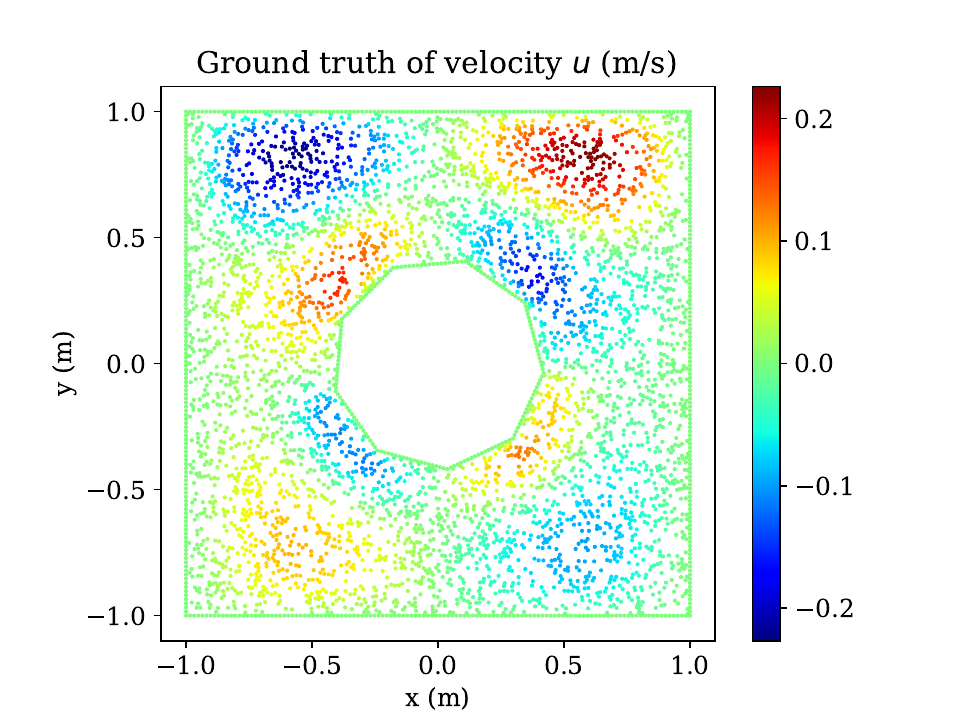}
    \end{subfigure}
    \begin{subfigure}[b]{0.24\textwidth}
        \centering
        \includegraphics[width=\textwidth]{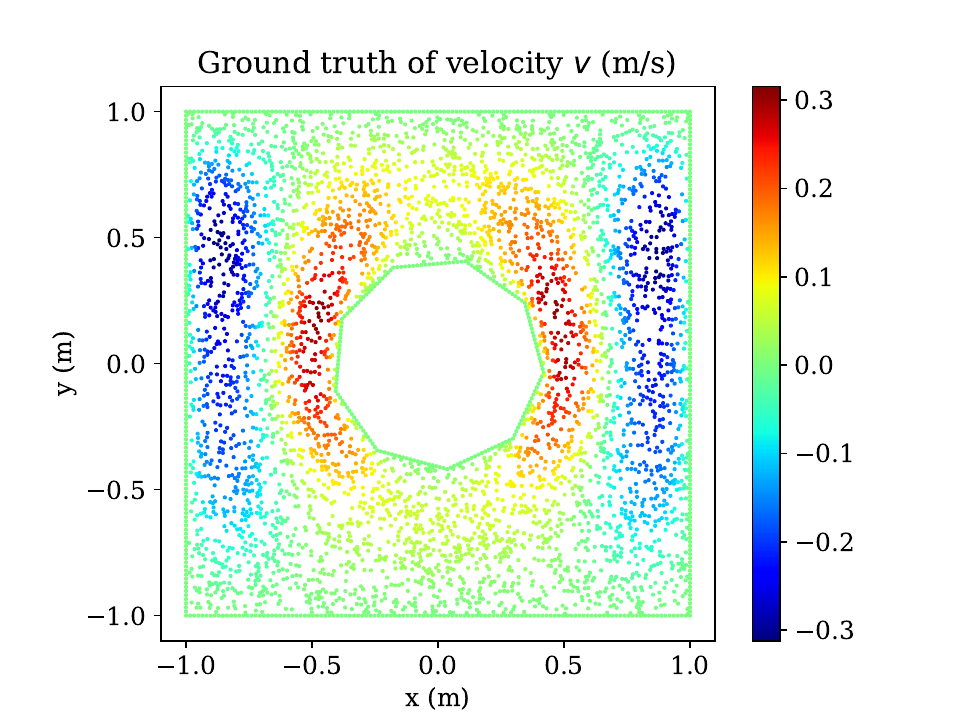}
    \end{subfigure}
    \begin{subfigure}[b]{0.24\textwidth}
        \centering
        \includegraphics[width=\textwidth]{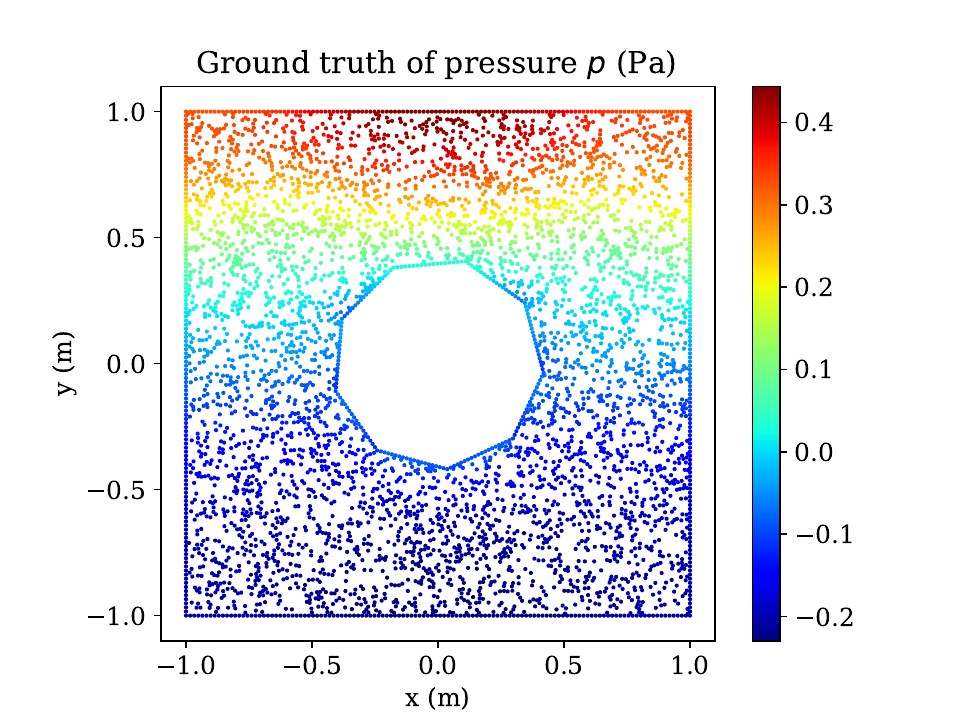}
    \end{subfigure}
     \begin{subfigure}[b]{0.24\textwidth}
        \centering
        \includegraphics[width=\textwidth]{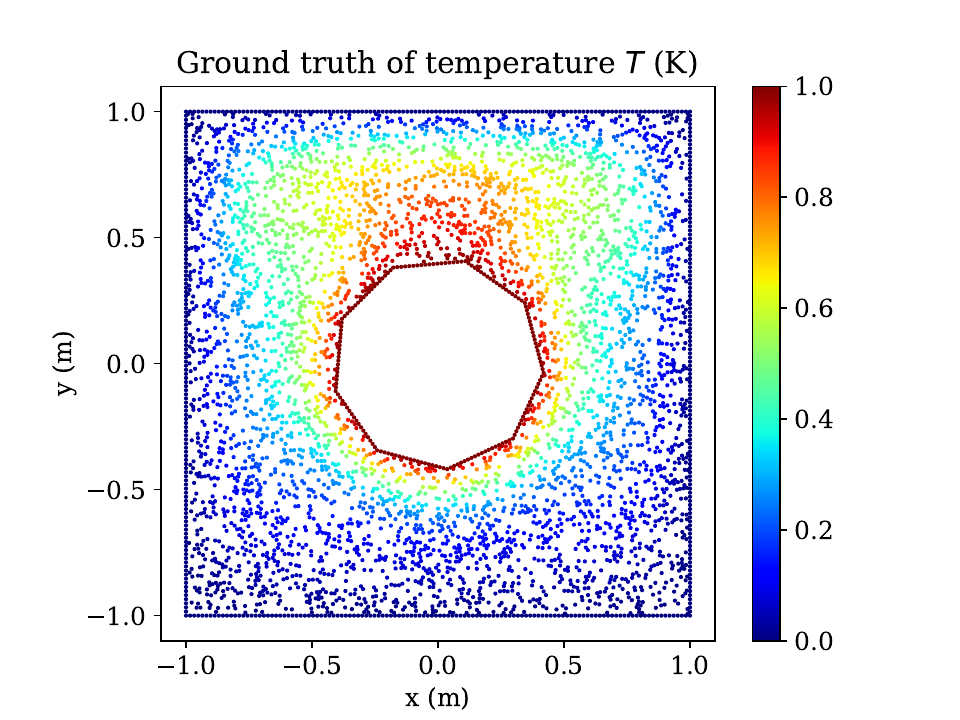}
    \end{subfigure}
    

  \centering 
      \begin{subfigure}[b]{0.24\textwidth}
        \centering
        \includegraphics[width=\textwidth]{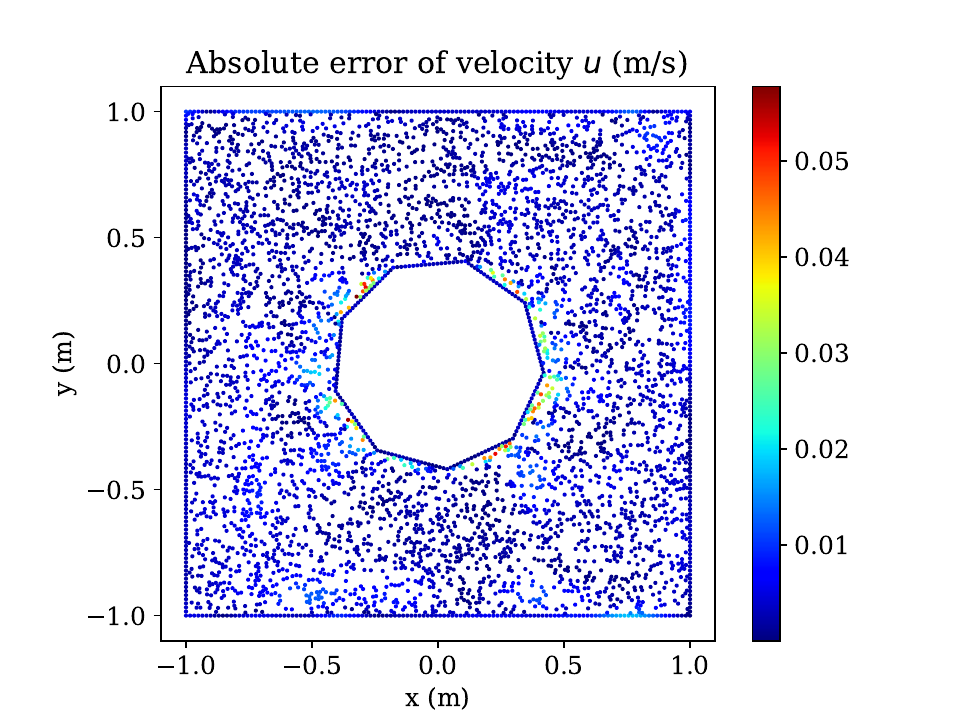}
    \end{subfigure}
    \begin{subfigure}[b]{0.24\textwidth}
        \centering
        \includegraphics[width=\textwidth]{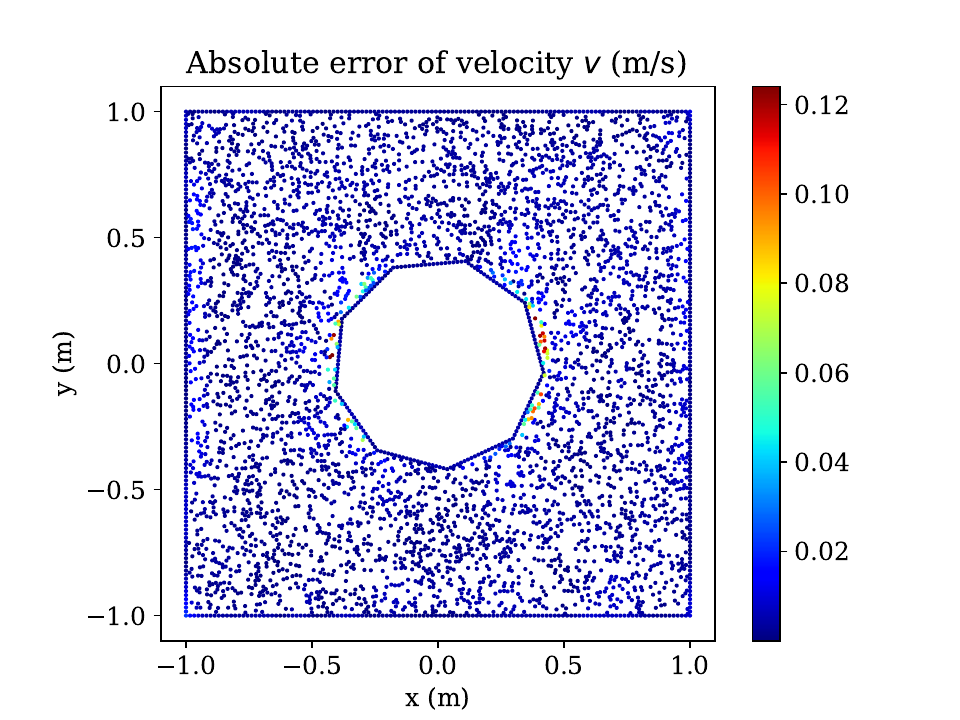}
    \end{subfigure}
    \begin{subfigure}[b]{0.24\textwidth}
        \centering
        \includegraphics[width=\textwidth]{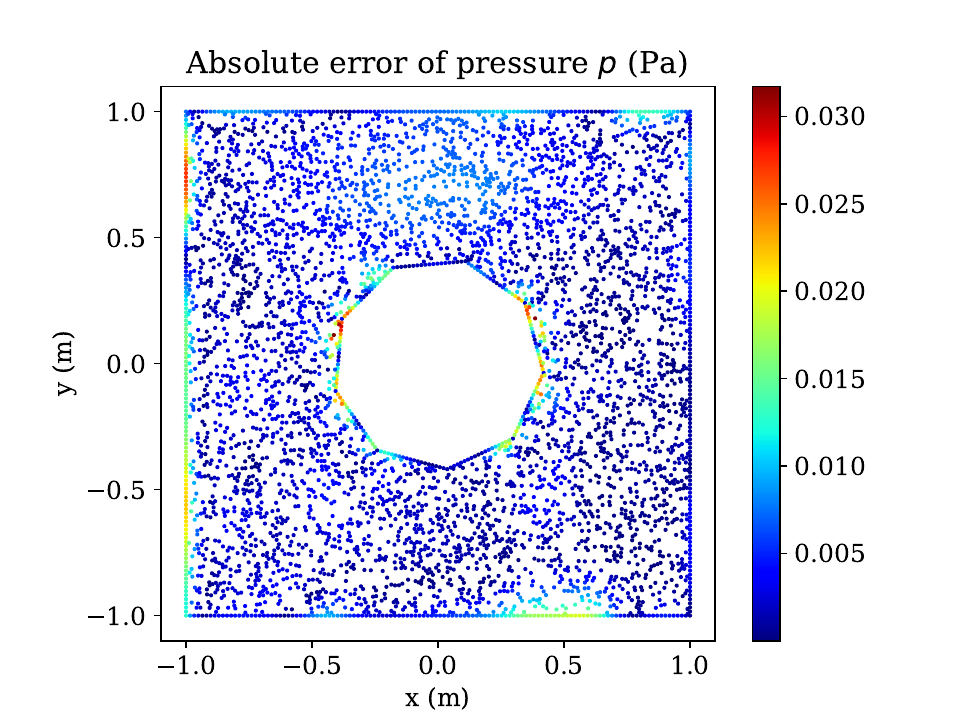}
    \end{subfigure}
     \begin{subfigure}[b]{0.24\textwidth}
        \centering
        \includegraphics[width=\textwidth]{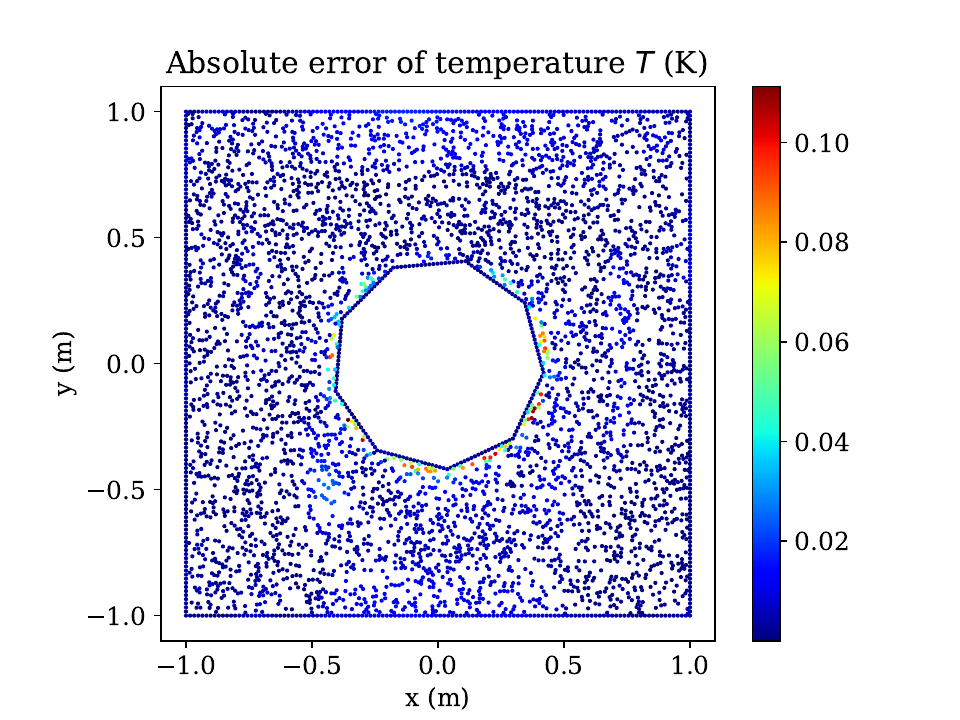}
    \end{subfigure}

  \caption{The third set of examples comparing the ground truth and predictions of the physics-informed KAN PointNet for the velocity, pressure, and temperature fields. The Jacobi polynomial used has a degree of 2, with $\alpha = \beta = -0.5$. Here, $n_s = 0.5$.}
  \label{Fig4}
\end{figure}


\begin{figure}[!htbp]
  \centering 
      \begin{subfigure}[b]{0.24\textwidth}
        \centering
        \includegraphics[width=\textwidth]{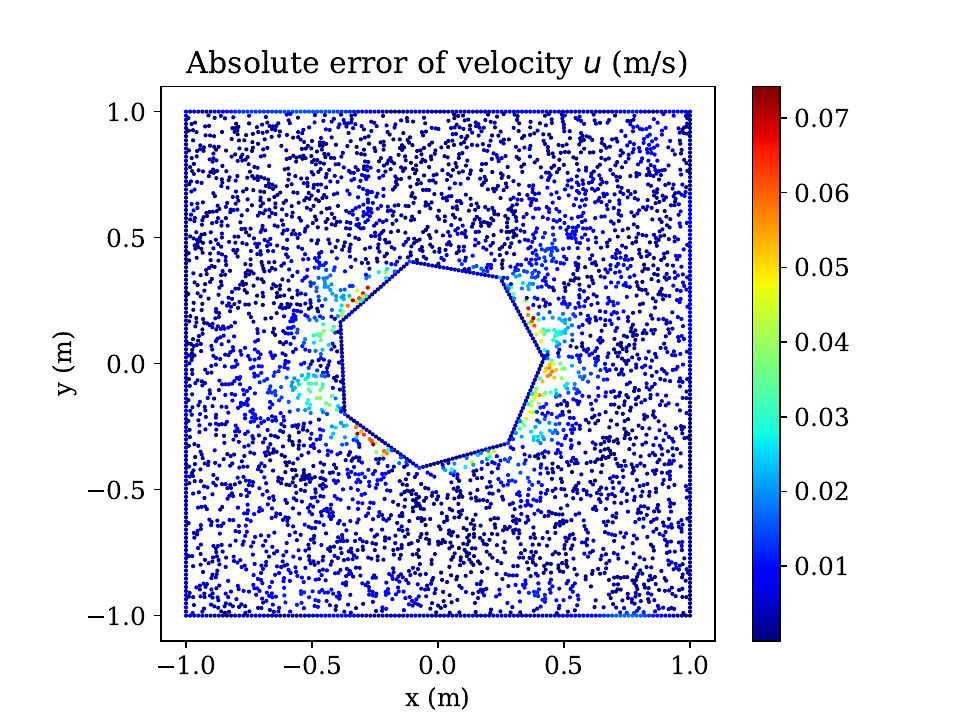}
    \end{subfigure}
    \begin{subfigure}[b]{0.24\textwidth}
        \centering
        \includegraphics[width=\textwidth]{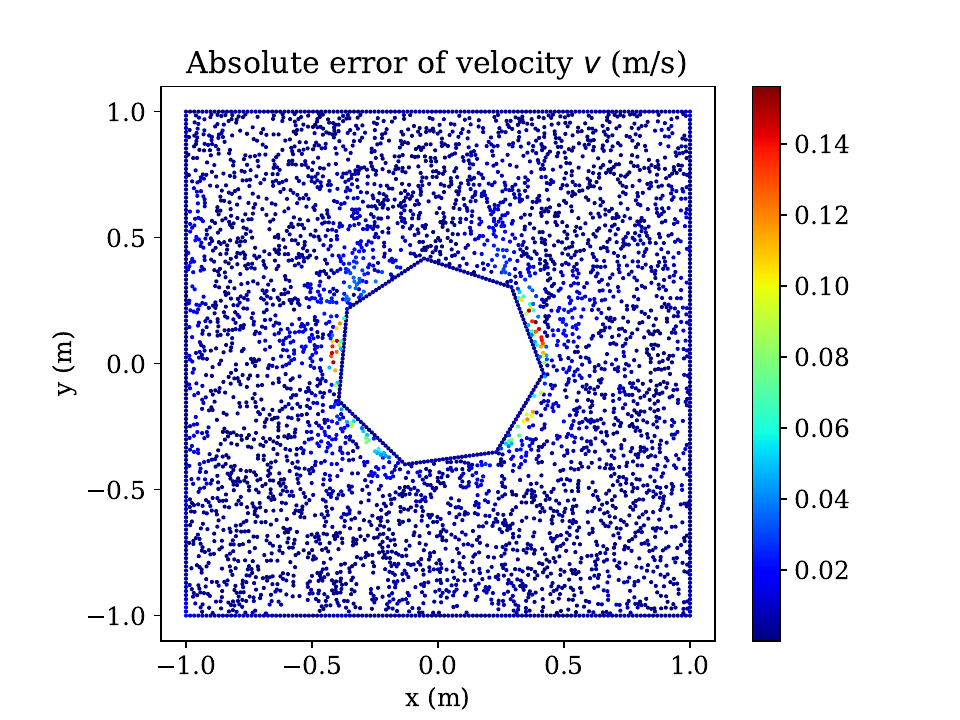}
    \end{subfigure}
    \begin{subfigure}[b]{0.24\textwidth}
        \centering
        \includegraphics[width=\textwidth]{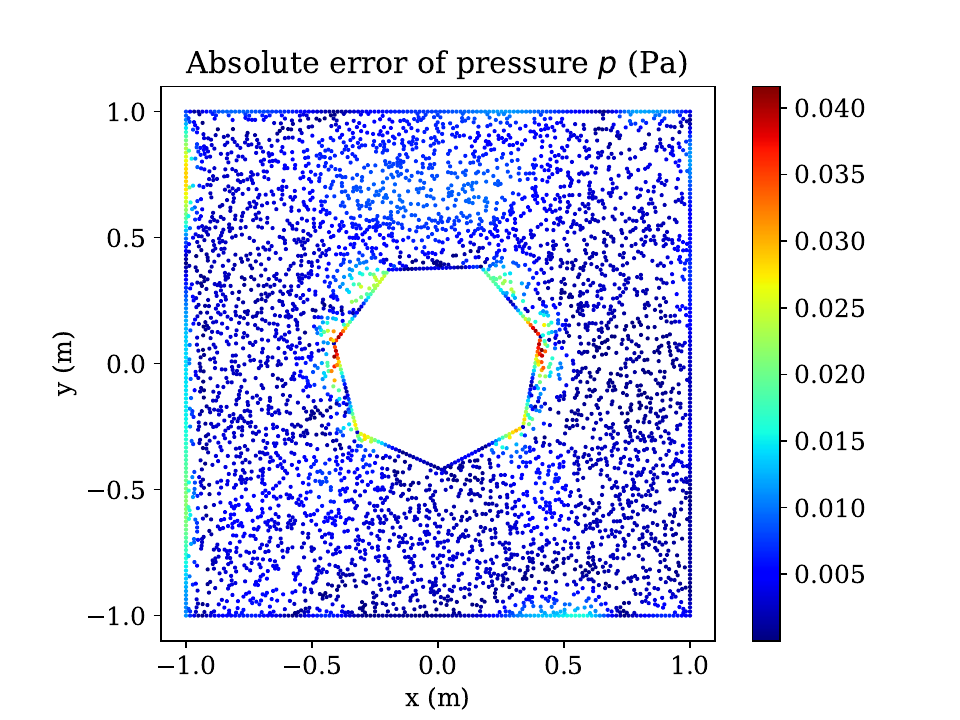}
    \end{subfigure}
     \begin{subfigure}[b]{0.24\textwidth}
        \centering
        \includegraphics[width=\textwidth]{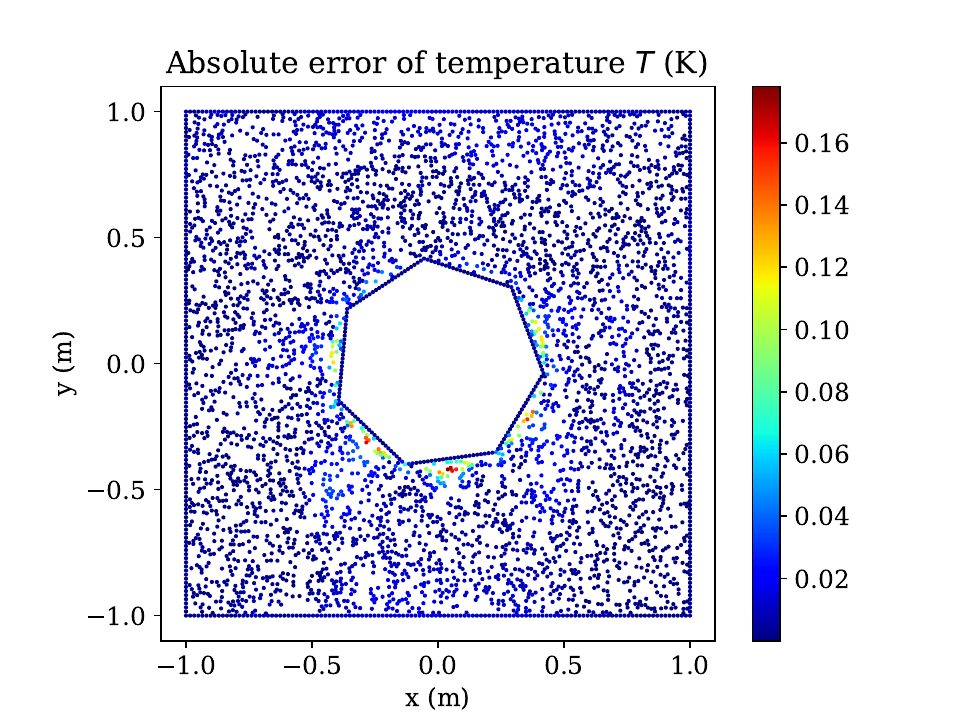}
    \end{subfigure}

    
      \begin{subfigure}[b]{0.24\textwidth}
        \centering
        \includegraphics[width=\textwidth]{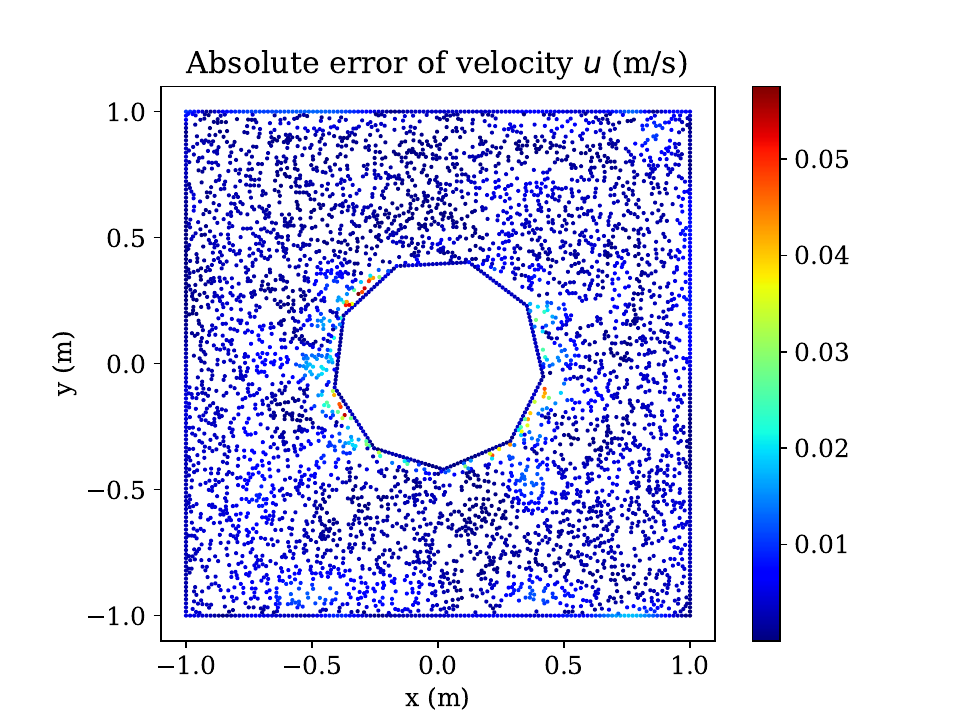}
    \end{subfigure}
    \begin{subfigure}[b]{0.24\textwidth}
        \centering
        \includegraphics[width=\textwidth]{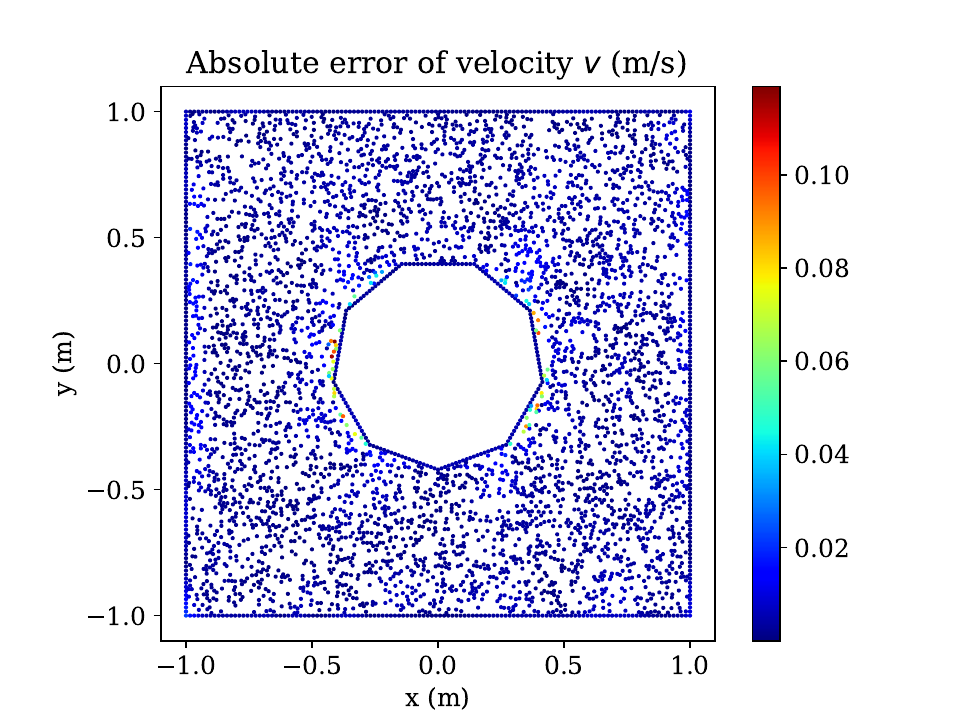}
    \end{subfigure}
    \begin{subfigure}[b]{0.24\textwidth}
        \centering
        \includegraphics[width=\textwidth]{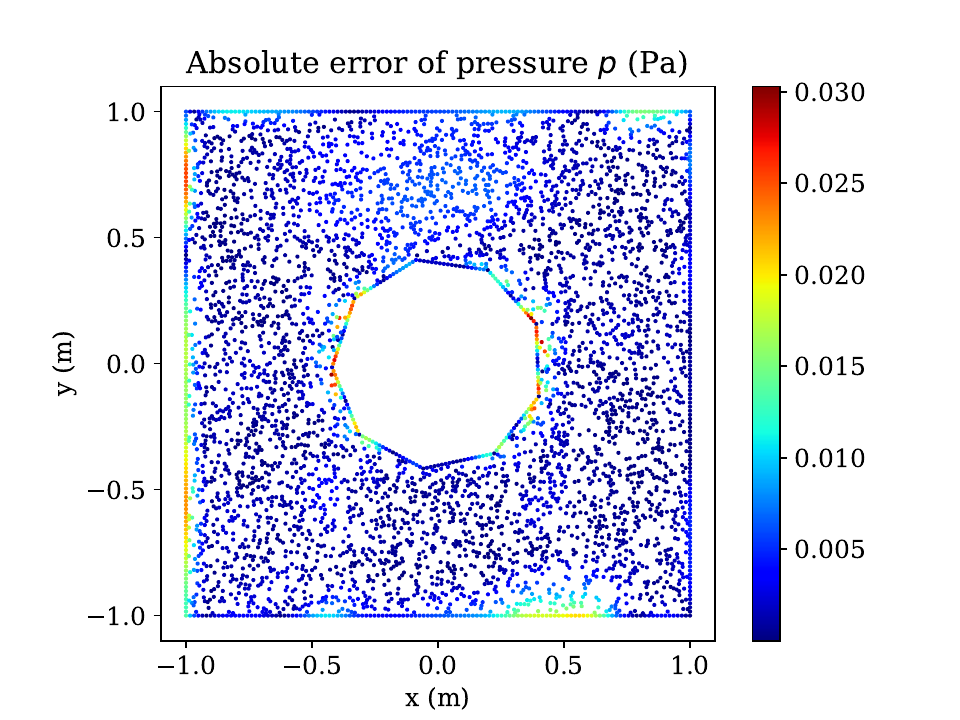}
    \end{subfigure}
     \begin{subfigure}[b]{0.24\textwidth}
        \centering
        \includegraphics[width=\textwidth]{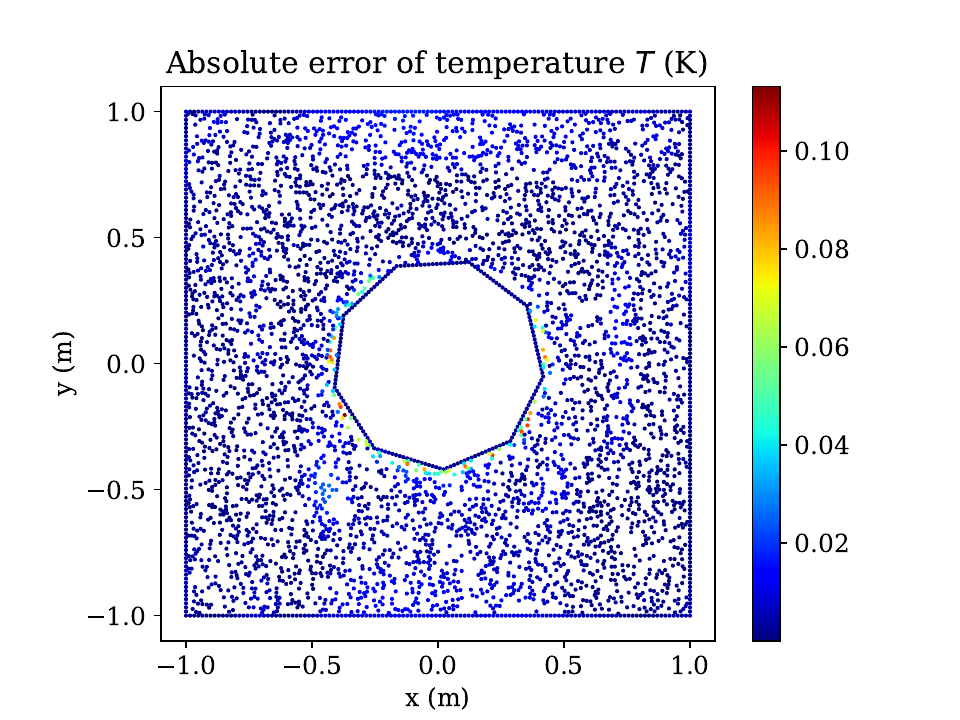}
    \end{subfigure}

  \caption{Distribution of absolute pointwise error for the prediction of the velocity, pressure, and temperature fields by the physics-informed KAN PointNet for the geometries when the relative pointwise error (\(L_2\) norm) becomes maximum (first row) and minimum (second row). The Jacobi polynomial used has a degree of 2, with \(\alpha = \beta = -0.5\). Here, \(n_s = 0.5\) is set.}
  \label{Fig6}
\end{figure}


\begin{figure}[!htbp]
  \centering 
      \begin{subfigure}[b]{0.49\textwidth}
       \caption{Equilateral octagon, $\alpha=\beta=-0.5$}
        \centering
        \includegraphics[width=\textwidth]{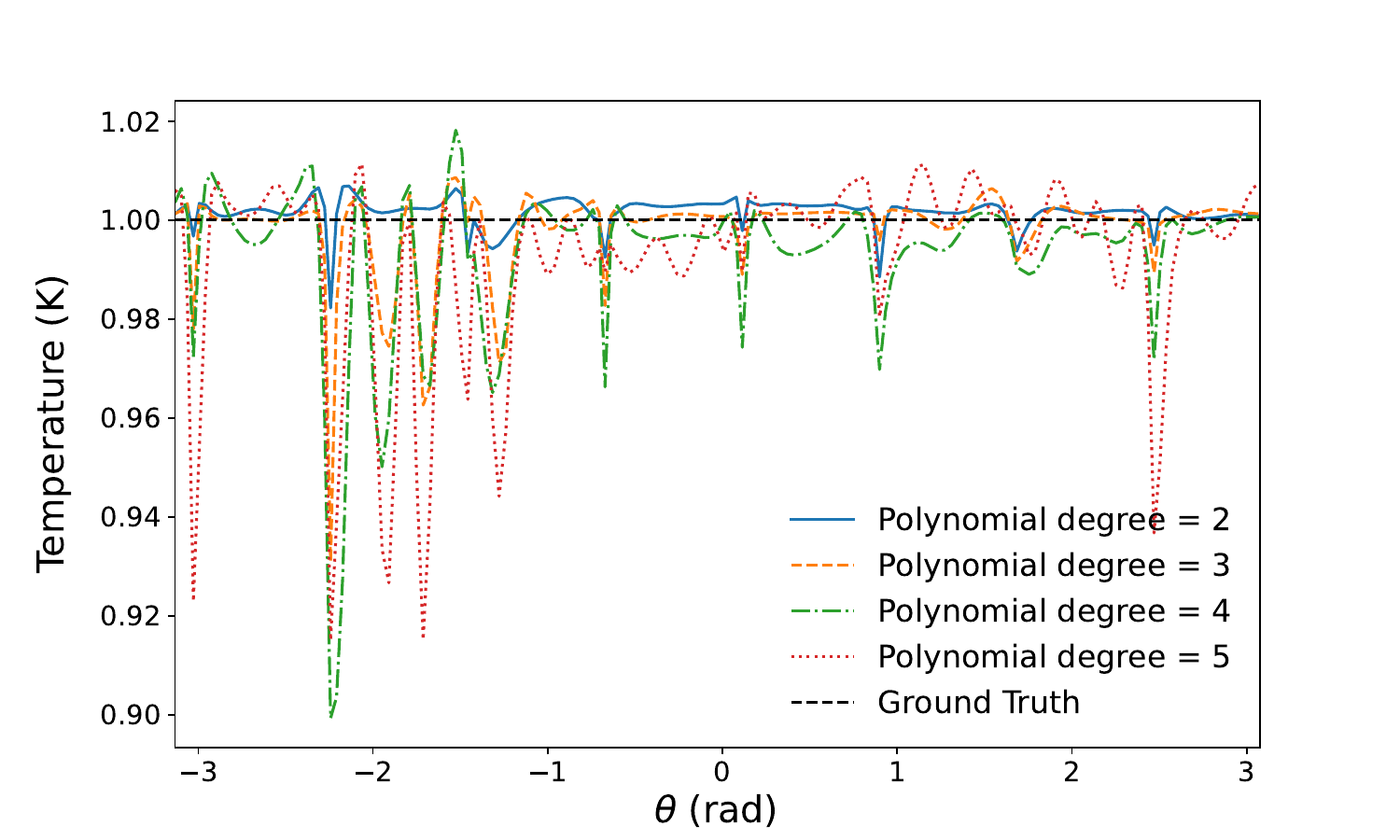}
    \end{subfigure}
    \begin{subfigure}[b]{0.49\textwidth}
    \caption{Equilateral octagon, polynomial degree $= 2$}
        \centering
        \includegraphics[width=\textwidth]{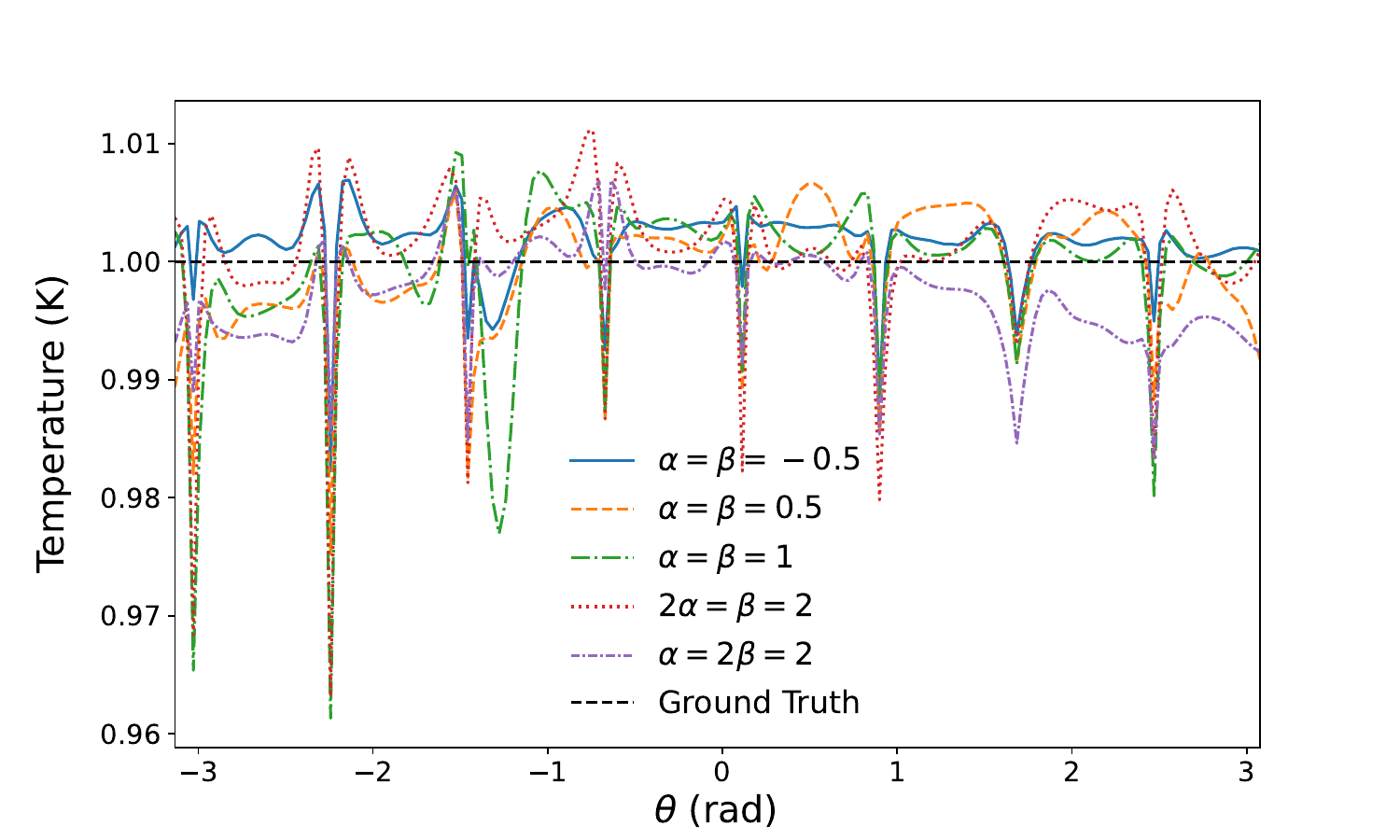}
    \end{subfigure}
 

       \begin{subfigure}[b]{0.49\textwidth}
       \caption{Equilateral heptagon, $\alpha=\beta=-0.5$}
        \centering
        \includegraphics[width=\textwidth]{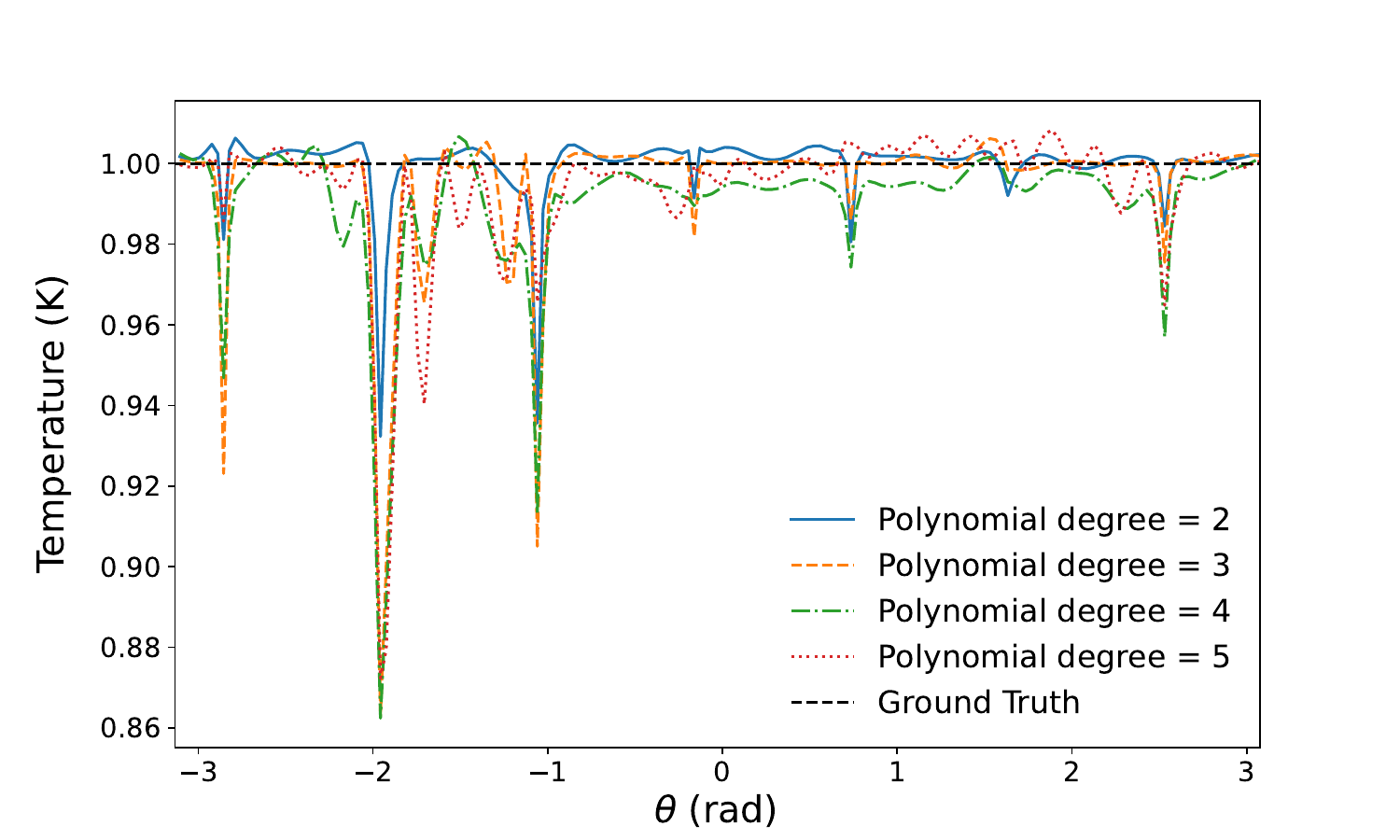}
    \end{subfigure}
    \begin{subfigure}[b]{0.49\textwidth}
    \caption{Equilateral heptagon, polynomial degree $= 2$}
        \centering
        \includegraphics[width=\textwidth]{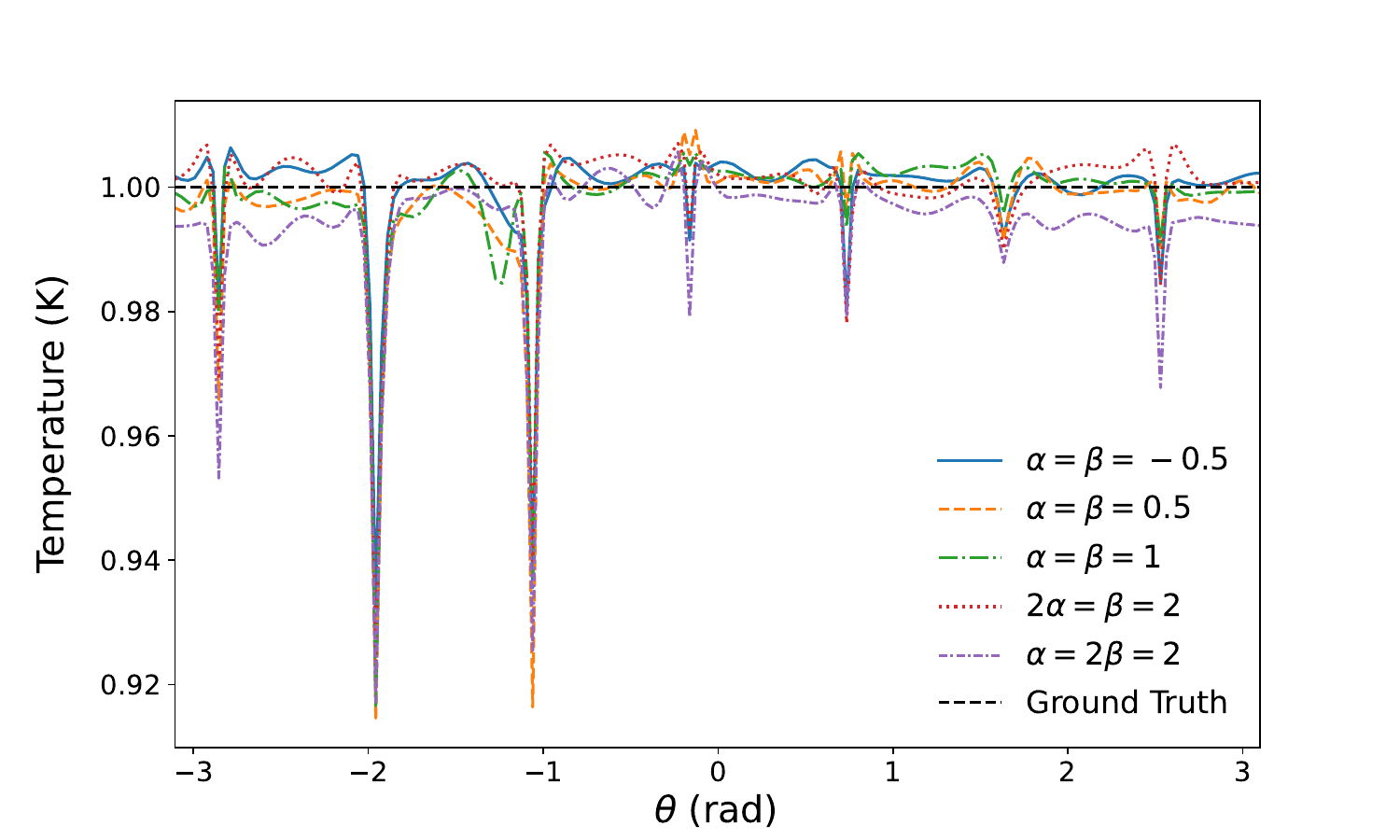}
    \end{subfigure}

       \begin{subfigure}[b]{0.49\textwidth}
       \caption{Equilateral nonagon, $\alpha=\beta=-0.5$}
        \centering
        \includegraphics[width=\textwidth]{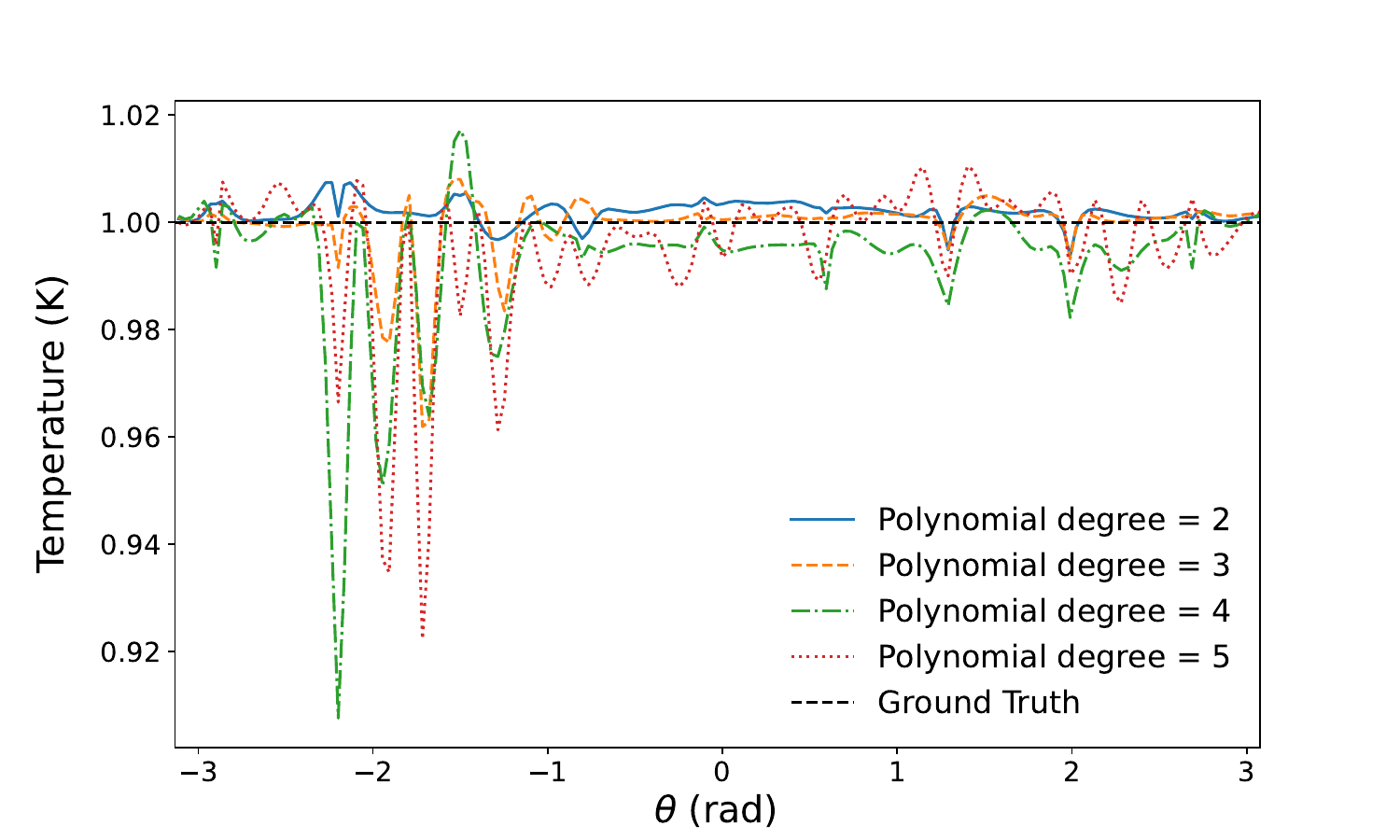}
    \end{subfigure}
    \begin{subfigure}[b]{0.49\textwidth}
    \caption{Equilateral nonagon, polynomial degree $= 2$}
        \centering
        \includegraphics[width=\textwidth]{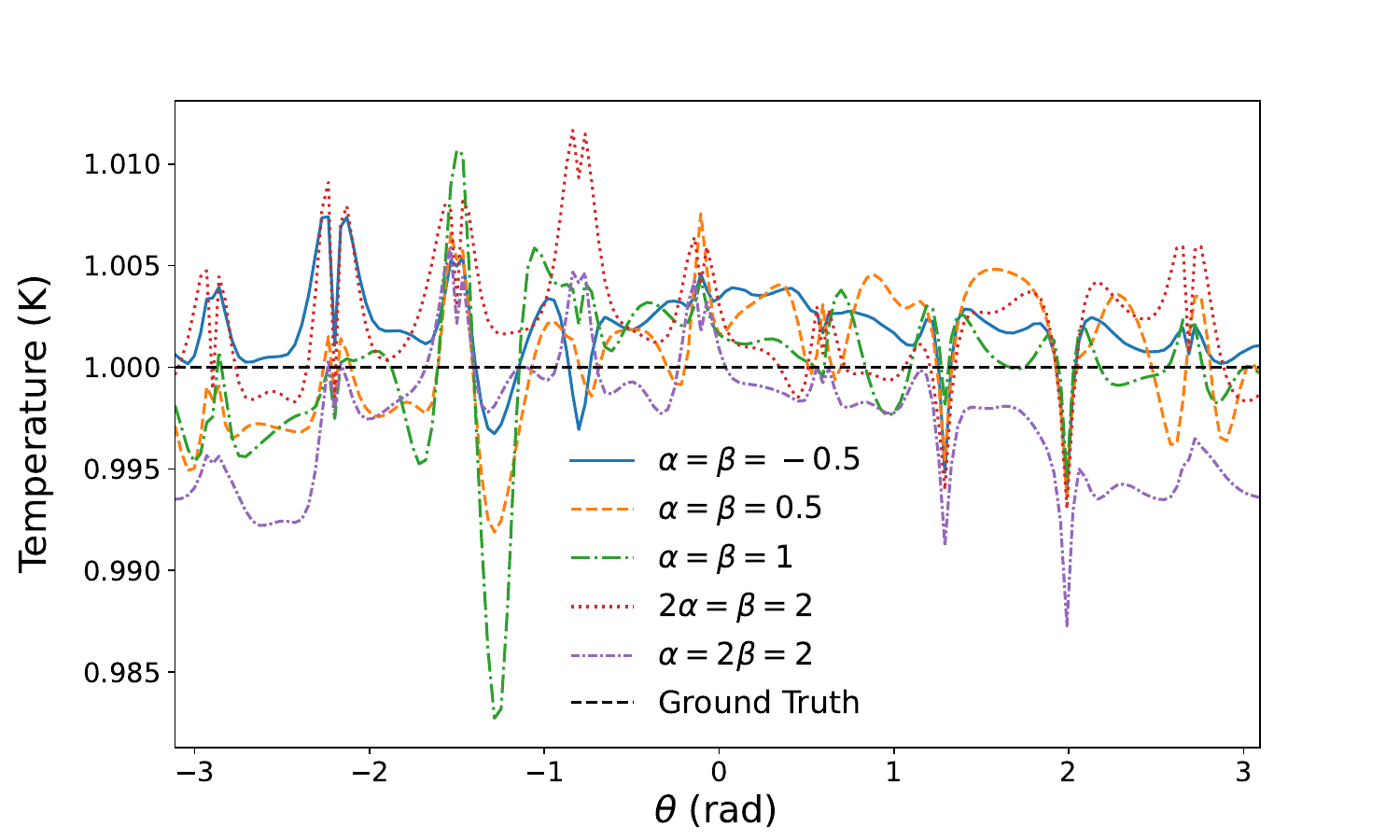}
    \end{subfigure}

  \caption{Temperature distributions predicted by the physics-informed KAN PointNet along the surface of the cylinder for the octagon with $\Omega=31^\circ$ (first row), the heptagon with $\Omega=6^\circ$ (second row), and the nonagon with $\Omega=26^\circ$ (third row), shown for different Jacobi polynomial degrees (first column) and various values of $\alpha$ and $\beta$ in the Jacobi polynomials (second column). Here, \(n_s = 0.5\) is used. The angle $\theta$ is defined with reference to the positive $x$-axis and increases counterclockwise (or decreases clockwise). See Table \ref{Table1} and the text for the definition of $\Omega$.}
  \label{Fig7}
\end{figure}


\begin{figure}[!htbp]
  \centering 
      \begin{subfigure}[b]{0.49\textwidth}
       \caption{$\alpha = \beta = -0.5$}
        \centering
        \includegraphics[width=\textwidth]{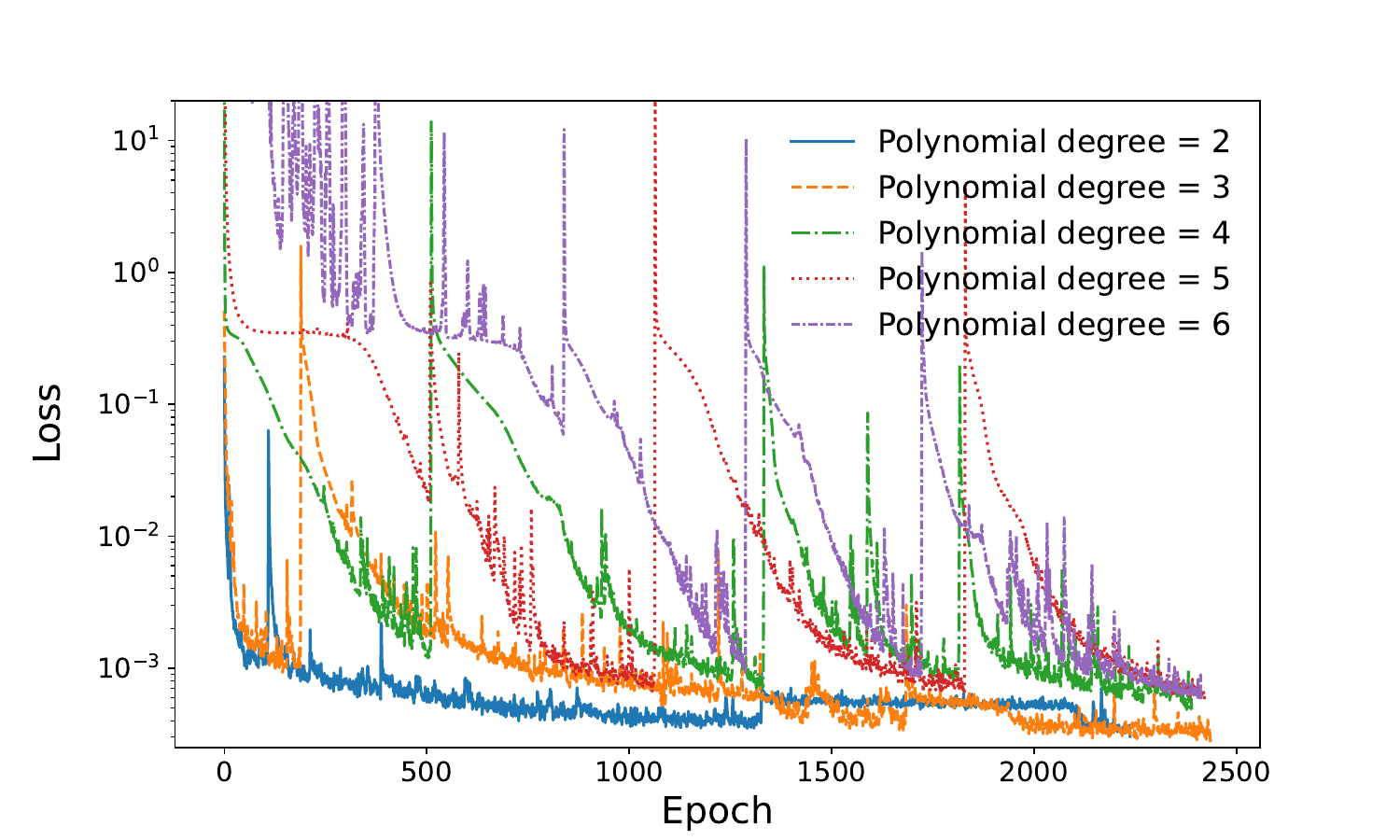}
    \end{subfigure}
    \begin{subfigure}[b]{0.49\textwidth}
    \caption{Polynomial degree $=2$}
        \centering
        \includegraphics[width=\textwidth]{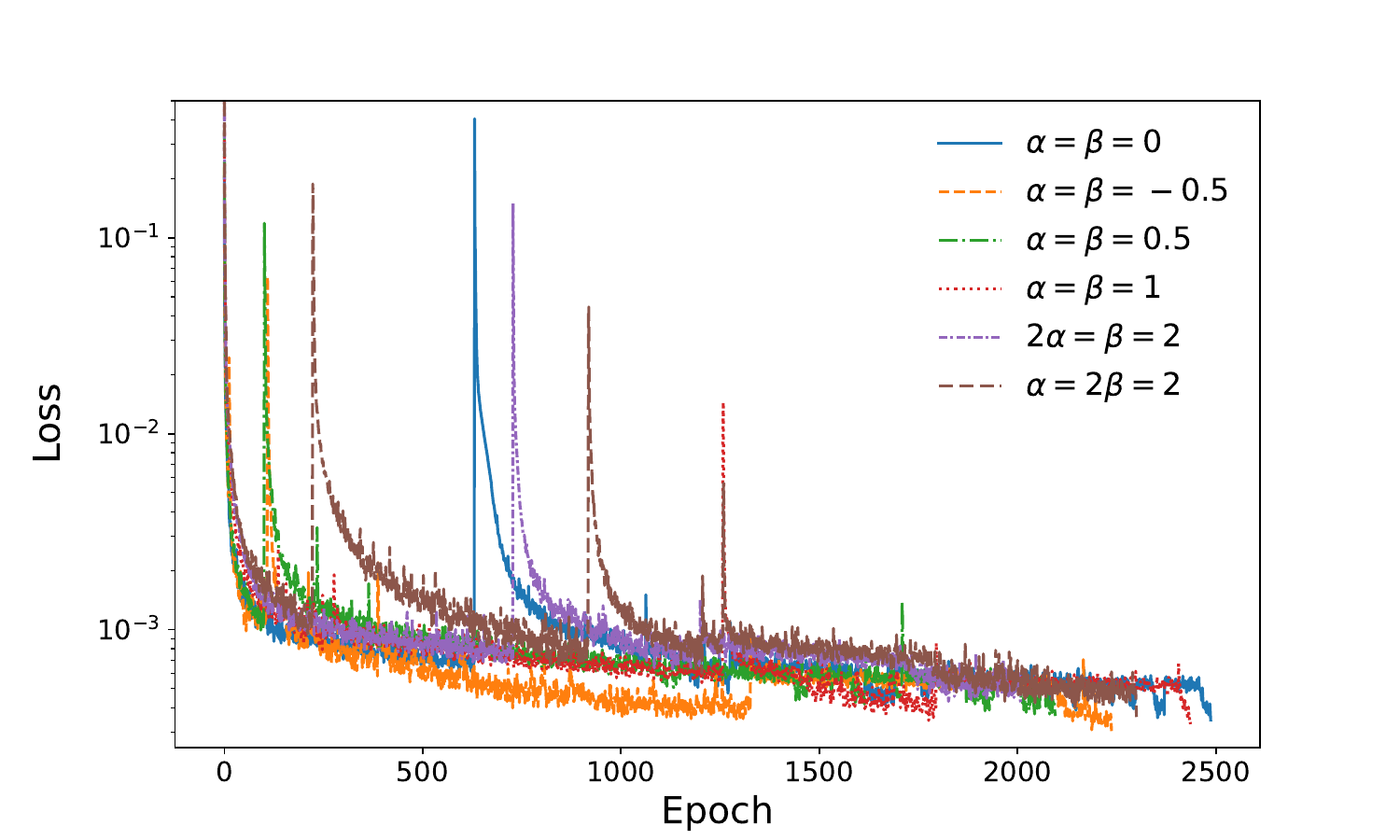}
    \end{subfigure}
    
 \caption{Loss evolution of the physics-informed KAN PointNet for different Jacobi polynomial degrees (left) and for various values of \( \alpha \) and \( \beta \) in the Jacobi polynomials (right). Here, \( n_s = 0.5 \) is used.}
  \label{Fig8}
\end{figure}


\begin{table}[width=1.0\linewidth,cols=6,pos=!htbp]
\caption{Computational cost and error analysis of the velocity, temperature, and pressure fields of 135 geometries predicted by the physics-informed KAN PointNet for different degrees of the Jacobi polynomial. Here, $n_s=0.5$ is set. In the Jacobi polynomial, $\alpha=\beta=-0.5$ is set. $||\cdot||_V$ shows the $L^2$ norm over the entire domain $V$ and $||\cdot||_\Gamma$ shows the $L^2$ norm over the inner cylinder surface.}
\label{Table2}
\begin{tabular*}{\tblwidth}{@{} LLLLLL@{} }
\toprule
Polynomial degree & 2  &  3 &  4 &  5 & 6 \\
\midrule
Average $||\Tilde{u}-u||_V/||u||_V$ & 1.08973E$-$1 & 1.19702E$-$1 & 1.13309E$-$1 & 1.88704E$-$1 & 2.05859E$-$1 \\
Maximum $||\Tilde{u}-u||_V/||u||_V$ & 1.31928E$-$1 & 1.40406E$-$1 & 1.43837E$-$1 & 2.17029E$-$1 & 2.38389E$-$1 \\
Minimum $||\Tilde{u}-u||_V/||u||_V$ & 8.78500E$-$2 & 1.00543E$-$1 & 9.28611E$-$2 & 1.68615E$-$1 & 1.86217E$-$1 \\
\midrule
Average $||\Tilde{v}-v||_V/||v||_V$ & 8.84278E$-$2 & 9.71973E$-$2 & 1.00067E$-$1 & 1.22098E$-$1 & 1.76641E$-$1 \\
Maximum $||\Tilde{v}-v||_V/||v||_V$ & 1.11167E$-$1 & 1.17739E$-$1 & 1.26364E$-$1 & 1.58934E$-$1 & 2.16676E$-$1 \\
Minimum $||\Tilde{v}-v||_V/||v||_V$ & 7.06122E$-$2 & 8.13682E$-$2 & 8.27909E$-$2 & 9.23098E$-$2 & 1.45240E$-$1 \\
\midrule
Average $||\Tilde{p}-p||_V/||p||_V$ & 2.97057E$-$2 & 2.93362E$-$2 & 5.60406E$-$2 & 7.13754E$-$2 & 7.03334E$-$2 \\
Maximum $||\Tilde{p}-p||_V/||p||_V$ & 3.41539E$-$2 & 3.42451E$-$2 & 5.91457E$-$2 & 7.50780E$-$2 & 7.29919E$-$2 \\
Minimum $||\Tilde{p}-p||_V/||p||_V$ & 2.63087E$-$2 & 2.48404E$-$2 & 5.24305E$-$2 & 6.85314E$-$2 & 6.79873E$-$2 \\
\midrule
Average $||\Tilde{T}-T||_V/||T||_V$ & 2.84486E$-$2 & 3.06295E$-$2 & 5.84228E$-$2 & 1.34718E$-$1 & 1.85275E$-$1 \\
Maximum $||\Tilde{T}-T||_V/||T||_V$ & 3.66225E$-$2 & 3.97039E$-$2 & 6.56761E$-$2 & 1.47693E$-$1 & 2.03668E$-$1 \\
Minimum $||\Tilde{T}-T||_V/||T||_V$ & 2.28196E$-$2 & 2.49564E$-$2 & 5.08258E$-$2 & 1.19578E$-$1 & 1.67913E$-$1 \\
\midrule
Average $||\Tilde{T}-T||_{\Gamma}/||T||_{\Gamma}$ & 5.03499E$-$3 & 1.19854E$-$2 & 1.67429E$-$2 & 1.77187E$-$2 & 6.24591E$-$2 \\
Maximum $||\Tilde{T}-T||_{\Gamma}/||T||_{\Gamma}$ & 9.08389E$-$3 & 1.98724E$-$2 & 2.44606E$-$2 & 2.60403E$-$2 & 8.56928E$-$2 \\
Minimum $||\Tilde{T}-T||_{\Gamma}/||T||_{\Gamma}$ & 2.59714E$-$3 & 5.87528E$-$3 & 1.18108E$-$2 & 9.29117E$-$3 & 4.72897E$-$2 \\
\midrule
Minimum loss achieved & 3.03649E$-$4 & 2.75245E$-$4 & 4.84736E$-$4 & 5.61811E$-$4 & 5.63230E$-$4 \\
\midrule
Training time & 15.4 & 23.8 & 52.7 & 74.8 & 102.7 \\
per epoch (s) &  &  &  &  &  \\
\midrule
Number of epochs to & 2238 & 2437 & 2390 & 2422 & 2414 \\
reach the minimum loss &  &  &  &  &  \\
\midrule
Number of trainable & 666880 & 888384 & 1109888 & 1331392 & 1552896 \\
parameters &  &  &  &  & \\
\bottomrule
\end{tabular*}
\end{table}


\subsection{General analysis}
\label{Sect61}

\subsubsection{Visual comparison and overall outcomes}
\label{Sect611}

To begin with, we consider the setup of PI-KAN-PointNet, where the degree of the Jacobi polynomial is set to 2, and $\alpha = \beta = -0.5$ (i.e., the Chebyshev polynomial), with $n_s = 0.5$. The error analysis of this configuration for simultaneously solving the inverse problem over 135 geometries ($m=135$) is reported in the second column of Table \ref{Table2}. As observed, the average relative pointwise error ($L^2$ norm) for the velocity, pressure, and temperature fields, as well as the temperature on the surface of the inner cylinder, is below 10.9\%. Specifically, this error for the pressure and temperature fields remains under 3\%. Moreover, PI-KAN-PointNet successfully estimates the unknown temperature on the surface of the inner cylinder, with a maximum relative pointwise error ($L^2$ norm) of less than 1\%.

As a few examples, the velocity, pressure, and temperature fields for domains with inner cylinders in the shapes of an octagon, heptagon, and nonagon are visualized in Fig. \ref{Fig2}, Fig. \ref{Fig3}, and Fig. \ref{Fig4}, respectively. In each of these figures, the predictions by PI-KAN-PointNet, the numerical solver solution (serving as the ground truth), and the corresponding absolute pointwise error are shown. This visual comparison demonstrates a good agreement between the predicted and ground truth solutions. The maximum error for all variables occurs near or on the surface of the inner cylinders. This is primarily due to the fact that, across different domains in the dataset, the most significant geometric variations occur along the inner surface. Moreover, from a mathematical perspective, since the temperature at the surface of the inner cylinder is unknown in the defined inverse problem, the temperature variable exhibits the highest error along this surface. This local temperature error propagates to the surface of the inner cylinder at the velocity and pressure fields as well, given that these variables are coupled (see Eqs. \ref{Eq1}--\ref{Eq3}).

To better understand the effects of the inner cylinders' geometry on the performance of PI-KAN-PointNet, we plot the absolute pointwise errors for the domains where the maximum and minimum relative pointwise errors of each quantity occur in Fig. \ref{Fig6}. Interestingly, the maximum error for all variables occurs in the domain where the inner cylinder has a hexagonal shape. Conversely, the minimum error for all variables is observed in the domain where the inner cylinder has a nonagonal shape. This observation demonstrates that sharper corners introduce greater challenges for PI-KAN-PointNet. Additionally, we observe that the pressure variable exhibits local errors on both the inner and outer boundaries. This is because no pressure boundary condition is imposed in the loss function (see Eq. \ref{Eq30}), as pressure is an implicit variable in the Navier-Stokes equations \citep{timmermans1996approximate}. Beyond the mathematical and physical reasoning behind the relatively higher local errors at the boundaries, this observation can also be interpreted from a computer graphics perspective. Considering the original application of PointNet for three-dimensional point cloud segmentation \citep{qi2017pointnet}, boundary points of an object are among the critical points, meaning that they contribute most significantly to the geometric features learned by the network through a max-pooling operation (see, e.g., Fig. 7 and Fig. 17 of Ref. \cite{qi2017pointnet}). This structure then leads to a situation where these points are also among the most challenging to classify in part segmentation \citep{qi2017pointnet}. A similar trend is observed in the context of the present study.


\subsubsection{Effect of Jacobi polynomial degree}
\label{Sect612}

To investigate the effect of the Jacobi polynomial degree, we fix $n_s=0.5$ and $\alpha=\beta = -0.5$ while varying the polynomial degree from 2 to 6. The results of this investigation appear in Table \ref{Table2}. According to the average relative pointwise error ($L^2$ norm) listed in Table \ref{Table2}, the general trend is that increasing the Jacobi polynomial degree leads to a higher relative error. This observation can be explained as follows. Increasing the polynomial degree for data fitting results in an oscillatory polynomial due to Runge's phenomenon (see e.g., Refs. \citep{trefethen1991two,henrici1982essentials,berrut2004barycentric,de2020polynomial}). This effect can also be observed in the first column of Fig. \ref{Fig7}, where the temperature distribution along the inner surfaces is plotted for inner cylinders in the shapes of an octagon, heptagon, and nonagon. As shown in Fig. \ref{Fig7}, the highest local error occurs at sharp corners and worsens as the degree of the Jacobi polynomial increases. Furthermore, Fig. \ref{Fig7} indicates that a Jacobi polynomial degree of 2 exhibits the best performance and provides the most accurate predictions. Note that we do not show the results for the Jacobi polynomial of degree 6 in Fig. \ref{Fig7} due to its extreme oscillatory behavior compared to other polynomials. \citet{shukla2024comprehensive} reported a similar observation when they used Physics-Informed Chebyshev–KAN \citep{shukla2024comprehensive} with high-order Chebyshev polynomials for solving forward and inverse problems.

It is important to explain the motivation behind increasing the order of the polynomial. The rationale is that, as the polynomial order increases, the number of trainable parameters grows, with the expectation that the network's predictive ability may improve. At the very least, there was hope that a trade-off might exist between the effects of Runge's phenomenon and the enhanced learning capacity of high-order Jacobi polynomials. However, our observations indicate that the negative effect of Runge's phenomenon completely dominates, outweighing any potential benefits. Note that increasing the degree of the Jacobi polynomial increases the number of trainable parameters, requiring more GPU memory and computational resources while also increasing the training time per epoch. Consequently, a higher polynomial degree not only imposes greater computational costs and GPU memory usage but also leads to increased error. Thus, the strategy of increasing the polynomial degree proves to be inefficient and fragile in the framework of PI-KAN-PointNet for solving inverse problems.

It should be noted that this negative effect arises because we solve an inverse problem using a physics-informed machine learning framework within the category of weakly supervised learning. In this framework, the loss function is designed to enforce agreement with sensor measurements. Imposing such constraints introduces oscillatory behavior when a high-order polynomial is used. In contrast, \cite{kashefi2024KANpointnet} demonstrated that KA-PointNet, which combines KAN and PointNet in a supervised learning framework, does not exhibit Runge's phenomenon. For instance, as shown in Fig. 20 of \citet{kashefi2024KANpointnet}, the pressure predictions on the surface of the cylinder remain stable even for a polynomial order of 6. The author of Ref. \cite{kashefi2024KANpointnet} did not investigate higher orders due to limitations in GPU memory.

Based on the information provided in Table \ref{Table2}, increasing the polynomial degree generally leads to a higher minimum loss achieved during the training (although for a polynomial degree of 3, the minimum loss is slightly lower than that of a polynomial degree of 2). This trend also explains the increase in error due to the implementation of high-order polynomials in shared KAN layers. Additionally, the number of epochs required to reach this minimum is greater for all polynomial degrees compared to degree 2. The left panel of Fig. \ref{Fig8} shows the loss evolution for different polynomial degrees. Accordingly, the loss evolution exhibits non-smooth behavior as the polynomial degree increases. This instability may arise due to oscillatory effects introduced by high-order polynomials. The loss values exhibit fluctuations at certain epochs, requiring additional optimization steps to stabilize, thereby increasing computational cost. Hence, PI-KAN-PointNet with a polynomial degree of 2 provides the optimal choice in terms of both computational cost and minimizing error.


\begin{table}[width=1.0\linewidth,cols=7,pos=!htbp]
\caption{Error analysis of the velocity, temperature, and pressure fields of 135 geometries predicted by the physics-informed KAN PointNet for different values of $\alpha$ and $\beta$ in Jacobi polynomials. The degree of the Jacobi polynomials used is 2. Here, $n_s=0.5$ is set. $||\cdot||_V$ indicates the $L^2$ norm over the entire domain $V$ and $||\cdot||_\Gamma$ indicates the $L^2$ norm over the inner cylinder surface.}
\label{Table3}
\begin{tabular*}{\tblwidth}{@{} LLLLLLL@{} }
\toprule
 & $\alpha = \beta = 0$ & $\alpha = \beta = -0.5$ & $\alpha = \beta = 0.5$ & $\alpha = \beta = 1$ & $2 \alpha = \beta = 2$ & $ \alpha = 2\beta = 2$ \\
\midrule
Average $||\Tilde{u}-u||_V/||u||_V$ & 1.08996E$-$1 & 1.08973E$-$1 & 1.21563E$-$1 & 1.13304E$-$1 & 1.21449E$-$1 & 1.21910E$-$1 \\
Maximum $||\Tilde{u}-u||_V/||u||_V$ & 1.32383E$-$1 & 1.31928E$-$1 & 1.44929E$-$1 & 1.39832E$-$1 & 1.45608E$-$1 & 1.42975E$-$1 \\
Minimum $||\Tilde{u}-u||_V/||u||_V$ & 8.96139E$-$2 & 8.78500E$-$2 & 1.02337E$-$1 & 9.30357E$-$2 & 1.02677E$-$1 & 1.05169E$-$1 \\
\midrule
Average $||\Tilde{v}-v||_V/||v||_V$ & 9.73026E$-$2 & 8.84278E$-$2 & 9.79996E$-$2 & 1.00921E$-$1 & 9.18341E$-$2 & 9.05697E$-$2 \\
Maximum $||\Tilde{v}-v||_V/||v||_V$ & 1.20330E$-$1 & 1.11167E$-$1 & 1.29262E$-$1 & 1.32655E$-$1 & 1.19910E$-$1 & 1.13413E$-$1 \\
Minimum $||\Tilde{v}-v||_V/||v||_V$ & 7.65053E$-$2 & 7.06122E$-$2 & 7.67489E$-$2 & 7.62320E$-$2 & 7.00090E$-$2 & 7.51902E$-$2 \\
\midrule
Average $||\Tilde{p}-p||_V/||p||_V$ & 3.32575E$-$2 & 2.97057E$-$2 & 3.57642E$-$2 & 3.13464E$-$2 & 2.61337E$-$2 & 3.67067E$-$2 \\
Maximum $||\Tilde{p}-p||_V/||p||_V$ & 3.76855E$-$2 & 3.41539E$-$2 & 4.23057E$-$2 & 3.69046E$-$2 & 3.04176E$-$2 & 4.18151E$-$2 \\
Minimum $||\Tilde{p}-p||_V/||p||_V$ & 2.88493E$-$2 & 2.63087E$-$2 & 3.12100E$-$2 & 2.80267E$-$2 & 2.25251E$-$2 & 3.35536E$-$2 \\
\midrule
Average $||\Tilde{T}-T||_V/||T||_V$ & 2.73711E$-$2 & 2.84486E$-$2 & 2.75284E$-$2 & 2.96868E$-$2 & 2.99346E$-$2 & 2.72577E$-$2 \\
Maximum $||\Tilde{T}-T||_V/||T||_V$ & 3.46309E$-$2 & 3.66225E$-$2 & 3.67476E$-$2 & 3.94755E$-$2 & 3.80191E$-$2 & 3.45827E$-$2 \\
Minimum $||\Tilde{T}-T||_V/||T||_V$ & 2.14836E$-$2 & 2.28196E$-$2 & 2.17336E$-$2 & 2.34157E$-$2 & 2.40676E$-$2 & 2.26196E$-$2 \\
\midrule
Average $||\Tilde{T}-T||_{\Gamma}/||T||_{\Gamma}$ & 8.20696E$-$3 & 5.03499E$-$3 & 6.73664E$-$3 & 7.29142E$-$3 & 6.41857E$-$3 & 7.30220E$-$3 \\
Maximum $||\Tilde{T}-T||_{\Gamma}/||T||_{\Gamma}$ & 1.73272E$-$2 & 9.08389E$-$3 & 1.14712E$-$2 & 1.43823E$-$2 & 9.60926E$-$3 & 1.31027E$-$2 \\
Minimum $||\Tilde{T}-T||_{\Gamma}/||T||_{\Gamma}$ & 2.24344E$-$3 & 2.59714E$-$3 & 3.06772E$-$3 & 3.33082E$-$3 & 3.15150E$-$3 & 4.02504E$-$3 \\
\midrule
Minimum loss achieved & 3.45998E$-$4 & 3.03649E$-$4  & 3.62887E$-$4 & 3.29284E$-$4 & 4.25615E$-$4 & 3.58046E$-$4 \\
\midrule
Number of epochs to & 2488 & 2238 & 2097 & 2437 & 2009 & 2301 \\
reach the minimum loss &  &  &  &  &  \\
\bottomrule
\end{tabular*}
\end{table}


\subsubsection{Effect of Jacobi polynomial type}
\label{Sect613}

Table \ref{Table3} presents the error analysis for velocity, pressure, and temperature fields predicted by PI-KAN-PointNet across 135 geometries ($m=135$), considering different values of $\alpha$ and $\beta$ in Jacobi polynomials. The degree of the polynomials is set to 2, with $n_s = 0.5$ held constant. Recall that the choice of $\alpha=\beta = 0$ leads to the Legendre polynomial, while selecting $\alpha=\beta = -0.5$ and $\alpha=\beta = 0.5$ corresponds to the Chebyshev polynomials of the first and second kinds, respectively. The Gegenbauer polynomial (or ultraspherical polynomials) is obtained when $\alpha=\beta$. As can be inferred from Table \ref{Table3}, all these types of polynomials overall demonstrate good performance in predicting the desired fields. Accordingly, the average pointwise relative error in the velocity field is lowest when $\alpha = \beta = -0.5$ compared to other choices. A similar trend is observed for the average error in predicting the temperature on the surface of the inner cylinder. The temperature distribution predicted by PI-KAN-PointNet on the surface of the inner cylinder with octagonal, heptagonal, and nonagonal shapes for different values of $\alpha$ and $\beta$ is shown in the right panel of Fig. \ref{Fig7}. Based on this, the most challenging aspect of the prediction for all cases occurs at the sharp corners, particularly for the heptagonal shape. For the octagonal and nonagonal shapes, the Chebyshev polynomial of the first kind (i.e., $\alpha=\beta=-0.5$) demonstrates superior performance compared to other polynomial types. This superiority stems from the distribution of Chebyshev polynomial roots within their interval, which helps minimize absolute errors and reduce potential oscillatory behavior. The advantage of Chebyshev polynomials over other Jacobi polynomial variants in implementing KANs has been recognized in the literature  \cite{shukla2024comprehensive,KANwithTANH}. Remarkably, PI-KAN-PointNet with the choice of $2\alpha=\beta = 2$ and $\alpha=2\beta = 2$ results in an average relative error of approximately 13\% for the $u$-component of the velocity vector, making these two members of the Jacobi polynomial family less suitable for engineering applications.

According to Table \ref{Table3}, from a training perspective, the lowest minimum loss is achieved when $\alpha = \beta = -0.5$. However, this configuration requires more epochs to converge compared to $\alpha = \beta = 0.5$ and $2\alpha=\beta = 2$. The right panel of Fig. \ref{Fig8} illustrates the evolution of the loss function for different values of $\alpha$ and $\beta$. As shown in Fig. \ref{Fig8}, all polynomial types exhibit oscillatory behavior, characterized by sharp jumps at certain epochs during training. Notably, the Chebyshev polynomials of the first and second kinds (i.e., $\alpha = \beta = -0.5$ and $\alpha = \beta = 0.5$) demonstrate smoother convergence, with smaller and less frequent oscillations, which may indicate greater training stability. In contrast, the Jacobi polynomials with settings $\alpha=2\beta=2$ and $\alpha=\beta=0$ exhibit the most pronounced oscillations, with large abrupt jumps in the loss function. Our conclusion is that, overall, the Chebyshev polynomial demonstrates superior performance, particularly its first kind, compared to other types of Jacobi polynomials. Considering our discussion in Sect. \ref{Sect612}, we conclude that the effect of the polynomial degree is more pronounced than the effect of the polynomial type.


\begin{table}[width=1.0\linewidth,cols=3,pos=!htbp]
\caption{Performance comparison of physics-informed PointNet with MLP and physics-informed PointNet with KAN for predicting the velocity, pressure, and temperature fields across 135 geometries. In physics-informed PointNet with KAN, the Jacobi polynomial degree is set to 2, with $\alpha=\beta=-0.5$ and $n_s=0.5$. In physics-informed PointNet with MLP, $n_s=0.85$. $||\cdot||_V$ denotes the $L^2$ norm over the entire domain $V$ and $||\cdot||_\Gamma$ represents the $L^2$ norm over the inner cylinder surface.}
\label{Table4}
\begin{tabular*}{\tblwidth}{@{} LLL@{} }
\toprule
 & Physics-informed PointNet with MLP & Physics-informed PointNet with KAN  \\
\midrule
Average $||\Tilde{u}-u||_V/||u||_V$ & 1.18367E$-$1 & 1.08973E$-$1  \\
Maximum $||\Tilde{u}-u||_V/||u||_V$ & 1.37381E$-$1 & 1.31928E$-$1  \\
Minimum $||\Tilde{u}-u||_V/||u||_V$ & 1.02223E$-$1 & 8.78500E$-$2  \\
\midrule
Average $||\Tilde{v}-v||_V/||v||_V$ & 8.39424E$-$2 & 8.84278E$-$2  \\
Maximum $||\Tilde{v}-v||_V/||v||_V$ & 1.05331E$-$1 & 1.11167E$-$1  \\
Minimum $||\Tilde{v}-v||_V/||v||_V$ & 6.86013E$-$2 & 7.06122E$-$2  \\
\midrule
Average $||\Tilde{p}-p||_V/||p||_V$ & 2.61402E$-$2 & 2.97057E$-$2  \\
Maximum $||\Tilde{p}-p||_V/||p||_V$ & 2.91599E$-$2 & 3.41539E$-$2  \\
Minimum $||\Tilde{p}-p||_V/||p||_V$ & 2.04838E$-$2 & 2.63087E$-$2  \\
\midrule
Average $||\Tilde{T}-T||_V/||T||_V$ & 2.58254E$-$2 & 2.84486E$-$2  \\
Maximum $||\Tilde{T}-T||_V/||T||_V$ & 3.10124E$-$2 & 3.66225E$-$2  \\
Minimum $||\Tilde{T}-T||_V/||T||_V$ & 2.00219E$-$2 & 2.28196E$-$2  \\
\midrule
Average $||\Tilde{T}-T||_{\Gamma}/||T||_{\Gamma}$ & 1.45697E$-$2 & 5.03499E$-$3  \\
Maximum $||\Tilde{T}-T||_{\Gamma}/||T||_{\Gamma}$ & 1.86680E$-$2 & 9.08389E$-$3  \\
Minimum $||\Tilde{T}-T||_{\Gamma}/||T||_{\Gamma}$ & 9.04055E$-$3 & 2.59714E$-$3  \\
\midrule
Minimum loss achieved & 3.72039E$-$4 & 3.03649E$-$4  \\
\midrule
Training time & 15.6 & 15.4 \\
per epoch (s) &  &   \\
\midrule
Number of epochs to & 2324 & 2238  \\
reach the minimum loss &  &   \\
\midrule
Number of trainable & 639611 &  666880 \\
parameters &  &   \\
\bottomrule
\end{tabular*}
\end{table}


\subsection{Comparison with physics-informed PointNet with MLP}
\label{Sect62}

\subsubsection{A brief overview of physics-informed PointNet with MLPs}
\label{Sect621}

The details of the technical implementation and formulation of physics-informed PointNet with MLPs (commonly referred to as PIPN) are presented in Refs. \citep{kashefi2022physics,kashefi2023PIPNelasticity}. Here, we provide a brief overview and highlight the key differences between PIPN and PI-KAN-PointNet. The architecture of physics-informed PointNet with MLPs is shown in Fig. 2 of Ref. \citep{kashefi2022physics} and Fig. 2 of Ref. \citep{kashefi2023PIPNelasticity}. At a high level, its structure is similar to what we present in Fig. \ref{Fig1} of the current article, with the primary difference being that instead of using shared KANs, we implement shared MLPs. An MLP consists of several fully connected (FC) layers, where the output of one layer serves as the input to the next. If the input to an FC layer is $\mathbf{r}$, then the output $\mathbf{s}$ of this layer is computed as

\begin{equation}
    \mathbf{s}_{d_\text{output}\times 1} = \sigma( \mathbf{W}_{d_\text{output}\times d_\text{input}} \mathbf{r}_{d_\text{input}\times 1} + \mathbf{b}_{d_\text{output}\times 1}),
\end{equation}
where $\mathbf{W}$ and $\mathbf{b}$ are the weight matrix and bias vector of the FC layer, respectively. $d_\text{input}$ and $d_\text{output}$ are defined similarly to Sect. \ref{Sect3}. The activation function of the FC layer, denoted by $\sigma$, is set to the hyperbolic tangent function, defined as  

\begin{equation}
    \tanh(\gamma) = \frac{e^{2\gamma} - 1}{e^{2\gamma} + 1}.
    \label{Eq31}
\end{equation}
It is important to note that since Eqs. \ref{Eq2}--\ref{Eq3} contain a second-order derivative term, the activation function must be twice differentiable for this specific application in physics-informed PointNet with shared MLPs. Further details about the PIPN architecture can be found in Refs. \citep{kashefi2022physics,kashefi2023PIPNelasticity}.

\subsubsection{PI-KAN-PointNet versus physics-informed PointNet with MLPs}
\label{Sect622}

To ensure a fair comparison between PI-KAN-PointNet and physics-informed PointNet with MLPs, we consider two configurations of these models such that the number of trainable parameters is approximately equal in both. Alternatively, fairness in comparison may also be considered in terms of training time (i.e., equal computational cost). To achieve this, we use PI-KAN-PointNet with a Chebyshev polynomial (\(\alpha = \beta = -0.5\)) of degree 2 and set \(n_s = 0.5\). In physics-informed PointNet with MLPs, we set \(n_s = 0.85\). Under these parameter choices, the number of trainable parameters is approximately 666880 for PI-KAN-PointNet and 639611 for physics-informed PointNet with MLPs.

\begin{figure}[!htbp]
  \centering 
      \begin{subfigure}[b]{0.49\textwidth}
        \centering
        \includegraphics[width=\textwidth]{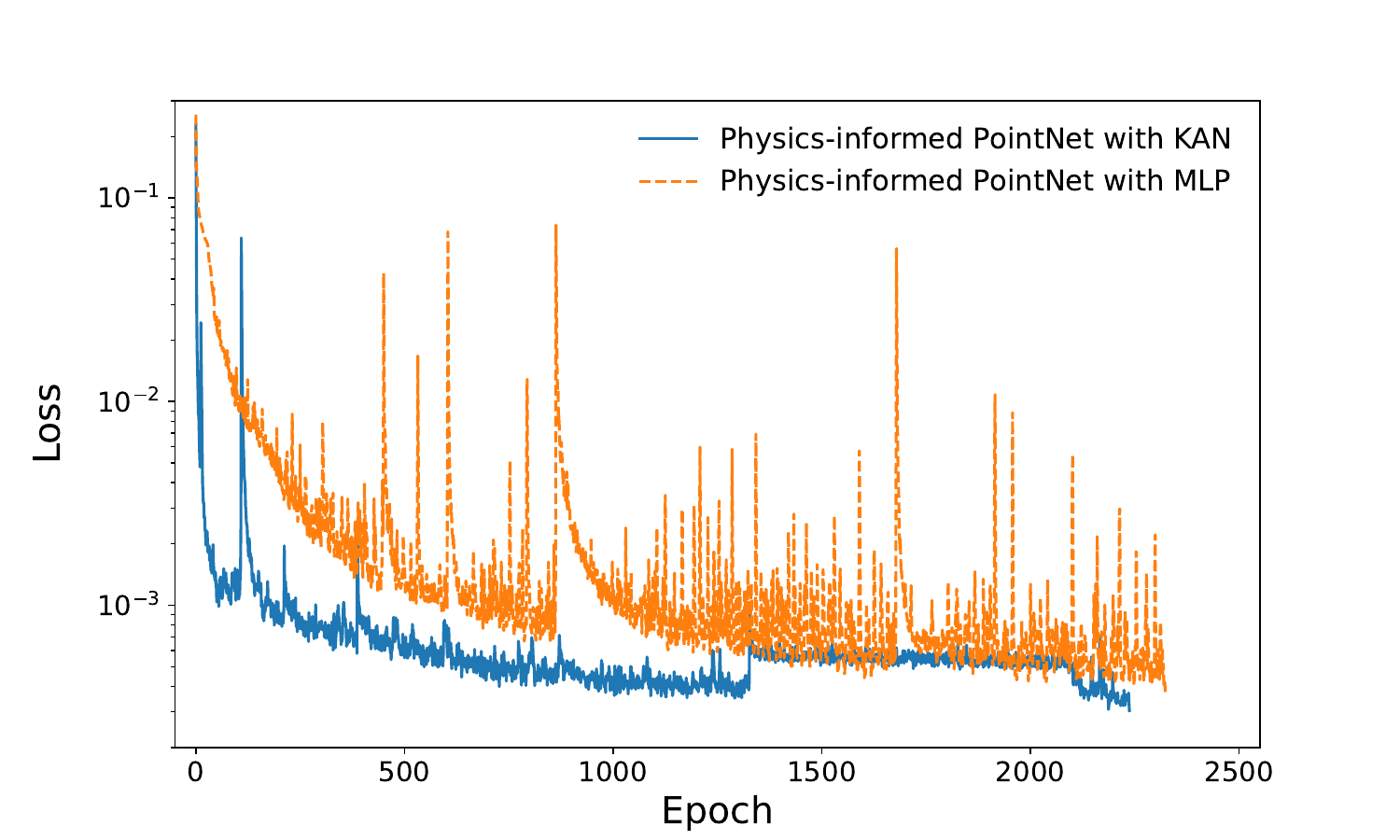}
    \end{subfigure}
    \begin{subfigure}[b]{0.49\textwidth}
        \centering
        \includegraphics[width=\textwidth]{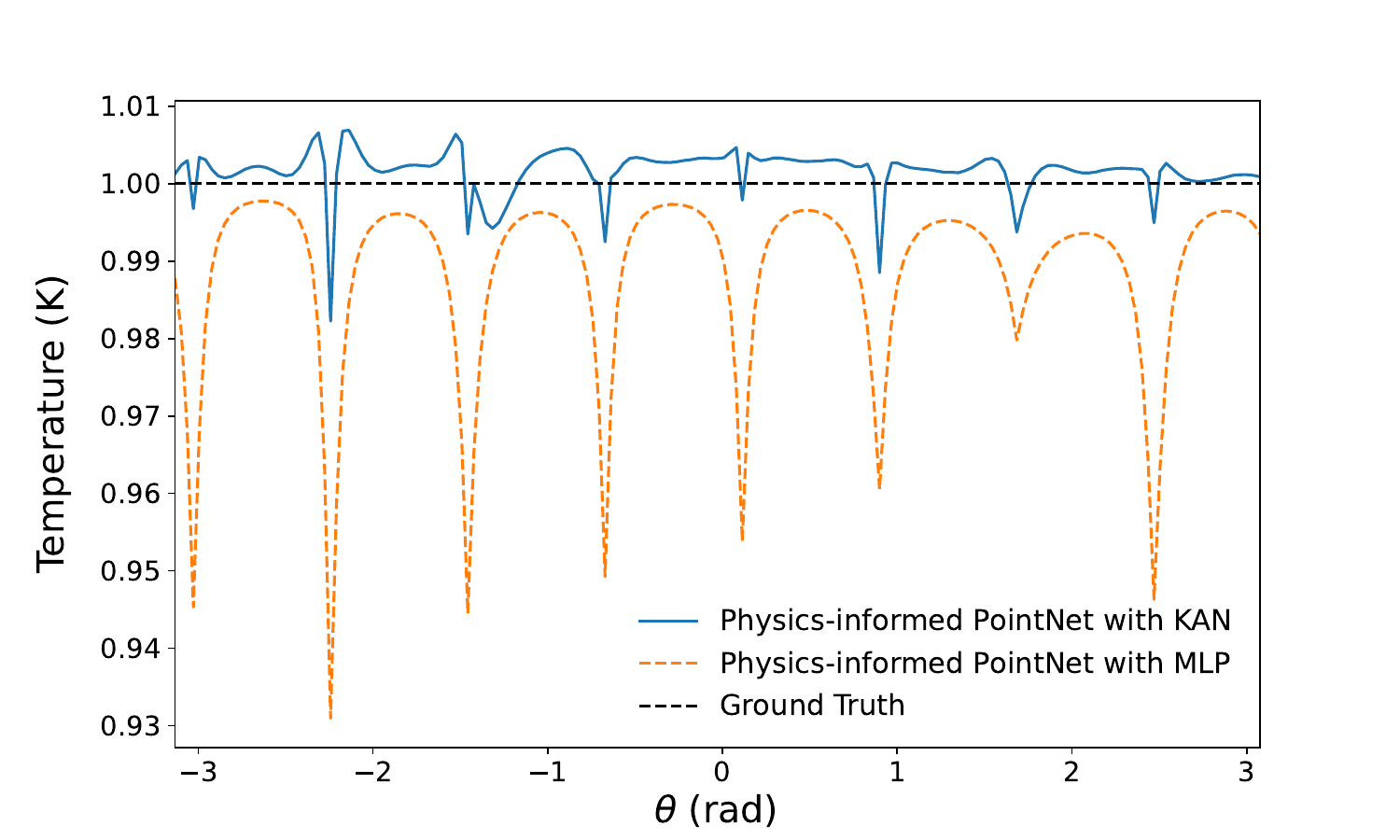}
    \end{subfigure}
    
 \caption{Comparison of loss evolution (left) and temperature distribution prediction along the cylinder surface for the octagon with $\Omega=31^\circ$ (right) between the physics-informed PointNet with KAN and the physics-informed PointNet with MLP. In physics-informed PointNet with KAN, the Jacobi polynomial degree is set to 2, with $\alpha=\beta=-0.5$ and $n_s=0.5$. In physics-informed PointNet with MLP, $n_s=0.85$. The angle $\theta$ is defined with reference to the positive $x$-axis and increases counterclockwise (or decreases clockwise). See Table \ref{Table1} and the text for the definition of $\Omega$.}
  \label{Fig9}
\end{figure}

The results of simultaneously solving the inverse problem on 135 geometries ($m=135$) using both models are presented in Table \ref{Table4}. The minimum loss achieved by PI-KAN-PointNet is approximately \(3.04 \times 10^{-4}\), whereas for physics-informed PointNet with MLPs, it is \(3.73 \times 10^{-4}\). Additionally, PI-KAN-PointNet reaches this minimum loss 100 epochs earlier than physics-informed PointNet with MLPs (2238 vs. 2324), while the required training time per epoch remains nearly identical (154 s vs. 156 s). The loss evolution during training for both models is shown in the left panel of Fig. \ref{Fig9}, where PI-KAN-PointNet exhibits significantly fewer fluctuations compared to physics-informed PointNet, demonstrating greater stability.

Based on the information tabulated in Table \ref{Table4}, the average relative pointwise error (\(L^2\) norm) of the velocity, pressure, and temperature fields predicted by both models is of the same order of accuracy. However, PI-KAN-PointNet's prediction of the temperature on the surface of the inner cylinder is significantly more accurate than that of physics-informed PointNet with MLPs, with average relative errors of approximately 0.5\% and 1.4\%, respectively.

These results highlight the capability of KANs compared to MLPs. For a more detailed analysis, the temperature distribution on the surface of an inner cylinder with an octagonal shape is plotted in the right panel of Fig. \ref{Fig9}. As shown in Fig. \ref{Fig9}, the prediction of physics-informed PointNet with shared MLPs is always less than 1 because the hyperbolic tangent activation function restricts the output range to \([-1,1]\). This limitation reduces the flexibility of the network and results in less accurate predictions. In contrast, since the activation function in PI-KAN-PointNet is learnable, it can better approximate the ground truth of 1, as it is not constrained by a fixed output range. Additionally, this flexibility allows PI-KAN-PointNet to provide more accurate predictions for the sharp corners of the octagonal shape compared to physics-informed PointNet with MLPs.

A more detailed examination of the data in Table \ref{Table4} reveals that PI-KAN-PointNet predicts the $u$-component of the velocity vector slightly more accurately than physics-informed PointNet with MLPs. On the other hand, physics-informed PointNet with MLPs provides slightly more accurate predictions for the $v$-component of the velocity vector, pressure, and temperature field across the entire domain. Given these trends, determining which model demonstrates superior performance is essential. We argue that PI-KAN-PointNet is more successful, as it achieves a lower minimum loss compared to physics-informed PointNet with MLPs. Note that the objective of both networks is to minimize the loss function (Eq. \ref{Eq30}). From an engineering and practical perspective, the exact solutions of the velocity, pressure, and temperature fields are unknown. Therefore, the model yielding the smallest loss function values is expected to be more reliable.


\begin{figure}[!htbp]
  \centering 
\includegraphics[width=\textwidth]{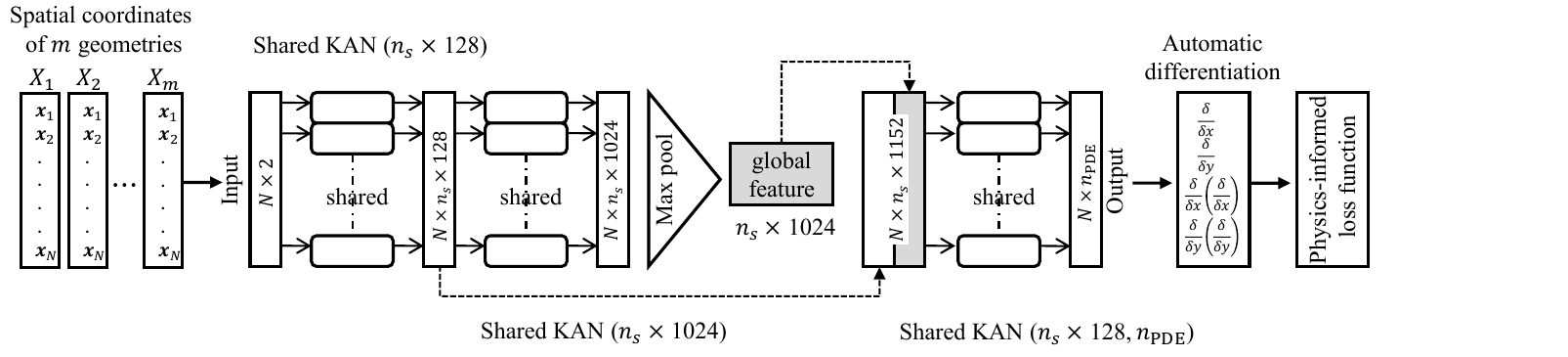}
 \caption{Architecture of the lightweight PI-KAN-PointNet. In this study, we set $n_s = 1$. Other components are defined similarly to the caption of Fig. \ref{Fig1}.}
  \label{Fig900}
\end{figure}


\subsection{Lightweight alternative architecture for physics-informed KAN PointNet}
\label{Sect63}

\citet{kashefi2024pointnetKAN3D} proposed PointNet with KAN for classification and segmentation of 3D point clouds and demonstrated that, while PointNet with KAN achieves the same performance as PointNet with MLPs, it requires a relatively shallower network (see Fig. 1 of Ref. \cite{kashefi2024pointnetKAN3D}) compared to PointNet with shared MLPs (see Fig. 2 and Fig. 9 of Ref. \cite{qi2017pointnet}). More specifically, \citet{kashefi2024pointnetKAN3D} showed that using just two layers of shared KANs in the encoder and a single layer of shared KAN in the decoder is sufficient for the part segmentation task of three-dimensional point sets. This finding motivates us to investigate whether a shallower and more lightweight PI-KAN-PointNet can achieve comparable accuracy. However, the answer to this question likely depends on the dataset size. Intuitively, deeper neural networks may be necessary for large datasets. Nevertheless, we first aim to answer this question in the context of the current dataset. To this end, we consider an alternative architecture of PI-KAN-PointNet, which is shown in Fig. \ref{Fig900}. Specifically, we use shared KAN layers with Chebyshev polynomials of degree 2, defined by $\alpha = \beta = -0.5$, and set $n_s = 1$. As illustrated in Fig. \ref{Fig900}, the input first passes through a shared KAN layer, converting it into an intermediate feature space of size 128. These local features are then processed by a second shared KAN layer, expanding them into a higher-dimensional space of size 1024. A max pooling operation extracts a global feature representing the entire point cloud, which is then expanded to match the number of points. The final combined feature, comprising local features of size 128 and a global feature of size 1024, undergoes another shared KAN layer to reduce the feature size to 128. A final shared KAN layer predicts the velocity, pressure, and temperature fields. Batch normalization is applied after each KAN layer, except for the last one.

The results indicate that the average relative pointwise error ($L^2$ norm) for the $u$-component of velocity is 1.23189E$-$1, while for the $v$-component, it is 9.09591E$-$2. The pressure field exhibits an average relative error of 2.96761E$\--$2, and the temperature field achieves an average relative error of 2.51027E$-$2. Additionally, the temperature error measured specifically on the surface of the inner cylinder is 1.81828E$-$2. The minimum loss achieved is 6.80781E$-$4, and the training time per epoch is 7.6 s. Compared to the performance of full PI-KAN-PointNet presented in Table \ref{Table1}, although the computational cost per epoch for the lightweight PI-KAN-PointNet is significantly lower, as expected, the minimum loss—and consequently, the error in predicting the desired fields—increases. This effect is particularly noticeable for the $u$-component of the velocity vector, where the relative error exceeds 10\%, surpassing the commonly accepted engineering threshold. However, the errors in the pressure and temperature fields remain within acceptable limits. We conclude that reducing the depth of PI-KAN-PointNet significantly impacts its performance, in contrast to the application of PointNet with KAN in part segmentation tasks in computer graphics.


\begin{figure}[!htbp]
  \centering 
        \includegraphics[width=\textwidth]{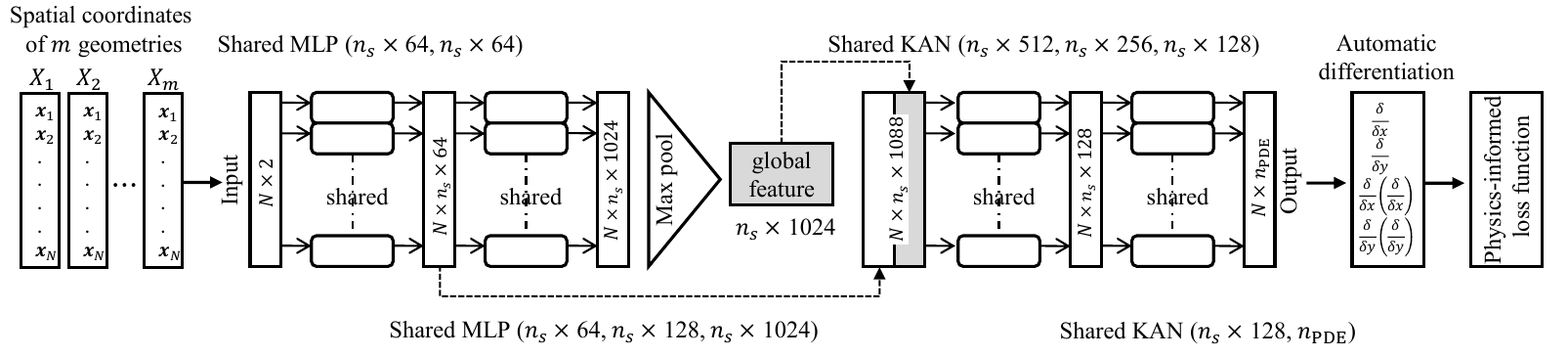}
 \caption{Architecture of the physics-informed PointNet with shared MLPs in the encoder and shared KANs in the decoder. In this study, we set $n_s = 1$ in the encoder, while $n_s = 0.5$ in the decoder. Other components are defined similarly to the caption of Fig. \ref{Fig1}.}
  \label{Fig100}
\end{figure}

\begin{figure}[!htbp]
  \centering 
        \includegraphics[width=\textwidth]{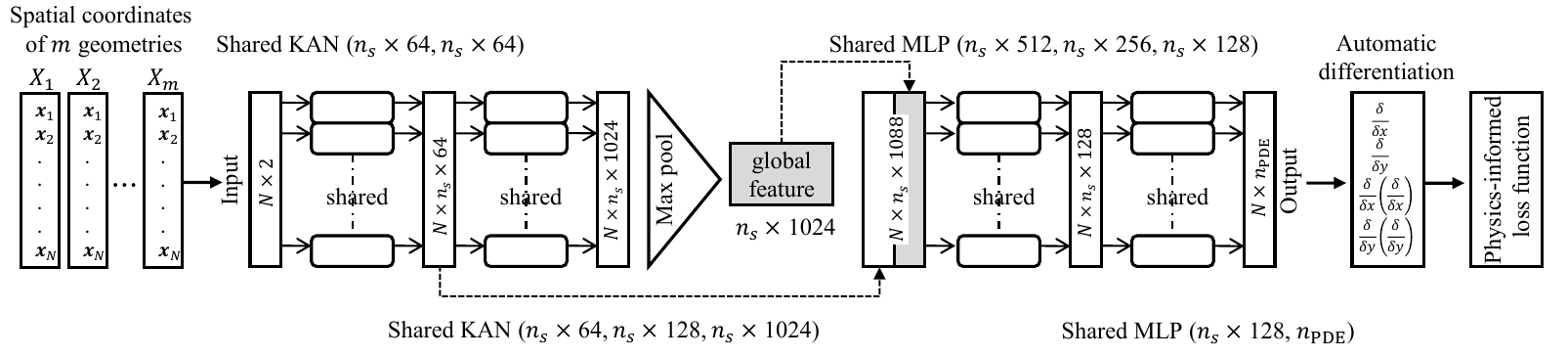}
  \caption{Architecture of the physics-informed PointNet with shared KANs in the encoder and shared MLPs in the decoder. In this study, we set $n_s = 0.5$ in the encoder, while $n_s = 1.0$ in the decoder. Other components are defined similarly to the caption of Fig. \ref{Fig1}.}
  \label{Fig101}
\end{figure}


\begin{table}[width=1.0\linewidth,cols=3]
\caption{Performance comparison of physics-informed PointNet with KAN encoder and MLP decoder and physics-informed PointNet with MLP encoder and KAN decoder for predicting the velocity, pressure, and temperature fields across 135 geometries. In KANs, the Jacobi polynomial degree is set to 2, with $\alpha=\beta=-0.5$ and $n_s=0.5$. In MLPs, $n_s=1.0$. $||\cdot||_V$ represents the $L^2$ norm over the entire domain $V$ and $||\cdot||_\Gamma$ denotes the $L^2$ norm over the inner cylinder surface.}
\label{Table6}
\begin{tabular*}{\tblwidth}{@{} LLL@{} }
\toprule
 & Physics-informed PointNet with & Physics-informed PointNet with \\
  & KAN encoder and MLP decoder & MLP encoder and KAN decoder  \\
\midrule
Average $||\Tilde{u}-u||_V/||u||_V$ & 1.25273E$-$1 & 8.62143E$-$2  \\
Maximum $||\Tilde{u}-u||_V/||u||_V$ & 1.45714E$-$1 & 9.63481E$-$2  \\
Minimum $||\Tilde{u}-u||_V/||u||_V$ & 1.10891E$-$1 & 7.54883E$-$2  \\
\midrule
Average $||\Tilde{v}-v||_V/||v||_V$ & 9.02736E$-$2 & 8.85543E$-$2  \\
Maximum $||\Tilde{v}-v||_V/||v||_V$ & 1.11110E$-$1 & 1.02025E$-$1  \\
Minimum $||\Tilde{v}-v||_V/||v||_V$ & 7.55460E$-$2 & 7.46082E$-$2  \\
\midrule
Average $||\Tilde{p}-p||_V/||p||_V$ & 2.73488E$-$2 & 2.98410E$-$2  \\
Maximum $||\Tilde{p}-p||_V/||p||_V$ & 3.22359E$-$2 & 3.24039E$-$2  \\
Minimum $||\Tilde{p}-p||_V/||p||_V$ & 2.25765E$-$2 & 2.76343E$-$2  \\
\midrule
Average $||\Tilde{T}-T||_V/||T||_V$ & 2.52252E$-$2 & 2.21639E$-$2  \\
Maximum $||\Tilde{T}-T||_V/||T||_V$ & 3.09241E$-$2 & 2.82095E$-$2  \\
Minimum $||\Tilde{T}-T||_V/||T||_V$ & 1.98086E$-$2 & 1.80277E$-$2  \\
\midrule
Average $||\Tilde{T}-T||_{\Gamma}/||T||_{\Gamma}$ & 1.05230E$-$2 & 3.46168E$-$3  \\
Maximum $||\Tilde{T}-T||_{\Gamma}/||T||_{\Gamma}$ & 1.30571E$-$2 & 5.33721E$-$3  \\
Minimum $||\Tilde{T}-T||_{\Gamma}/||T||_{\Gamma}$ & 5.91177E$-$3 & 2.56398E$-$3  \\
\midrule
Minimum loss achieved & 4.58812E$-$4 & 1.25308E$-$4  \\
\midrule
Training time & 23.0 & 25.49  \\
per epoch (s) &  &   \\
\midrule
Number of epochs to &  2396 &  2498 \\
reach the minimum loss &  &   \\
\bottomrule
\end{tabular*}
\end{table}


\subsection{Physics-informed PointNet with an MLP (KAN) encoder and KAN (MLP) decoder}
\label{Sect64}

After a comprehensive evaluation of PI-KAN-PointNet and an in-depth comparison with physics-informed PointNet with shared MLPs, an idea worth investigating is the integration of both shared KANs and MLPs into PointNet to potentially leverage the strengths of both architectures. There are multiple ways to achieve this integration. One reasonable and straightforward approach is to construct the encoder of PointNet using shared KAN layers while utilizing shared MLP layers for the decoder, or vice versa. Figure \ref{Fig100} and Figure \ref{Fig101} illustrate the schematic representation of these two proposed models. Specifically, in shred KAN layers, the Chebyshev polynomial (i.e., $\alpha=\beta = -0.5$) with a degree of 2 and $n_s=0.5$ was used, while in shared MLP layers, $n_s$ was set to 1.

\begin{figure}[!htbp]
  \centering 
      \begin{subfigure}[b]{0.49\textwidth}
        \centering
        \includegraphics[width=\textwidth]{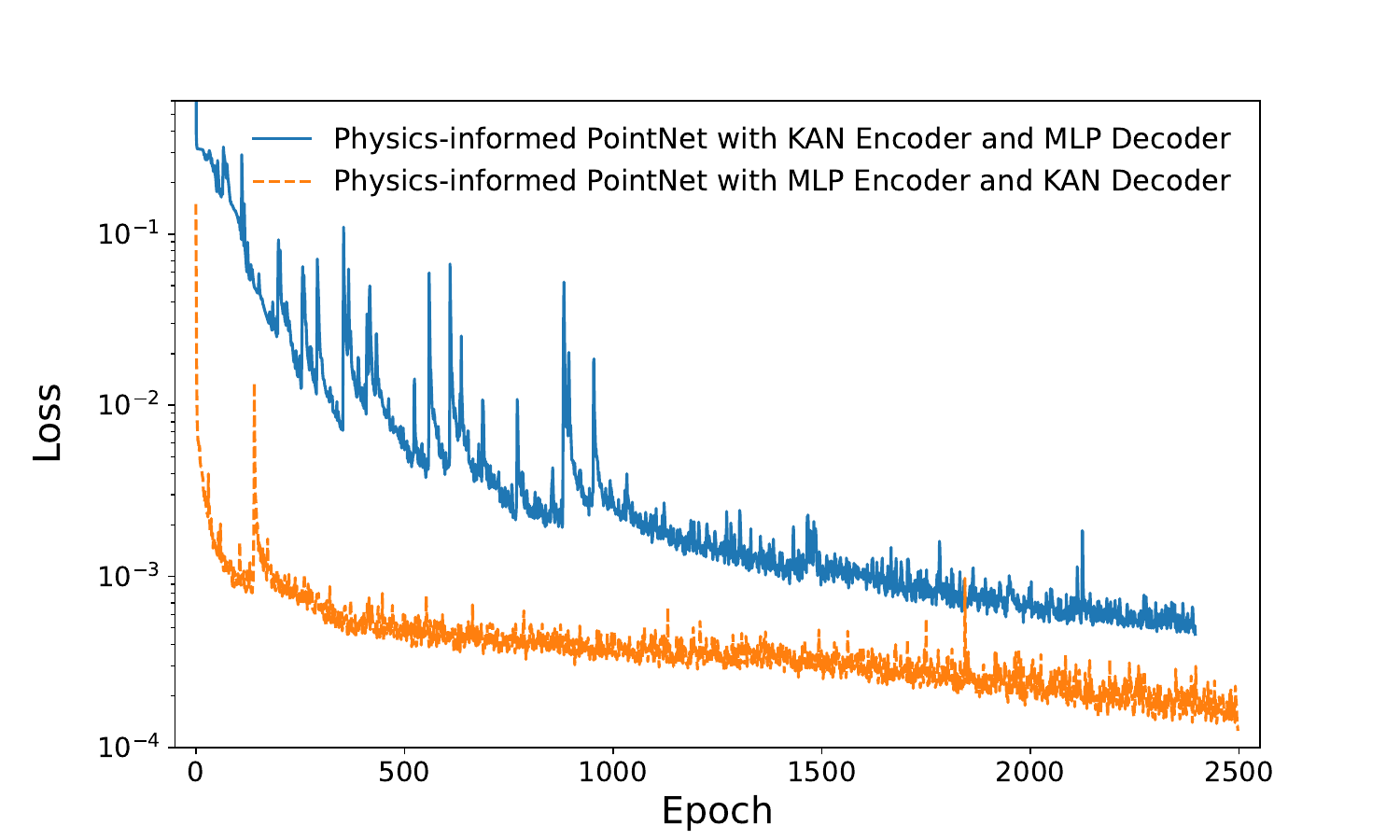}
    \end{subfigure}
    \begin{subfigure}[b]{0.49\textwidth}
        \centering
        \includegraphics[width=\textwidth]{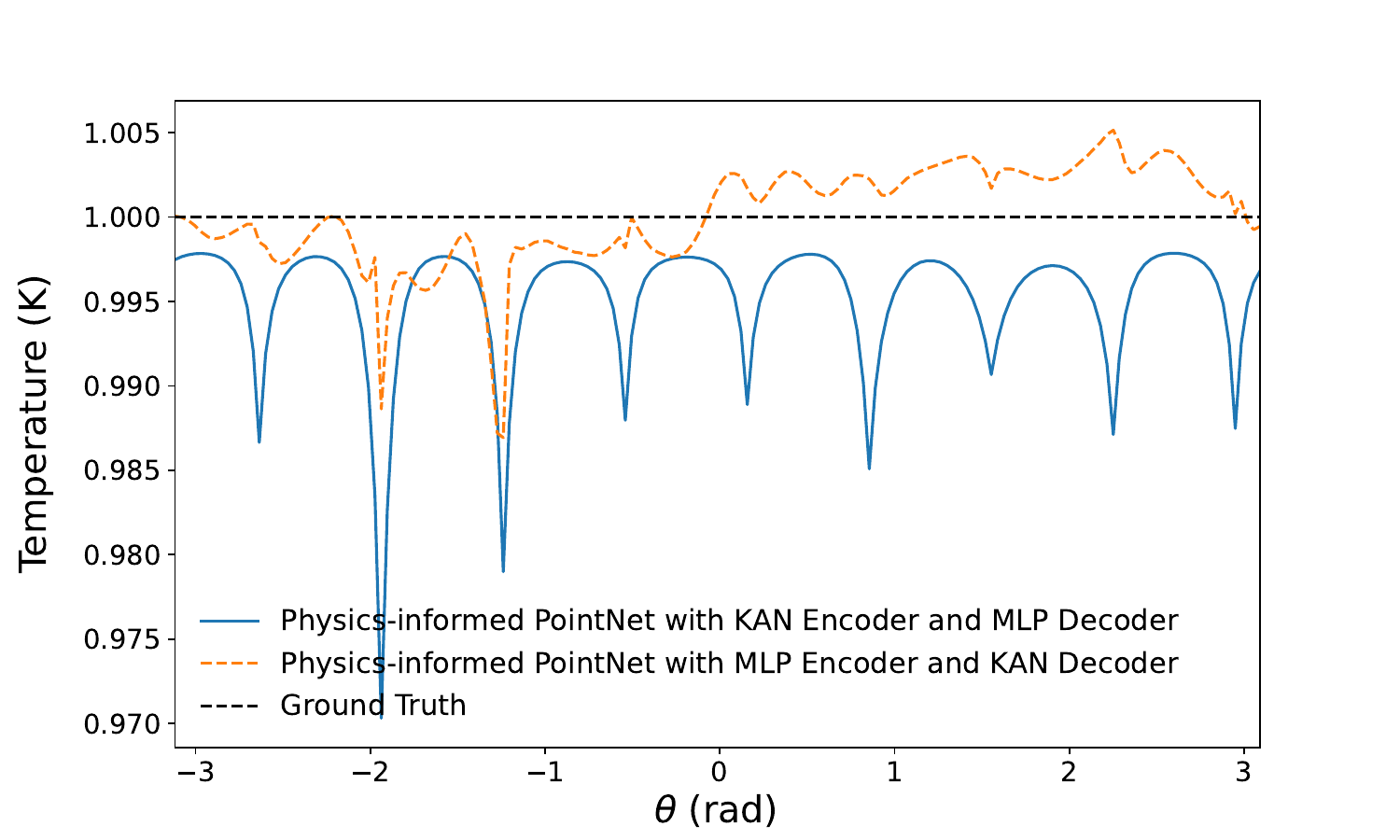}
    \end{subfigure}
    
 \caption{Comparison of loss evolution (left) and temperature distribution prediction along the cylinder surface for the octagon with $\Omega=31^\circ$ between the physics-informed PointNet with a KAN encoder and MLP decoder, and the physics-informed PointNet with an MLP encoder and KAN decoder. In KANs, the Jacobi polynomial degree is set to 2, with $\alpha=\beta=-0.5$ and $n_s=0.5$. In MLPs, $n_s=1.0$. See the text for further details about the setup of these two models. The angle $\theta$ is defined with reference to the positive $x$-axis and increases counterclockwise (or decreases clockwise). See Table \ref{Table1} and the text for the definition of $\Omega$.}
  \label{Fig10}
\end{figure}

Table \ref{Table6} compares these two models in predicting velocity, pressure, and temperature fields across 135 geometries ($m=135$). The evaluation, based on relative pointwise error ($L^2$ norm) over the entire domain and the inner cylinder surface (for temperature), reveals that the configuration with the MLP encoder and KAN decoder generally produces lower errors in velocity and temperature predictions, particularly along the inner cylinder surface, while pressure predictions remain comparable between the two approaches. Additionally, this configuration achieves a lower minimum loss (1.25308E$-$4 vs. 4.58812E$-$4), although it requires slightly longer training times per epoch and a marginally increased number of epochs to converge. The loss evolution for both models is shown in the left panel of Fig. \ref{Fig10}. Accordingly, the loss function of the physics-informed PointNet with an MLP encoder and KAN decoder decreases to approximately $8.7\times10^{-4}$ after just 200 epochs, whereas this value remains at $3.8\times10^{-2}$ at the same epoch for the model with a KAN encoder and MLP decoder. Additionally, the loss plot of the first model exhibits fewer fluctuations and appears smoother.

Regarding the prediction of the temperature field on the surface of the inner cylinder as an unknown boundary condition in the inverse problem, the average relative error ($L^2$ norm) of the predicted quantity is approximately 0.346\% and 1.052\% using the first and second models, respectively. Additionally, an example of these predictions is visually presented in the right panel of Fig. \ref{Fig10}. Similar to the discussion in Sect. \ref{Sect622} regarding the temperature distribution in Fig. \ref{Fig9}, the learnable activation function in KAN layers provides greater flexibility compared to the fixed hyperbolic activation function in MLP layers. Since the first model employs KAN layers in the decoder, which is responsible for mapping the geometric features to the velocity, pressure, and temperature fields, it achieves better performance compared to the second model, which utilizes MLP layers in its decoder. The second reason for the success of the first model can be outlined as follows. The encoder is primarily responsible for extracting the geometric features of the point clouds. As shown in Fig. \ref{Fig1}, a max pooling operation is applied at the end of the encoder to construct the vector of global features. When using MLP layers with the hyperbolic tangent activation function, the score of each point is constrained within the range \([-1,1]\), leading to a more consistent comparison when computing the maximum values. In contrast, when KAN layers are used in the encoder, there is no restriction on the output range. Consequently, the distribution of scores among the point clouds may become uneven, and after applying the max pooling operator, the global features may not represent the geometric features as accurately as in the case of MLP layers.

At the end of this subsection, it is worthwhile noting that if we compare the physics-informed PointNet model, containing an encoder built on shared MLP layers and a decoder built on shared KAN layers, with those models consisting entirely of KAN layers discussed in the previous sections, particularly those listed in Table \ref{Table2} and Table \ref{Table3}, the results reveal that the combined MLP-KAN model provides more accurate predictions for the velocity (especially the $u$-component) and temperature fields while achieving the lowest loss among the compared models. Therefore, we suggest using the physics-informed PointNet with shared MLP layers as the encoder and shared KAN layers as the decoder as the optimal model.


\section{Summary and future studies}
\label{Sect7}

In this research article, we introduced Physics-Informed Kolmogorov-Arnold PointNet (PI-KAN-PointNet) as a novel machine learning framework for solving time-independent inverse problems. PI-KAN-PointNet allowed for the simultaneous solution of inverse problems across multiple irregular geometries in a single training run, reducing computational costs compared to conventional physics-informed neural networks, which required retraining for each new geometry. PI-KAN-PointNet combined PointNet \citep{qi2017pointnet} with KANs and employed a loss function based on the residuals of the governing equations and sparse data obtained from sensor locations, using the automatic differentiation tools in deep learning libraries such as PyTorch \citep{paszke2019pytorch} and TensorFlow \citep{tensorflow2015-whitepaper}. In contrast, PIPN \citep{kashefi2022physics} took the opposite approach, incorporating MLPs within the physics-informed PointNet framework instead of KANs. To construct KANs in PI-KAN-PointNet, we used different Jacobi polynomial families with varying degrees. We used the PyTorch library \citep{paszke2019pytorch} and Adam optimizer \citep{kingma2014adam}. As a benchmark test case, we considered natural convection in a square enclosure containing a cylinder across 135 different geometries. To assess PI-KAN-PointNet's performance, we conducted both quantitative and visual error analyses. Our findings showed that using a polynomial degree of 2 and the Chebyshev polynomial of the first kind resulted in the most accurate solution. A comparison between PI-KAN-PointNet and PIPN demonstrated that, for approximately the same computational cost and number of trainable parameters, PI-KAN-PointNet produced more accurate predictions, particularly for temperature estimation on the surface of inner cylinders, which served as unknown boundary conditions. Furthermore, we explored a hybrid architecture that integrated both KAN and MLP layers within the physics-informed PointNet framework. The results indicated that using MLPs in the encoder and KANs in the decoder improved prediction accuracy and represented the optimal configuration, outperforming both PIPN and PI-KAN-PointNet models that relied exclusively on MLPs or KANs.

There are several possible directions for extending this research. One of our plans is to expand the current framework to three-dimensional and unsteady problems, as well as to other areas of computational physics, such as nonlinear elasticity, plasticity, and compressible flows. Additionally, we aim to explore the integration of KANs with neural operators \citep{li2020fourier,anandkumar2020neural,bonev2023spherical,kashefi2024novelFNO}  and large language models \citep{lewkowycz2022solving,imani2023mathprompter,frieder2024mathematical,Kashefi2024misleading,kashefi2023chatgpt}. Furthermore, we plan to investigate the performance of PI-KAN-PointNet using alternative basis functions, including B-splines \citep{liu2024kan,liu2024kan2}, wavelet functions \cite{bozorgasl2024wav}, and residual basis functions \cite{li2024KANradial}. 


\section*{CRediT authorship contribution statement}
\textbf{Ali Kashefi:} Conceptualization, Methodology, Formal analysis, Software, Visualization, Writing - original draft, Writing - review \& editing. \textbf{Tapan Mukerji:} Conceptualization, Formal analysis, Writing - review \& editing, Funding acquisition, Resources.

\section*{Declaration of competing interest}
The authors declare that they have no known competing financial interests or personal relationships that could have appeared to influence the work reported in this paper.

\section*{Acknowledgement}
We gratefully acknowledge funding from the Shell-Stanford collaborative project. We also extend our thanks to the Stanford Research Computing Center for providing the computational resources that supported this research.

\section*{Data availability}

The Python codes and data are available on the following GitHub repository: \url{https://github.com/Ali-Stanford/PI-KAN-PointNet}.


\printcredits

\bibliographystyle{model1-num-names}

\bibliography{cas-refs}


\end{document}